\documentclass[11pt]{arxivTemplate}
\usepackage[utf8]{inputenc}
\usepackage[
    style=numeric-comp,
    doi=true,
    url=false,
    uniquename=false,
    giveninits=true,
    natbib,
    backend=biber]{biblatex}
\PassOptionsToPackage{most}{tcolorbox}
\usepackage{colortbl}

\usepackage[utf8]{inputenc} 
\usepackage[T1]{fontenc}    
\usepackage{url}            
\usepackage{booktabs}       
\usepackage{amsfonts}       
\usepackage{nicefrac}       
\usepackage{microtype}      
\usepackage{tabularx}
\usepackage{makecell}
\usepackage{amsmath}
\usepackage{multirow}
\usepackage{array}
\usepackage{caption}
\usepackage{wrapfig}
\usepackage{enumitem}
\usepackage{tikz}
\usepackage{lipsum}
\usepackage{subcaption}
\usepackage{multirow}
\usepackage{adjustbox}
\usepackage{varwidth}

\usepackage{amssymb}
\usepackage{unicode}  
\usepackage[capitalize]{cleveref} 

\usepackage{pgf}
\usepackage{colortbl}

\usepackage{tipa}

\usepackage{tcolorbox}
\usepackage{rotating}
\usepackage[abs]{overpic}
\usepackage{makecell}
\usepackage{longtable}

\usepackage{tocloft}  

\usepackage[english]{babel}
\usepackage{csquotes}
\usepackage{listings}
\definecolor{codekeyword}{rgb}{0.0, 0.0, 0.5}   
\definecolor{codecomment}{rgb}{0.0, 0.5, 0.0}   
\definecolor{codestring}{rgb}{0.56, 0.0, 1.0}   
\definecolor{deepgreen}{RGB}{0,100,45}

\usepackage{xcolor}
\usepackage{pifont}

\usepackage{algorithm}
\usepackage{algpseudocode}
\usepackage{lineno}

\usepackage{amsmath}
\usepackage{amssymb}

\newcommand{\greencheck}{\textcolor{deepgreen}{\ding{51}}}
\newcommand{\rederror}{\textcolor{red}{\ding{55}}}

\newcommand{\dataset}{PAI-Bench}

\definecolor{lightred}{HTML}{f6c6ad}
\definecolor{lightgreen}{HTML}{b4e5a2}
\definecolor{lightblue}{HTML}{a6caec}

\definecolor{darkgreen}{rgb}{0.0, 0.5, 0.0}
\definecolor{darkred}{rgb}{0.7, 0.0, 0.0}
\newcommand{\gain}[1]{\textcolor{darkgreen}{\small(+#1)}}
\newcommand{\loss}[1]{\textcolor{darkred}{\small(-#1)}}

\lstdefinestyle{pythonstyle}{
    language=Python,                          
    basicstyle=\ttfamily\small,               
    keywordstyle=\color{codekeyword}\bfseries,
    commentstyle=\color{codecomment}\itshape, 
    stringstyle=\color{codestring},           
    showstringspaces=false,                   
    breaklines=true,                          
    tabsize=4,                                
    numbers=none,                             
    frame=none,                               
    backgroundcolor=\color{white},            
    captionpos=b,                             
    morekeywords={self, __init__, __name__, __main__}, 
}

\usepackage{fontawesome}
\usepackage{xspace}
\newcommand{\huggingface}{\raisebox{-1.5pt}{\includegraphics[height=1.05em]{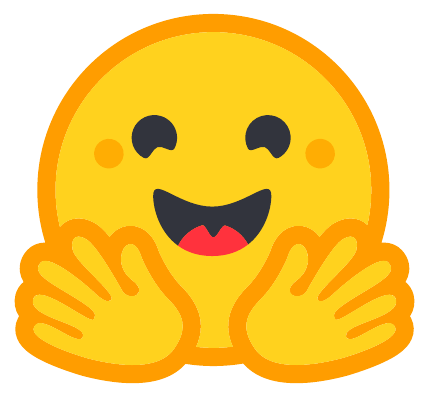}}\xspace}
\newcommand{\github}{\raisebox{-1.5pt}{\includegraphics[height=1.05em]{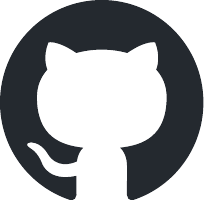}}\xspace}

\newcommand{\eg}{{e.g.}}
\newcommand{\ie}{{i.e.}}

\newcommand{\thinparagraph}[1]{\vspace{0.1em}\noindent\textbf{#1}\enspace}

\title{\center \includegraphics[height=3.7em]{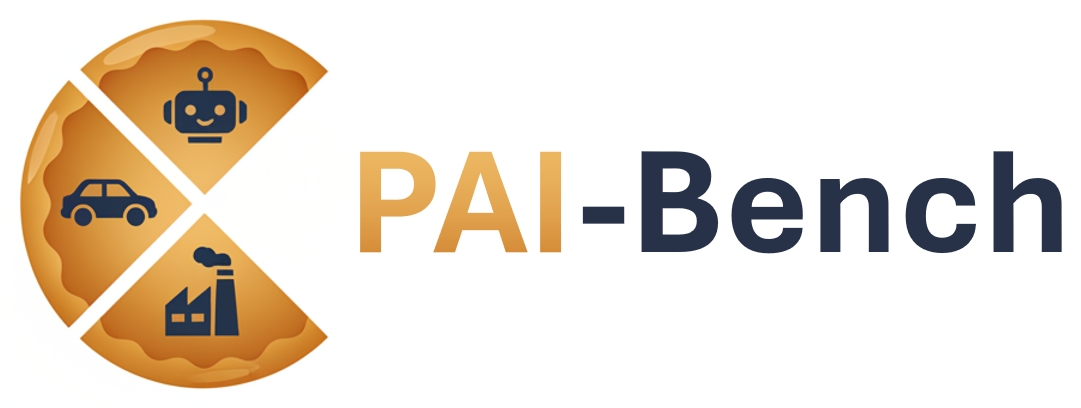} \\  \emph{PAI-Bench}: A Comprehensive Benchmark For Physical AI}

\renewcommand{\thefootnote}{\fnsymbol{footnote}}

\author{
    Fengzhe Zhou\textsuperscript{1,*} \enspace
    Jiannan Huang\textsuperscript{1,*} \enspace
    Jialuo Li\textsuperscript{1,*} \enspace
    Deva Ramanan\textsuperscript{2} \enspace
    Humphrey Shi\textsuperscript{1} \\
    \textsuperscript{1}Georgia Tech \quad
    \textsuperscript{2}CMU \\
}

\begin{abstract}
Physical AI aims to develop models that can perceive and predict real-world dynamics; yet, the extent to which current multi-modal large language models and video generative models support these abilities is insufficiently understood. We introduce Physical AI Bench (PAI-Bench), a unified and comprehensive benchmark that evaluates perception and prediction capabilities across video generation, conditional video generation, and video understanding, comprising 2,808 real-world cases with task-aligned metrics designed to capture physical plausibility and domain-specific reasoning. Our study provides a systematic assessment of recent models and shows that video generative models, despite strong visual fidelity, often struggle to maintain physically coherent dynamics, while multi-modal large language models exhibit limited performance in forecasting and causal interpretation. These observations suggest that current systems are still at an early stage in handling the perceptual and predictive demands of Physical AI. In summary, PAI-Bench establishes a realistic foundation for evaluating Physical AI and highlights key gaps that future systems must address.
\end{abstract}

\projectlinks{
    \begin{tabular}{rl @{\hspace{1.0cm}} l}
        \github & \textbf{Code} & \href{https://github.com/SHI-Labs/physical-ai-bench}{\texttt{SHI-Labs/physical-ai-bench}}\\
        \huggingface & \textbf{Leaderboard} & \href{https://huggingface.co/spaces/shi-labs/physical-ai-bench-leaderboard}{\texttt{shi-labs/physical-ai-bench-leaderboard}}\\
        \huggingface & \textbf{PAI-Bench-G Dataset} & \href{https://huggingface.co/datasets/shi-labs/physical-ai-bench-generation}{\texttt{shi-labs/physical-ai-bench-generation}}\\
        \huggingface & \textbf{PAI-Bench-C Dataset} & \href{https://huggingface.co/datasets/shi-labs/physical-ai-bench-conditional-generation}{\texttt{shi-labs/physical-ai-bench-conditional-generation}}\\
        \huggingface & \textbf{PAI-Bench-U Dataset} & \href{https://huggingface.co/datasets/shi-labs/physical-ai-bench-understanding}{\texttt{shi-labs/physical-ai-bench-understanding}}\\
    \end{tabular}
}

\begin{document}
\maketitle
\footnotetext[1]{Equal contribution}
\setcounter{footnote}{0}  
\renewcommand{\thefootnote}{\arabic{footnote}}  

\newpage

\section{Introduction}
\label{sec:intro}

\textit{``Human intelligence was not forged in a vacuum. It was sculpted by the relentless demands of the physical world."}
\vspace{3pt}

We learned to think because we had to move, grasp, and survive. AI systems stand at a similar threshold. To mature from a mere information processor into a truly capable partner, they must leave the clean, predictable confines of digital simulation and enter the messy, unforgiving reality. This is the core mission of \textbf{Physical AI}:
the embodiment of intelligent algorithms into autonomous systems that must \textit{perceive, predict, and act} directly within the physical world.

In this work, we focus on the \textit{perception} and \textit{prediction} components of this mission through visual signals, specifically video streams. Regarding \textit{perception}, while recent multi-modal large language models (MLLMs) have demonstrated escalating capabilities with increased scale~\cite{liu2023llava, Qwen-VL, tong2024cambrian}, their advancements are predominantly validated using benchmarks for abstract reasoning (e.g., OCR~\cite{fu2025ocrbenchv2improvedbenchmark}, mathematical problems~\cite{song2025videommlumassivemultidisciplinelecture}) and simple daily-life perception~\cite{zhou2025mlvubenchmarkingmultitasklong, fu2025videommefirstevercomprehensiveevaluation, shangguan2025tomatoassessingvisualtemporal}. Their true efficacy within specialized Physical AI applications remains largely uncharted.
Similarly, in the domain of \textit{prediction}, a Physical AI agent must rely on an internal \textit{world model} to forecast the consequences of actions and complex physical dynamics. Video generative models (VGMs)~\cite{openai2025sora, cen2025worldvlaautoregressiveactionworld} represent a powerful new paradigm for learning such dynamics, as they are implicitly trained to understand physics rules by predicting future video frames. However, existing benchmarks~\cite{huang2023vbench,huang2024vbench++,zheng2025vbench2,liu2023evalcrafter} for these models have largely focused on assessing aesthetic appeal and temporal consistency, with far less attention paid to evaluating their fundamental understanding of real-world rules. It thus remains unclear whether they can model complex interactions, adhere to physical laws, or make reasonable predictions in practical applications.

To address these deficiencies and advance Physical AI, we introduce Physical AI Bench~(\textbf{\dataset}), a comprehensive benchmark designed to systematically evaluate model performance on tasks grounded in Physical AI.~\dataset~is structured into three distinct tracks:
\vspace{-6pt}
\begin{itemize}[leftmargin=*]
\item \textbf{PAI-Bench-G} (Generation): Targets \textit{prediction} by evaluating visual quality and physical plausibility of VGMs.
\item \textbf{PAI-Bench-C} (Conditional Generation): Further probes \textit{prediction} by evaluating the fidelity of conditional VGMs, specifically their conformance to input control signals (e.g., depth maps).
\item \textbf{PAI-Bench-U} (Understanding): Targets the \textit{perception} component by evaluating MLLM capabilities on video understanding tasks specific to the Physical AI domain.
\end{itemize}
\vspace{-6pt}
\begin{figure}[t]
    \centering
    \includegraphics[width=0.97\textwidth]{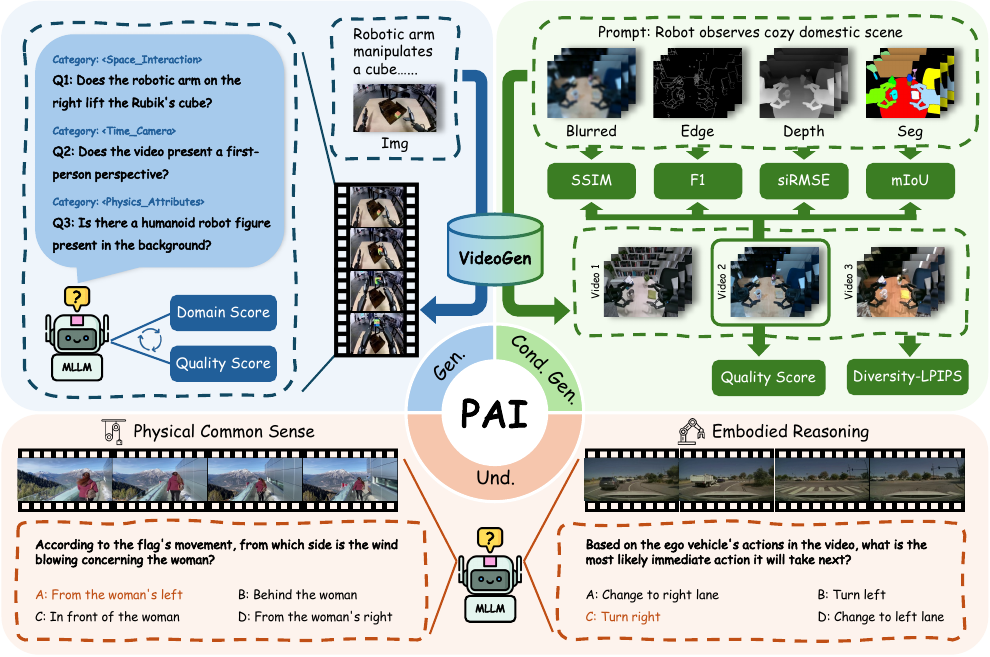}
    \caption{\textbf{Overview of PAI-Bench framework.} PAI-Bench is a comprehensive bench designed for diverse topics in Physical AI, including evaluation for text and condition to physical World Generation, and physical world understanding. }
    \label{fig:frame}
    \vspace{-6pt}
\end{figure}

All three tracks adhere to a unified design principle: grounding evaluation in physically meaningful tasks and real-world data. Specifically, all videos are sourced from real-world captures, such as dashcam recordings. Furthermore,~\dataset~covers several sub-domains, such as autonomous vehicles, robotics, ego-view scenes, \emph{etc.} , across three tracks, comprising 2,808 high-quality cases.

Based on~\dataset, we conducted a comprehensive evaluation of leading VGMs and MLLMs. Extensive experiments reveal that while SOTA VGMs like Veo3~\cite{deepmind2025veo} achieve high visual fidelity, their \textit{prediction} abilities are limited, as they often fail to adhere to basic physical laws or to model complex real-world dynamics. Concurrently, the \textit{perception} capabilities of powerful MLLMs, such as GPT-5~\cite{openai2025gpt5}, lag substantially behind human performance. These findings suggest that the core \textit{prediction} and \textit{perception} capabilities required for Physical AI remain in a nascent stage.

In summary, our contributions are three-fold:
\vspace{-6pt}
\begin{itemize}[leftmargin=*]
    \item We introduce~\dataset, the first benchmark to provide a unified and comprehensive review of video generation, conditional video generation, and video understanding.
    \item As the first high-quality benchmark grounded in Physical AI,~\dataset~spans diverse scenarios (e.g., industrial environments, everyday activities, physical common sense) and incorporates a suite of task-aligned metrics.
    \item We conduct a large-scale evaluation of 15 VGMs, 4 conditional VGMs with 5 control settings, and 16 MLLMs, establishing the current landscape of Physical AI and highlighting key challenges to guide future research.
\end{itemize}
\vspace{-6pt}

\section{Related Work}
\label{sec:rel}

\textit{Prediction} and \textit{Perception} are foundational capabilities for Physical AI applications. However, a significant gap persists in current benchmarking: the absence of a comprehensive framework specifically for the Physical AI domain. Furthermore, existing benchmarks are largely fragmented, typically assessing prediction or perception in isolation.

In the realm of \textit{prediction}, world models~\cite{https://doi.org/10.5281/zenodo.1207631, bar2025navigationworldmodels, liu2025world, cen2025worldvlaautoregressiveactionworld} are increasingly vital for applications such as robotics and autonomous driving. A prominent paradigm utilizes VGMs to forecast high-fidelity future frames~\cite{liu2025world, openai2025sora, deepmind2025veo, predict1, chen2024videocrafter2overcomingdatalimitations,xing2023dynamicrafter, yang2025cogvideoxtexttovideodiffusionmodels,wan2025, hong2022cogvideolargescalepretrainingtexttovideo, deepmind2025veo, wiedemer2025videomodelszeroshotlearners, openai2025sora, openai2024video, yang2025cogvideoxtexttovideodiffusionmodels}. This progress has spurred the proposal of numerous benchmarks~\cite{feng2024tcbenchbenchmarkingtemporalcompositionality, Ji_2024_CVPR, liu2023evalcrafter, liu2023fetvbenchmarkfinegrainedevaluation, meng2024worldsimulatorcraftingphysical,yuan2024chronomagicbenchbenchmarkmetamorphicevaluation}. However, existing evaluation methodologies are often narrow in scope. Some benchmarks~\cite{huang2023vbench, zheng2025vbench2,huang2024vbench++} primarily assess visual quality and temporal consistency, while others~\cite{meng2024worldsimulatorcraftingphysical, bordes2025intphys2benchmarkingintuitive,bansal2025videophy2challengingactioncentricphysical,duan2025worldscore} concentrate exclusively on physical plausibility. We introduce \textbf{PAI-Bench-G}, which distinguishes itself by integrating both aspects via a domain score and a quality score, with a strong emphasis on Physical AI applications. Furthermore, a significant evaluation gap has emerged as VGMs~\cite{transfer1,liu2025world,wan2025,deepmind2025veo,hong2022cogvideolargescalepretrainingtexttovideo,yang2025cogvideoxtexttovideodiffusionmodels,openai2025sora} increasingly leverage multimodal control signals (e.g., depth maps) for guided synthesis. No existing benchmark systematically tests this controllability. \textbf{PAI-Bench-C} is therefore introduced as the first benchmark to systematically evaluate this condition guided generation capability. 

Similarly, in the \textit{perception} domain, while numerous benchmarks exist for general video understanding~(e.g., action recognition)~\cite{shangguan2025tomatoassessingvisualtemporal, zhou2025mlvubenchmarkingmultitasklong, fu2025videommefirstevercomprehensiveevaluation, wu2024longvideobenchbenchmarklongcontextinterleaved, song2025videommlumassivemultidisciplinelecture,li2024mvbenchcomprehensivemultimodalvideo, hu2025videommmuevaluatingknowledgeacquisition}, evaluations for physically grounded reasoning remain notably limited, particularly for video-based analysis~\cite{foss2025causalvqaphysicallygroundedcausal, geminiroboticsteam2025geminiroboticsbringingai, li2025sciencet2iaddressingscientificillusions, chow2025physbench, motamed2025generative}. \textbf{PAI-Bench-U} bridges this gap by unifying tasks for general video comprehension with a dedicated suite of physically grounded challenges.

\begin{table*}[t]
\centering

\caption{\small \textbf{Comparison with existing benchmarks.} Our PAI-Bench is designed with comprehensive domains in Physical AI. Specifically, we focus on scenarios involving practical applications, such as autonomous vehicles~(AV), industry, embodied AI, and ego-centric views.}
\vspace{-6pt}

\resizebox{0.98\textwidth}{!}{%
\begin{tabular}{l|c|ccc|ccccc} 
\toprule
Benchmarks & \# Examples & \makecell{Video \\ Generation} & \makecell{Conditional Video \\ Generation} & \makecell{Video \\ Understanding}  & \makecell{Embodied \\ AI}  & AV & Industry & \makecell{Egocentric \\ Views} & \makecell{Physical \\ Common Sense}\\
\midrule
EvalCrafter~\cite{liu2023evalcrafter} & 700 &\greencheck & \rederror & \rederror & \rederror & \rederror &\rederror &\rederror &\rederror  \\
VBench~\cite{huang2023vbench} & 800 & \greencheck & \rederror & \rederror & \rederror & \rederror & \rederror  & \rederror &\rederror\\
Physics-IQ~\cite{motamed2025generative} & 198 & \greencheck & \rederror & \rederror & \rederror & \rederror & \rederror & \rederror & \greencheck \\
EgoSchema~\cite{mangalam2023egoschema} & 5000 &  \rederror & \rederror &\greencheck & \rederror & \rederror & \rederror & \greencheck &\rederror\\
VideoMME~\cite{fu2024video} & 2700 & \rederror & \rederror & \greencheck & \rederror & \rederror &  \rederror &\greencheck & \rederror\\
MVP Bench~\cite{krojer2025shortcutawarevideoqabenchmarkphysical} & ~54800 &  \rederror & \rederror & \greencheck & \greencheck & \rederror & \rederror & \greencheck & \greencheck  \\
CausalVQA~\cite{foss2025causalvqaphysicallygroundedcausal} & 793 & \rederror & \rederror & \greencheck & \rederror & \rederror & \rederror & \greencheck & \rederror \\
PhyGenBench~\cite{meng2024worldsimulatorcraftingphysical} & 160 & \greencheck & \rederror & \rederror & \rederror & \rederror & \rederror & \rederror & \greencheck \\
IntPhys2~\cite{bordes2025intphys2benchmarkingintuitive} & 1070 & \greencheck  & \rederror & \greencheck & \greencheck & \rederror &  \rederror & \rederror & \greencheck \\
\midrule
PAI-Bench~(Ours) & 2808 & \greencheck & \greencheck & \greencheck & \greencheck & \greencheck & \greencheck &\greencheck & \greencheck \\
\bottomrule
\end{tabular}
}
\vspace{-6pt}
\label{tab:comparison}
\end{table*}

In summary,~\dataset~is the first comprehensive benchmark to focus specifically on the Physical AI domain. By evaluating the capabilities of \textit{prediction} and \textit{perception}, it provides a holistic framework for assessing the preparedness of AI models for real-world physical interaction.

\section{PAI-Bench: A Comprehensive Benchmark For Physical AI}
\label{sec:method}

\noindent \textbf{Overview.}
This section details the foundational principles and construction process for each track of~\dataset. We elaborate on the core motivation, the design principles, and the data curation process. The overall framework is illustrated in Figure~\ref{fig:frame}. Detailed expositions of PAI-Bench-G, PAI-Bench-C, and PAI-Bench-U are provided in Sections~\ref{sec:paibench-predict}, \ref{sec:paibench-transfer}, and \ref{sec:paibench-reason}, respectively.

\subsection{PAI-Bench-G: Video Generation} \label{sec:paibench-predict}
For \textit{prediction} capability, models should be able to forecast the next physically plausible states based on existing information to dynamically adjust their policies. Within the current community, VGMs are considered the most promising approach for this task. To evaluate the efficacy of these models, we propose two metrics: 1) \textit{Quality Score} to assess the fidelity and coherence of the generated video, and 2) \textit{Domain Score} to validate physical plausibility.

\begin{figure}
    \centering
    \includegraphics[width=0.8\linewidth]{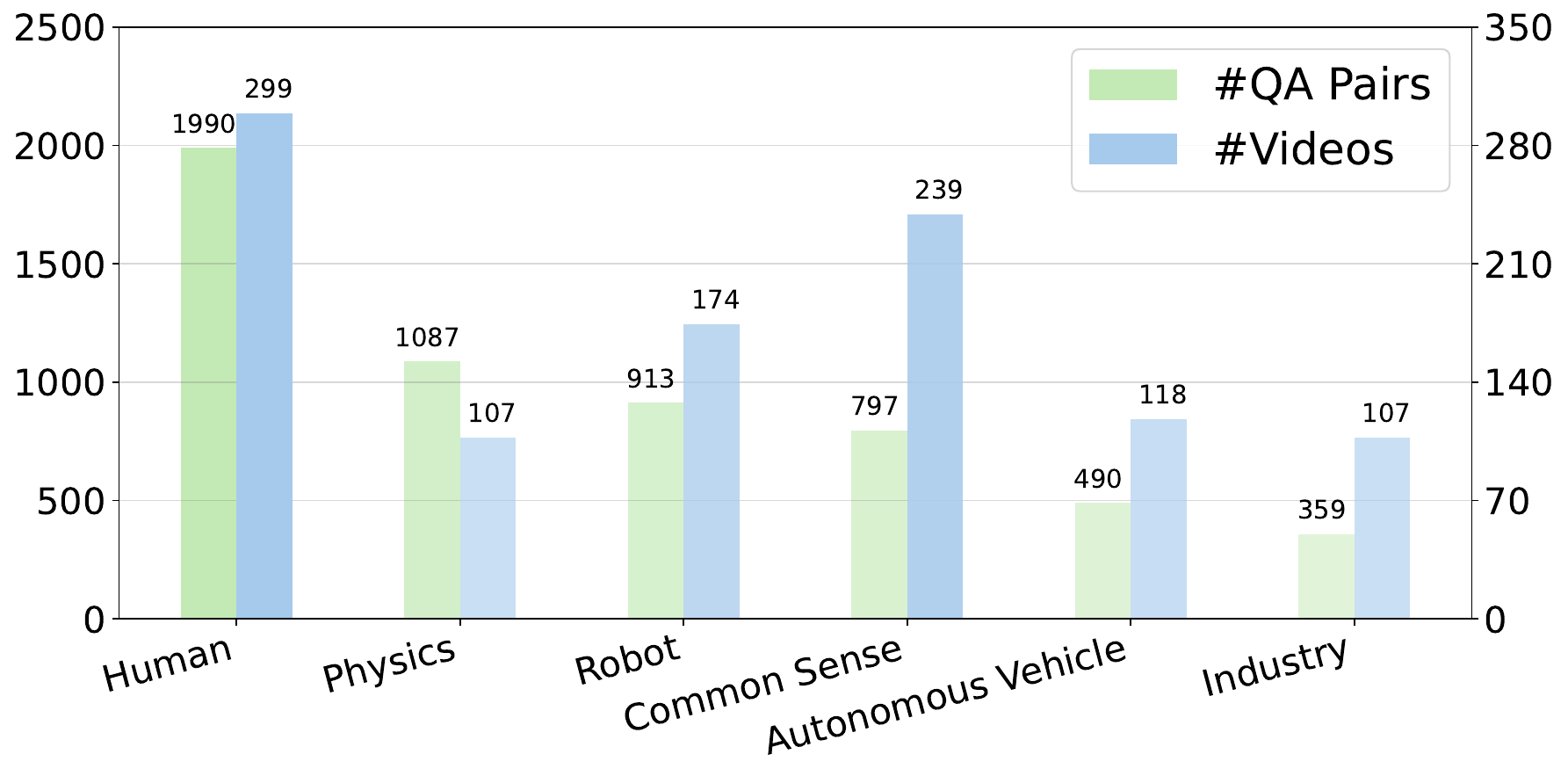}
    \vspace{-8pt}
    \caption{\textbf{Distribution of videos and QA pairs in PAI-Bench-G.} These pairs facilitate \textit{Domain Score} evaluation with an average density of 5-6 QA pairs per video.}
    \vspace{-6pt}
    \label{fig:paibench-g-dist}
\end{figure}

\begin{figure*}
    \centering
    \includegraphics[width=0.98\linewidth]{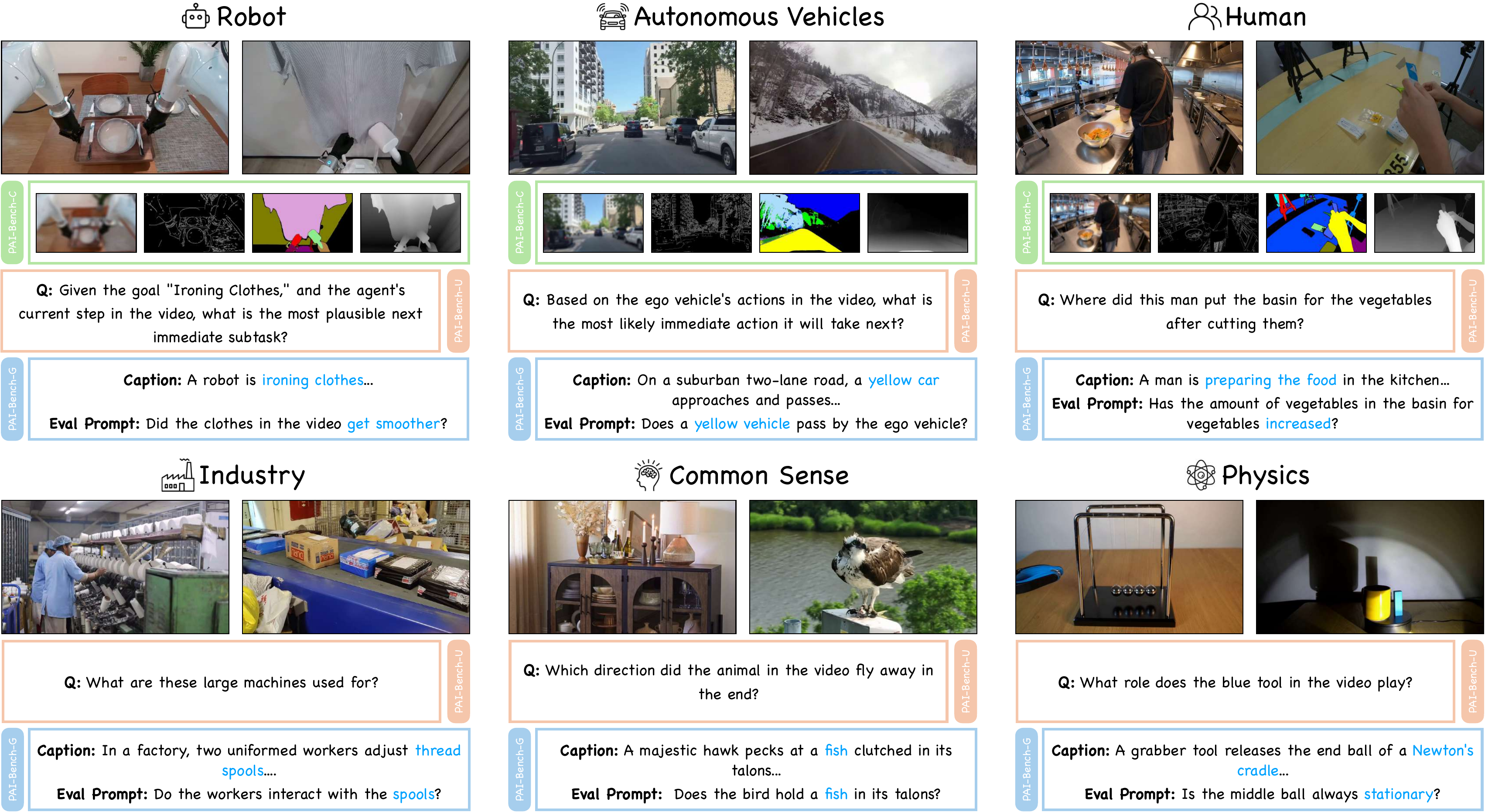}
    \vspace{-3pt}
    \caption{\textbf{Examples of PAI-Bench.} PAI-Bench focuses on Physical AI application scenarios across six domains, where only the first frame of each video is shown for brevity.
    For \textcolor[HTML]{3b7d23}{PAI-Bench-C}, we present the blurred, edge, segmentation, and depth videos that serve as control signals.
    For \textcolor[HTML]{c04f15}{PAI-Bench-U}, we show the questions used for video understanding.
    For \textcolor[HTML]{215f9a}{PAI-Bench-G}, we show the input captions used for generation and a derived prompt used for \textit{Domain Score} evaluation.}
    \vspace{-6pt}
    \label{fig:paibench-case}
\end{figure*}

\thinparagraph{Data Curation.}
We curated videos focusing on Physical AI domains (e.g., autonomous driving, robotics, and industrial applications) from open-source datasets and public web sources (detailed in Appendix~\ref{appen:source}). The annotation involved a two-stage process. First, to generate high-fidelity prompts, we employed an advanced MLLM, \ie, Qwen2.5-VL-72B-Instruct~\cite{bai2025qwen25vltechnicalreport}, for initial video captioning, followed by rigorous manual curation and correction. Second, to support \textit{Domain Score} assessment, we generated QA pairs by leveraging the ontology from~\cite{reason1}. This process used the same MLLM for candidate generation, with subsequent manual refinement ensuring accuracy and relevance. The final dataset comprises 1,044 video-prompt pairs and 5,636 QA pairs across 6 domains. The data distribution and qualitative examples are presented in Figure~\ref{fig:paibench-g-dist} and Figure~\ref{fig:paibench-case}, respectively.

\thinparagraph{Quality Score.} 
To provide a comprehensive assessment of temporal consistency, visual fidelity, and semantic alignment, we adhere to the evaluation protocols established by VBench and VBench++~\cite{zheng2025vbench2,huang2024vbench++,huang2023vbench}. The assessment comprises eight metrics categorized into two groups:
(1) \textit{General Generation Quality}: We evaluate the video's temporal stability (\textit{Subject/Background Consistency}), motion realism (\textit{Motion Smoothness}), and perceptual quality (\textit{Aesthetic/Imaging Quality}) using a combination of frame-level embedding similarities (e.g., DINO~\cite{caron2021emerging}) and specialized predictors (e.g., LAION~\cite{LAIONaes}). Additionally, \textit{Overall Consistency} measures video-text alignment via ViCLIP~\cite{weng2023openvcliptransformingclipopenvocabulary}. (2) \textit{Reference Fidelity}: For image-conditioned tasks, we assess the preservation of the input's identity and layout (\textit{Image-to-Video Subject/Background}) by computing feature distances between the reference image and generated frames. Implementation details are provided in Appendix~\ref{appen:quality_score}.

\thinparagraph{Domain Score.}
Evaluating the physical plausibility of the generated video necessitates analyzing fine-grained spatio-temporal details against foundational world knowledge. Conventional feature-level alignment metrics are inadequate for this task, as they fail to capture high-level semantic reasoning. To address this, we adopt the "MLLM-as-Judge" paradigm, leveraging Qwen3-VL-235B-A22B-Instruct~\cite{qwen_vl_3}. Specifically, We utilize the curated QA pairs, which define the expected physical and semantic content, to query the MLLM against the generated video. The Domain Score is defined as the MLLM's response accuracy across this QA set, quantifying the video's adherence to the specified physical and semantic constraints.

\subsection{PAI-Bench-C: Conditional Video Generation} \label{sec:paibench-transfer}

Control signals provide an interface through which humans can constrain the solution trajectory of VGMs, making them highly valuable in practical applications. In conditional video generation, an ideal outcome should satisfy three key criteria: 1) faithfulness to the control signals, 2) visual quality of the generated videos, and 3) diversity in the generated results under the same configurations of control signals.

\thinparagraph{Evaluation Metrics.}
Based on these principles, we evaluate faithfulness using a suite of fidelity metrics: \textit{Blur SSIM}, \textit{Edge F1}, \textit{Depth si-RMSE}, and \textit{Mask mIoU}. These metrics respectively quantify adherence to blur, edge, depth, and segmentation map controls. The computation projects the generated video into the specific modality space, utilizing Blur Kernel~\cite{1284395}, Canny~\cite{van1979information}, Video Depth Anything~\cite{video_depth_anything}, and GroundingDINO~\cite{liu2023grounding} with SAM2~\cite{ravi2024sam2}, and then measures the similarity to the ground truth control signal. Visual quality is measured using DOVER~\cite{wu2022fastvqa,wu2023dover,end2endvideoqualitytool}, and diversity is assessed with LPIPS~\cite{zhang2018perceptual}. Detailed calculation protocols are presented in Appendix~\ref{appen:pai-c-score}.

\thinparagraph{Dataset Curation.} We curated 600 videos by sampling 200 clips from each of the three datasets: AgiBot~\cite{bu2025agibot,shi2025diversity} for robotics, OpenDV~\cite{yang2024genad} for autonomous driving, and Ego-Exo-4D~\cite{grauman2024egoexo4dunderstandingskilledhuman} for ego-centric tasks. Ground-truth control signals were extracted using the modality-specific models described above. For diversity evaluation, we adopted a human-in-the-loop pipeline similar to PAI-Bench-G: first, Qwen2.5-VL-72B-Instruct~\cite{bai2025qwen25vltechnicalreport} generated a base high-fidelity caption; second, the MLLM iteratively modified visually dominant objects to create novel scenes. Both stages included manual refinement to ensure caption coherence, producing one original and five variant captions per video.

\subsection{PAI-Bench-U: Physical Video Understanding}
\label{sec:paibench-reason}

Human cognition leverages multimodal sensory integration to build robust representations of the physical world. This \textit{perception} capability is also foundational for Physical AI. Focusing on the interpretation of visual signals from videos, we posit that robust Physical AI models must possess two critical capabilities: 1) \textit{Physical Common Sense Reasoning} and embodiment-agnostic understanding of environmental physics to assess real world plausibility, and 2) \textit{Embodied Reasoning}, the capacity for an agent to perceive, reason about, and plan future interactions.

\thinparagraph{Physical Common Sense Reasoning.}
Humans acquire physical common sense primarily through passive observation~\cite{riochet2018intphys}, forming an implicit knowledge base of real-world plausibility. To formalize this concept, we delineate an ontology comprising three domains:
\vspace{-6pt}
\begin{itemize}[leftmargin=*]
    \item \textit{Space} governs object relations, interactions, and environmental context. This includes determining spatial feasibility and understanding scene composition.
    \item \textit{Time} concerns actions and events unfolding over a duration. This includes understanding event timestamps, sequential order, and causality.
    \item \textit{Physical World} addresses physical principles and intrinsic object properties. It includes understanding object states and identifying situations that defy physical laws.
\end{itemize}
\vspace{-6pt}
\thinparagraph{Embodied Reasoning.} Unlike abstract symbolic logic, embodied reasoning is grounded in the real-world actions of an agent operating in dynamic environments. This active grounding enables the agent to transcend passive understanding and develop predictive plans for future interactions. We delineate this capability into two components:
\vspace{-6pt}
\begin{itemize}[leftmargin=*]
    \item \textit{Predicting Action Effects} involves reasoning about physical cause-and-effect to forecast the consequences of an agent's interactions. We assess this via two tasks: task-completion verification (determining if a goal state is achieved) and next-plausible action prediction (forecasting subsequent steps toward a goal).
    \item \textit{Adherence to Physical Constraints} applies real-world physical principles to generate action plans that are feasible, stable, and safe. We assess this via action affordance, which evaluates the viability of performing a specific action to achieve a goal.
\end{itemize}
\vspace{-6pt}

\begin{figure*}[t]
    \centering

    \begin{minipage}[t]{0.3\linewidth} %
        \centering
        \begin{figure}[H]
            \includegraphics[width=\linewidth]{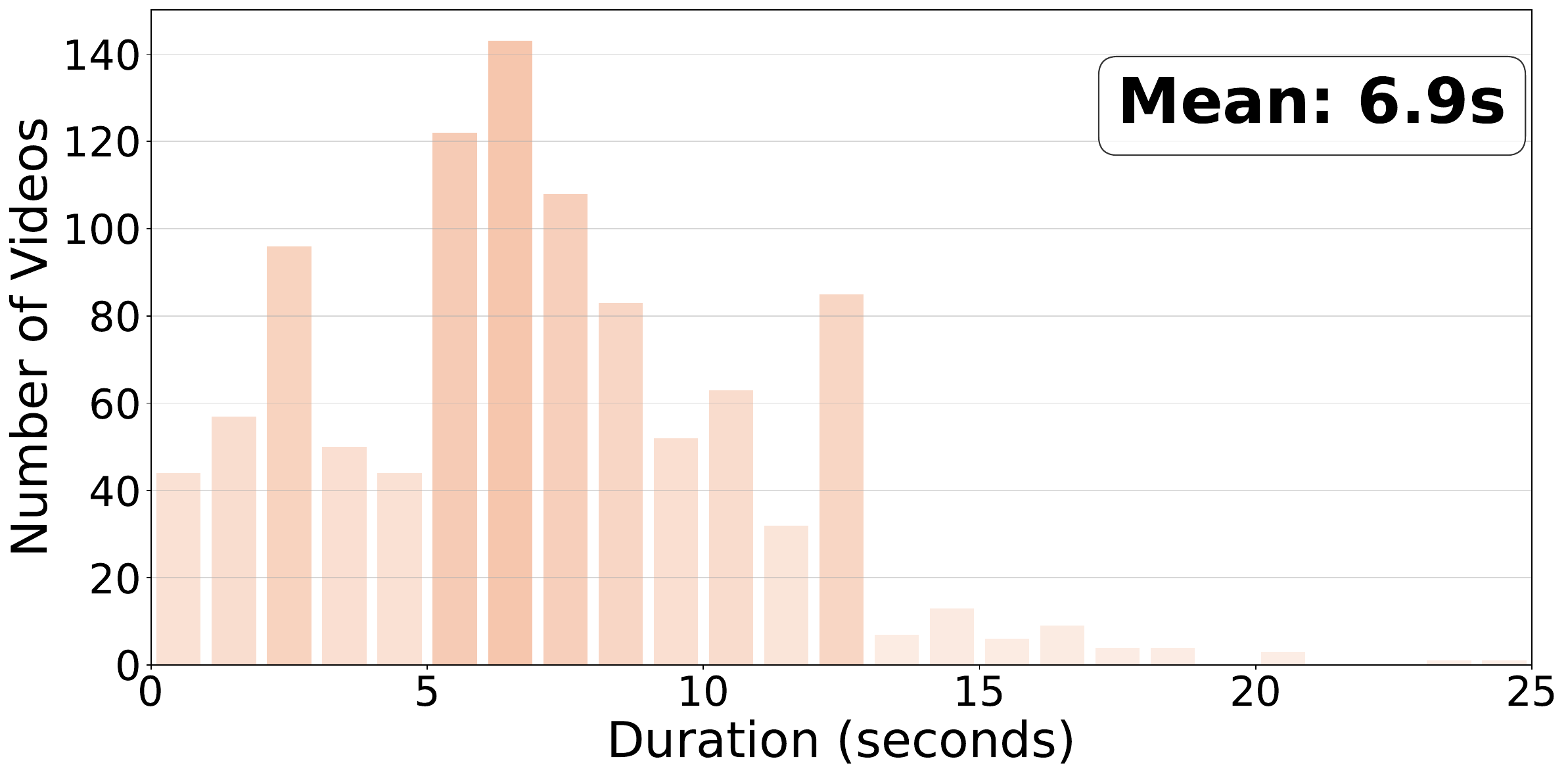}
            \caption{\small Video duration distribution in PAI-Bench-U.}
            \label{fig:paibench-u-duration}
        \end{figure}
    \end{minipage}
    \hfill
    \begin{minipage}[t]{0.29\linewidth}
        \centering
        \begin{figure}[H]
            \includegraphics[width=\linewidth]{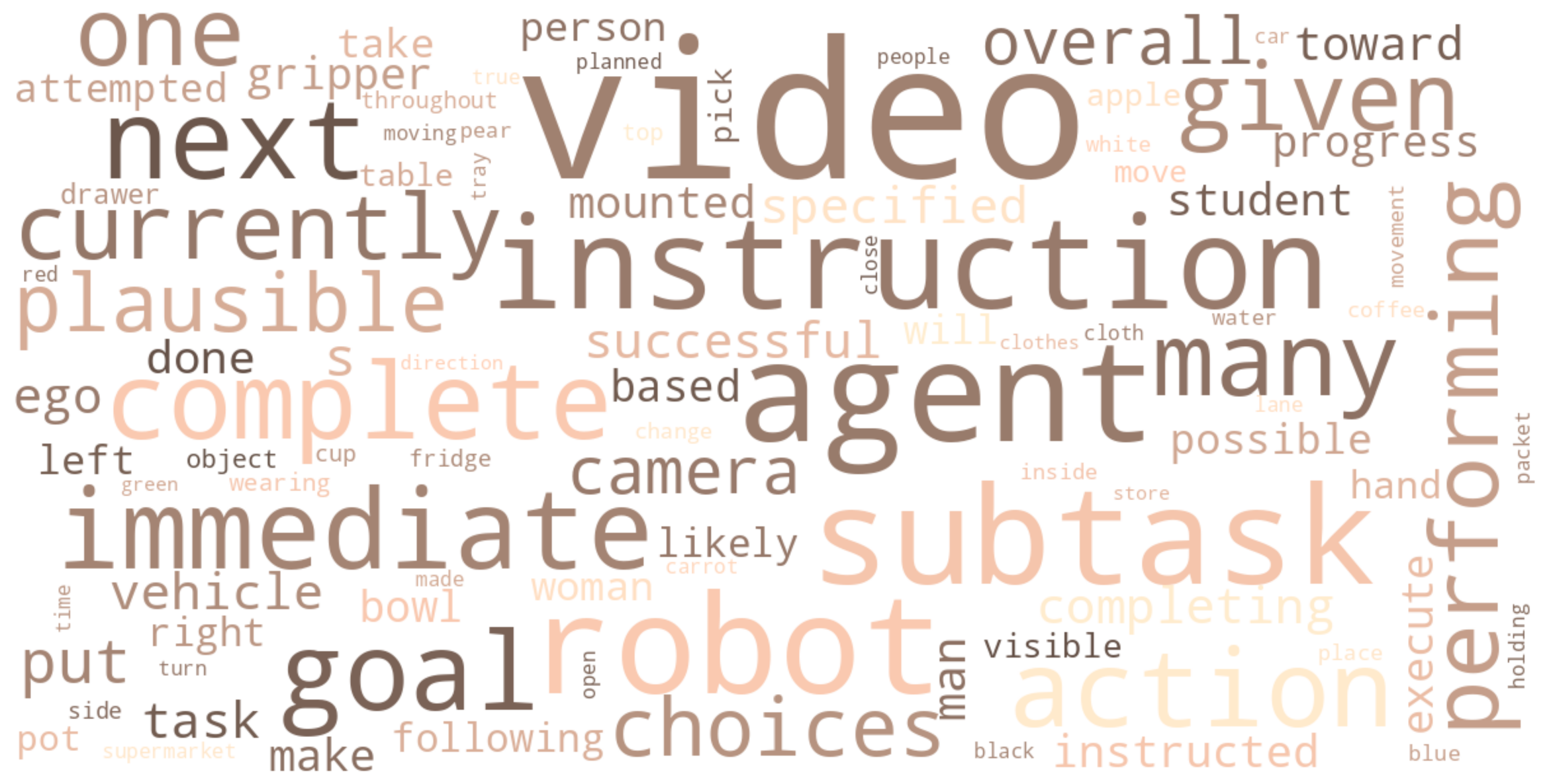}
            \caption{\small Word cloud of questions in PAI-Bench-U.}
            \label{fig:paibench-u-wordcloud}
        \end{figure}
    \end{minipage}
    \hfill
    \begin{minipage}[t]{0.38\linewidth}
        \centering
        \begin{table}[H]
            \centering
            \caption{\small Question counts across categories in PAI-Bench-U.}
            \label{tab:paibench-u-ques}
            \resizebox{0.5\linewidth}{!}{
            \begin{tabular}{ccc}
                \toprule
                Space & Time & Physical World \\
                \midrule
                 80 & 298 & 226 \\
                \bottomrule
            \end{tabular}
            \vspace{30pt}
            }
            \\[5mm]
            \resizebox{0.98\linewidth}{!}{
            \begin{tabular}{cccccc}
                \toprule
                 BridgeData & RoboVQA & RoboFail & Agibot & HoloAssist & AV \\
                \midrule
                 100 & 101 & 100 & 100 & 100 & 100 \\
                \bottomrule
            \end{tabular}
            }
        \end{table}
    \end{minipage}%
    \vspace{-12pt}
\end{figure*}

\thinparagraph{Data Curation.}
For physical common sense reasoning, we first collected over 1,000 videos from online sources and manually annotated 5,737 questions based on the defined ontology. Following a rigorous manual review process, this collection was refined to a final set of 604 high-quality QA pairs across 426 videos. For embodied reasoning, we sourced 601 videos from diverse existing datasets, including RoboVQA~\cite{sermanet2023robovqamultimodallonghorizonreasoning}, RoboFail~\cite{liu2023reflectsummarizingrobotexperiences}, BridgeData~\cite{walke2024bridgedatav2datasetrobot}, AgiBot~\cite{bu2025agibot}, HoloAssist~\cite{HoloAssist2023}, and a proprietary AV dataset. Guided by the established criteria for this domain, 610 QA pairs were manually annotated for this subset. Comprehensive meta-data and statistical distributions are detailed in Figures~\ref{fig:paibench-u-duration},~\ref{fig:paibench-u-wordcloud} and Table~\ref{tab:paibench-u-ques}. Full curation details are in Appendix~\ref{sec:appendix-reason}.

\section{Experiments}
\label{sec:exp}

\thinparagraph{Overview.} We conduct a systematic evaluation of current Physical AI capabilities through PAI-Bench suite. Our experiments cover three distinct aspects: video generation fidelity and plausibility on \textbf{PAI-Bench-G} (\S\ref{subsec:exp-g}), the efficacy of multi-signal control on \textbf{PAI-Bench-C} (\S\ref{subsec:exp-c}), and the physical reasoning capability of MLLMs on \textbf{PAI-Bench-U} (\S\ref{subsec:exp-u}). In the following sections, we outline the model configurations and evaluation protocols, followed by an in-depth discussion of the quantitative results and observed phenomena.

\subsection{PAI-Bench-G} \label{subsec:exp-g}

\thinparagraph{Models.} We evaluate 15 leading VGMs, covering open-source models from Wan~\cite{wan2025}, Cosmos-Predict~\cite{liu2025world}, MAGI~\cite{ai2025magi1autoregressivevideogeneration}, and CogVideoX~\cite{yang2025cogvideoxtexttovideodiffusionmodels} families, along with DynamicCrafter~\cite{xing2023dynamicrafter}, LTX-Video~\cite{HaCohen2024LTXVideo}, HunyuanVideo-I2V~\cite{kong2024hunyuanvideo}, and the proprietary model Veo3~\cite{deepmind2025veo}.

\thinparagraph{Evaluation Suites.} We evaluate all models using their default settings (\eg, resolution, frame count, and frame rate). To account for stochasticity, scores for open-source models are averaged over five videos generated per caption, each with different random seeds. We generated one video per caption for Veo3~\cite{deepmind2025veo}. The complete results are presented in Table~\ref{tab:predict_results}.

\begin{figure*}[!t]
    \centering
    \includegraphics[width=0.32\textwidth]{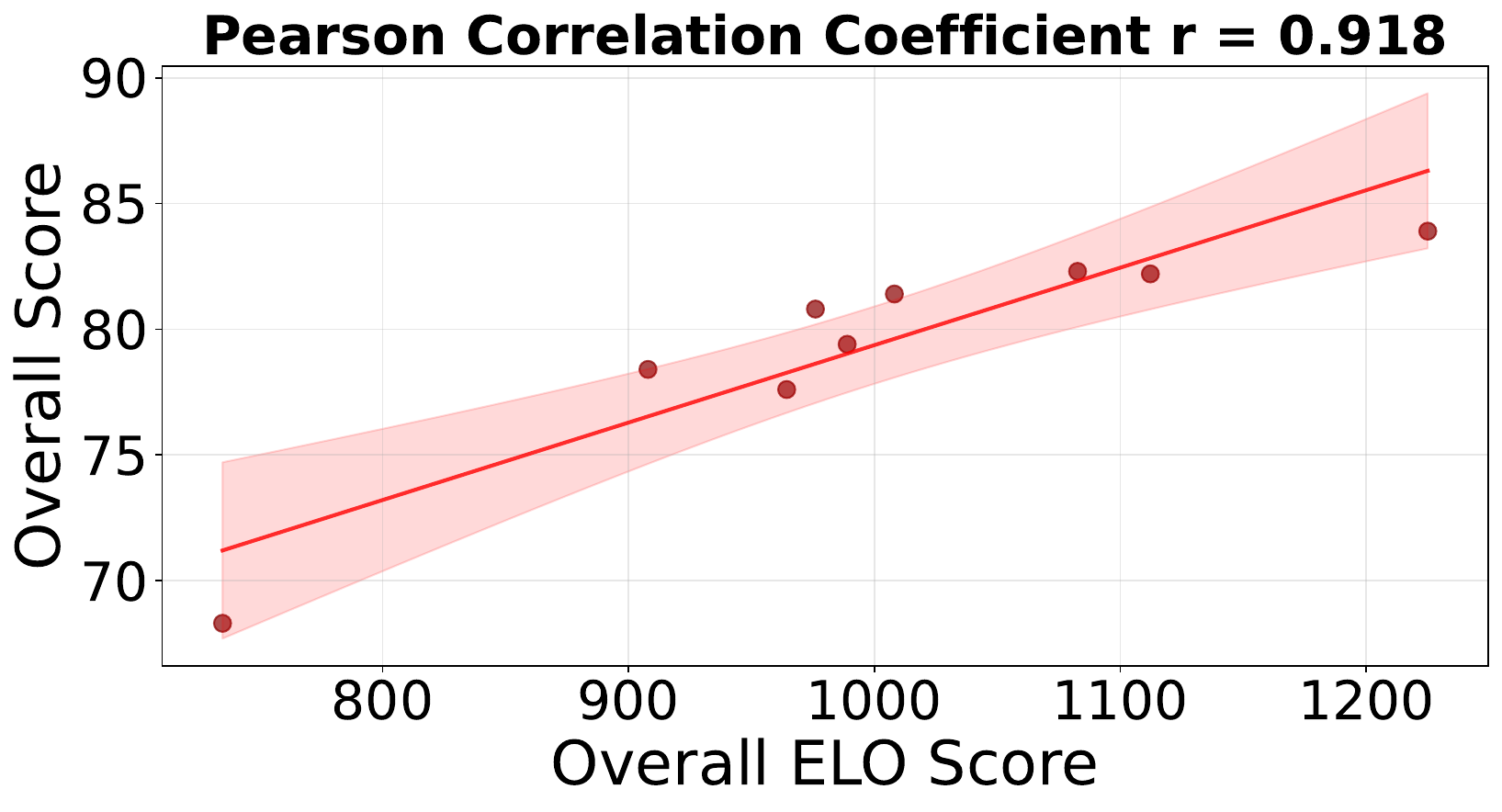}
    \hfill
    \includegraphics[width=0.32\textwidth]{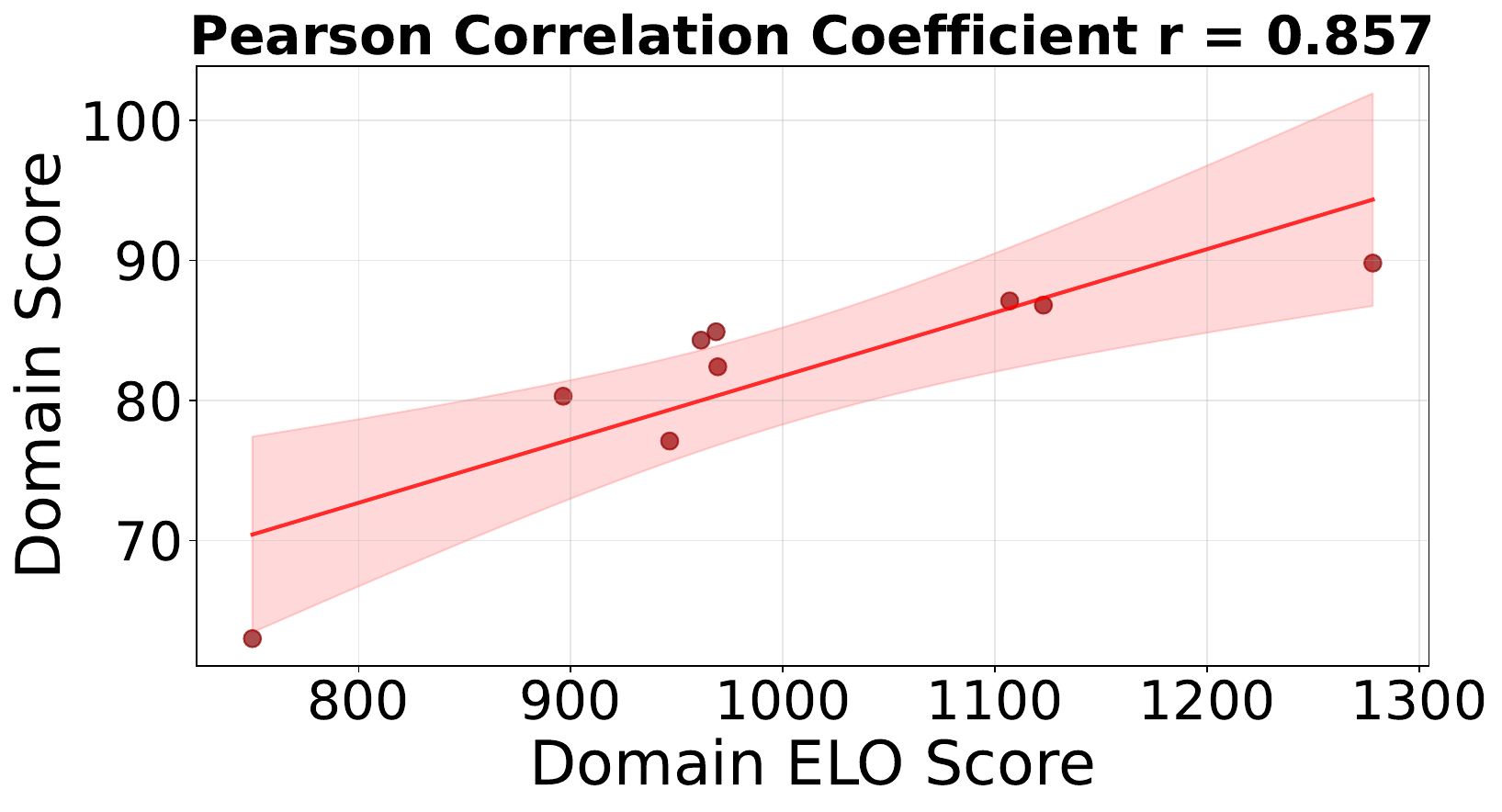}
    \hfill
    \includegraphics[width=0.32\textwidth]{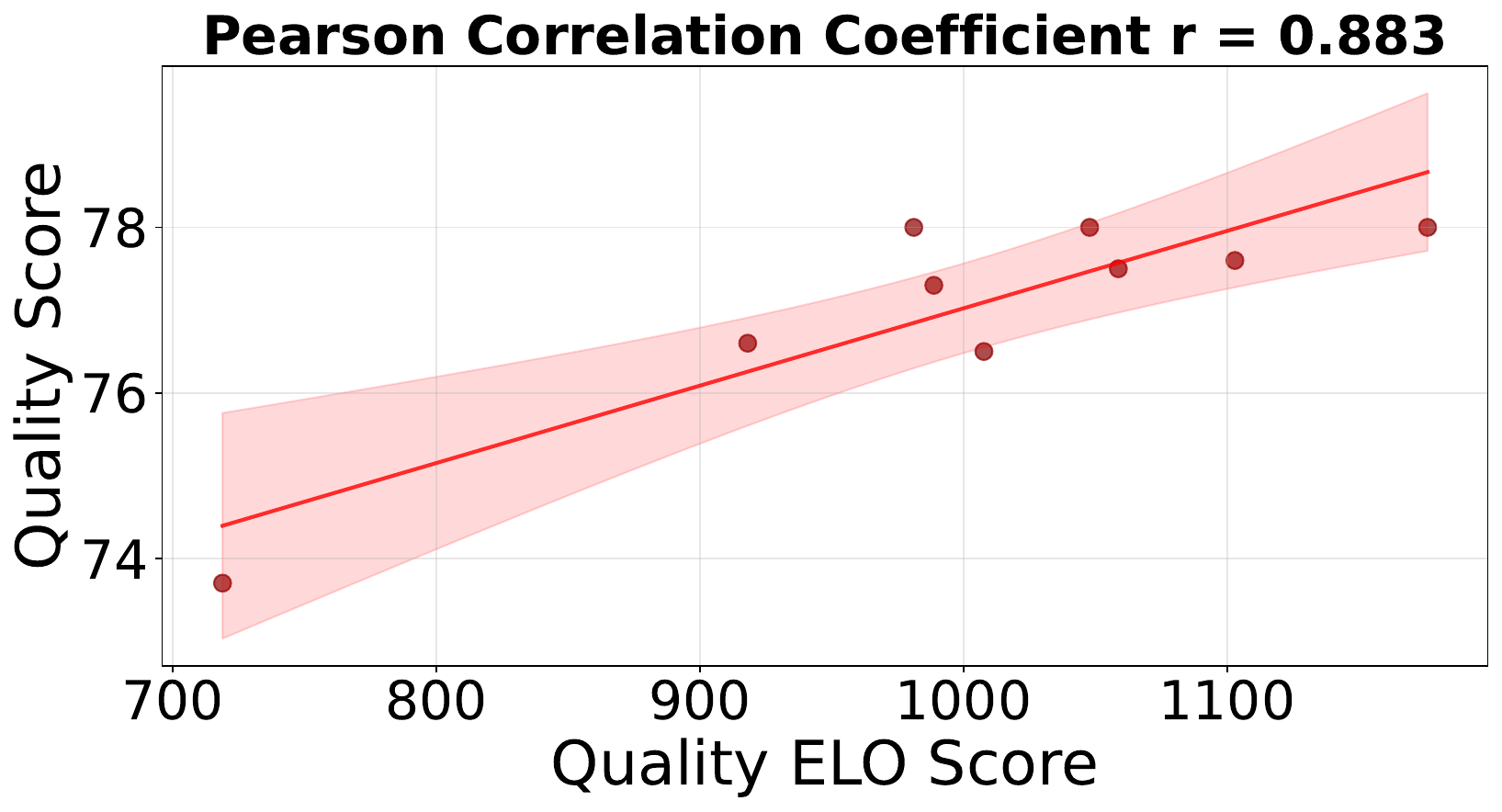}
    \vspace{-6pt}
    \caption{\textbf{Pearson correlation analysis on PAI-Bench-G.} The red shaded regions indicate the 0.95 confidence intervals.}
    \vspace{-6pt}
    \label{fig:pearson}
\end{figure*}

\thinparagraph{Metrics in PAI-Bench-G align with human preferences.}
To validate our proposed metrics against human preferences, we conduct an arena-based human study. We sample video pairs generated by various VGMs from identical captions. Participants then perform pairwise comparisons, selecting the superior video or noting a tie based on two separate criteria: video quality, corresponding to \textit{Quality Score}, and physical plausibility, corresponding to \textit{Domain Score}. We aggregate the pairwise preferences into ELO scores~\cite{elo1967proposed} and compute the Pearson correlation between these ELO scores and our metric scores. As shown in Figure~\ref{fig:pearson}, the results demonstrate a strong alignment, yielding an overall Pearson correlation coefficient of $\mathbf{r=0.918}$.

\begin{table*}[t]
  \centering
  \caption{
    \small \textbf{Evaluation results of 15 VGMs on PAI-Bench-G.}
    Metrics in Domain Score: Common Sense~(CS), Autonomous Vehicle~(AV), Robot~(RO), Industry~(IN), Human~(HU), Physics~(PH);
    and Quality Score: Subject Consistency~(SC), Background Consistency~(BC), Motion Smoothness~(MS), Aesthetic Quality~(AQ), Imaging Quality~(IQ), Overall Consistency~(OC), I2V Subject~(IS), I2V Background~(IB).
    \colorbox{lightblue!25}{Blue} means the best across open-source models.
  }
  \vspace{-4pt}
  \resizebox{\textwidth}{!}{%
  \small
  \begin{tabular}{@{}l|c|ccccccc|ccccccccc@{}}
  \toprule
  \multirow{2}{*}{\; Models} &
  \multirow{2}{*}{Overall} &
  \multicolumn{7}{c|}{Domain Score} &
  \multicolumn{9}{c}{Quality Score} \\
  \cmidrule(lr){3-9} \cmidrule(lr){10-18}
  & & CS & AV & RO & IN & HU & PH & Avg. & SC & BC & MS & AQ & IQ & OC & IS & IB & Avg. \\
  \midrule

\; Source Videos & 83.9 & 96.6 & 78.7 & 91.8 & 89.0 & 86.6 & 93.1 & 89.8 & 93.3 & 94.2 & 99.1 & 51.7 & 68.4 & 21.5 & 97.8 & 98.2 & 78.0 \\
  \midrule
  \multicolumn{18}{l}{\cellcolor{gray!10}\textit{Proprietary Models}} \\
\; Veo3~\cite{deepmind2025veo} & 82.2 & 93.0 & 72.1 & 88.8 & 89.0 & 84.4 & 89.7 & 86.8 & 91.4 & 93.1 & 99.2 & 51.9 & 69.8 & 21.7 & 97.0 & 96.9 & 77.6 \\
  \midrule
  \multicolumn{18}{l}{\cellcolor{gray!10}\textit{Open-source Models}} \\

\; DynamicCrafter~\cite{xing2023dynamicrafter} & 68.3 & 73.9 & 43.9 & 48.9 & 70.9 & 60.7 & 80.8 & 63.0 & 91.1 & 92.5 & 94.9 & 51.5 & 68.0 & 21.2 & 84.5 & 86.2 & 73.7 \\

\; LTX-Video-2B~\cite{HaCohen2024LTXVideo} & 77.5 & 88.1 & 60.5 & 69.4 & 82.6 & 77.1 & 86.2 & 78.0 & 89.2 & 92.7 & 98.7 & 53.2 & \cellcolor{lightblue!25}71.3 & 21.1 & 95.0 & 95.9 & 77.1 \\

\; MAGI-1-4.5B~\cite{ai2025magi1autoregressivevideogeneration} & 77.5 & 88.3 & 63.4 & 71.3 & 82.5 & 77.3 & 86.3 & 78.7 & 92.1 & 93.3 & 99.0 & 50.4 & 61.8 & \cellcolor{lightblue!25}21.6 & 94.5 & \cellcolor{lightblue!25}98.1 & 76.3 \\

\; HunyuanVideo-I2V~\cite{kong2024hunyuanvideo} & 77.6 & 86.7 & 60.3 & 68.9 & 82.4 & 75.8 & 85.8 & 77.1 & \cellcolor{lightblue!25}94.3 & \cellcolor{lightblue!25}95.3 & \cellcolor{lightblue!25}99.5 & 52.1 & 65.2 & 21.5 & \cellcolor{lightblue!25}98.6 & 97.6 & \cellcolor{lightblue!25}78.0 \\

\; CogVideoX1.5~\cite{yang2025cogvideoxtexttovideodiffusionmodels} & 78.4 & 88.7 & 64.4 & 73.6 & 84.6 & 79.1 & 88.7 & 80.3 & 91.6 & 93.9 & 98.5 & 50.0 & 66.5 & 21.2 & 95.0 & 96.1 & 76.6 \\

\; LTX-Video-13B~\cite{HaCohen2024LTXVideo} & 78.4 & 88.6 & 60.0 & 71.8 & 84.6 & 79.1 & 88.8 & 79.5 & 90.6 & 93.5 & 99.0 & \cellcolor{lightblue!25}53.5 & 69.5 & 21.4 & 95.7 & 96.0 & 77.4 \\

\; CogVideoX~\cite{yang2025cogvideoxtexttovideodiffusionmodels} & 78.6 & 88.4 & 62.7 & 75.3 & 85.0 & 81.0 & 88.5 & 80.9 & 91.4 & 93.4 & 98.0 & 51.2 & 64.6 & 21.3 & 94.1 & 95.9 & 76.3 \\

\; MAGI-1-24B~\cite{ai2025magi1autoregressivevideogeneration} & 79.4 & 91.4 & 68.0 & 76.5 & 85.4 & 81.5 & 87.0 & 82.4 & 90.0 & 92.4 & 99.0 & 50.2 & 64.2 & 21.4 & 96.8 & 97.9 & 76.5 \\

\; Cosmos-Predict2-2B~\cite{liu2025world} & 79.7 & 91.2 & 70.7 & 81.9 & 86.4 & 83.1 & 88.1 & 84.2 & 88.7 & 92.1 & 97.6 & 49.3 & 65.9 & \cellcolor{lightblue!25}21.6 & 91.9 & 94.6 & 75.2 \\

\; Cosmos-Predict2-14B~\cite{liu2025world} & 80.0 & 92.5 & 69.8 & 81.3 & 86.6 & 82.4 & 89.2 & 84.3 & 89.6 & 92.8 & 98.0 & 49.8 & 67.5 & 21.5 & 92.2 & 94.9 & 75.8 \\

\; Wan2.1-I2V-14B~\cite{wan2025} & 80.8 & 92.5 & 69.4 & 79.6 & 87.2 & 83.4 & 89.5 & 84.3 & 90.4 & 93.0 & 98.0 & 51.7 & 70.0 & 21.4 & 96.5 & 97.2 & 77.3 \\

\; Wan2.2-TI2V-5B~\cite{wan2025} & 81.3 & 92.2 & 69.1 & 82.9 & 86.4 & 83.3 & 90.3 & 84.7 & 92.0 & 93.9 & 98.8 & 52.3 & 69.6 & 21.4 & 97.4 & 98.0 & 77.9 \\

\; Cosmos-Predict2.5-2B~\cite{liu2025world} & 81.4 & 91.5 & 70.4 & 82.7 & 86.8 & 83.5 & 91.7 & 84.9 & 92.3 & 94.2 & 99.1 & 52.8 & 70.7 & 21.2 & 96.2 & 97.2 & \cellcolor{lightblue!25}78.0 \\

\; Wan2.2-I2V-A14B~\cite{wan2025} & \cellcolor{lightblue!25}82.3 & \cellcolor{lightblue!25}94.1 & \cellcolor{lightblue!25}73.0 & \cellcolor{lightblue!25}86.8 & \cellcolor{lightblue!25}88.6 & \cellcolor{lightblue!25}84.5 & \cellcolor{lightblue!25}92.7 & \cellcolor{lightblue!25}87.1 & 91.2 & 93.2 & 98.3 & 51.8 & 69.2 & 21.5 & 97.2 & 97.8 & 77.5 \\

  \bottomrule
  \end{tabular} %
  }
  \label{tab:predict_results}
  \vspace{-3pt}
\end{table*}

\thinparagraph{VGMs excel in visual quality but lack physical plausibility.} As detailed in Table~\ref{tab:predict_results}, a clear trend emerges: most leading VGMs achieve Quality Score that is highly competitive with, and in some cases identical to, that of the source videos (78.0). This suggests that current models have largely succeeded in generating videos with high visual fidelity and aesthetic appeal. In stark contrast, a significant disparity exists in Domain Score. The source videos, being real-world captures and thus inherently physically coherent, set a high baseline. However, all evaluated models fall short of this benchmark. This discrepancy highlights that while VGMs excel at visual synthesis, they still struggle to generate content that consistently adheres to fundamental physical laws. Consequently, enhancing the physical plausibility of generated videos remains a critical challenge and a key direction for Physical AI.

\subsection{PAI-Bench-C} \label{subsec:exp-c}

\thinparagraph{Models.} Our evaluation targets two families of controllable video generation models: Cosmos-Transfer~\cite{transfer1,liu2025world} and Wan-Fun~\cite{wan22fun}.

\thinparagraph{Evaluation Suites.}
We standardize generation settings across all models, configuring them to produce 121-frame videos, preserve the input video's aspect ratio, and use their default configurations. We assess model performance using five distinct conditioning signals: blurred video (\textit{Blur}), edge video (\textit{Edge}), depth video (\textit{Depth}), segmentation maps (\textit{Seg}), and a combination of all signals with equal weighting (\textit{All}). The \textit{All} condition is exclusively evaluated on Cosmos-Transfer~\cite{transfer1}. For each control signal, we generate six videos using six unique captions, maintaining a fixed random seed. Notably, since Wan2.2-Fun-5B-Control~\cite{wan22fun} failed to produce coherent results when conditioned on blur or segmentation maps, these are omitted from our evaluation. All results are summarized in Table~\ref{tab:transfer_results}.

\thinparagraph{Multi-signal conditioning enhances video quality.} The results in Table~\ref{tab:transfer_results} clearly demonstrate the benefits of multi-signal conditioning. For both Cosmos-Transfer~\cite{transfer1} models, the \textit{All} condition achieves the highest Quality Score, significantly outperforming any single control signal. This finding suggests a valuable practical application: instead of relying on a single degraded input, such as a blurry or noisy video, users can first extract complementary control signals from the source. By providing the model with this comprehensive set of control signals, it becomes possible to reconstruct a video of substantially higher quality.

\thinparagraph{Segmentation signals induce poor segmentation fidelity.} For Cosmos-Transfer~\cite{transfer1} models, surprisingly, we observe that using segmentation maps as the control signal results in the lowest Mask mIoU. We hypothesize that this stems from segmentation maps being the noisiest supervision signal among all conditions. Even state-of-the-art segmentation models, \eg, SAM2~\cite{ravi2024sam2}, can produce masks with lower temporal consistency than other control signals, such as the occasional missing object masks across frames. Consequently, the resulting training supervision is likely to be highly unreliable, degrading the model's ability.

\begin{table*}[t]
\centering
\caption{
    \small \textbf{Evaluation results of 4 conditional VGMs on PAI-Bench-C.}
    For each model, the control signal settings consist of either a single video or a combination of multiple signal videos.
    \colorbox{lightgreen!25}{Green} means the best across control signal settings for each model.
}
\vspace{-6pt}

\resizebox{0.97\textwidth}{!}{%
\begin{tabular}{l|c|cccccc}
\toprule
Model & Condition & Blur SSIM $\uparrow$ & Edge F1 $\uparrow$ & Depth si-RMSE $\downarrow$ & Mask mIoU $\uparrow$ & Quality Score $\uparrow$ & Diversity $\uparrow$ \\
\midrule

\multirow{5}{*}{Cosmos-Transfer1-7B~\cite{transfer1}}
 & Blur  & \cellcolor{lightgreen!25}0.89 & 0.20 & \cellcolor{lightgreen!25}0.66 & 0.73 & 6.56 & 0.19 \\
 & Edge & 0.77 & \cellcolor{lightgreen!25}0.38 & 0.85 & 0.73 & 6.76 & 0.28 \\
 & Depth & 0.67 & 0.15 & 0.76 & 0.71 & 6.89 & 0.39 \\
 & Seg & 0.62 & 0.11 & 1.13 & 0.70 & 6.02 & \cellcolor{lightgreen!25}0.42 \\
 & All & 0.82 & 0.26 & 0.70 & \cellcolor{lightgreen!25}0.74 & \cellcolor{lightgreen!25}9.24 & 0.22 \\
\midrule

\multirow{5}{*}{Cosmos-Transfer2.5-2B~\cite{liu2025world}}
 & Blur & \cellcolor{lightgreen!25}0.91 & 0.26 & \cellcolor{lightgreen!25}0.54 & 0.75 & 8.77 & 0.18 \\
 & Edge & 0.76 & 0.39 & 0.74 & 0.74 & 8.05 & 0.36 \\
 & Depth & 0.70 & 0.17 & 0.83 & 0.72 & 7.30 & 0.41 \\
 & Seg & 0.66 & 0.13 & 1.07 & 0.71 & 7.87 & \cellcolor{lightgreen!25}0.44 \\
 & All & 0.90 & \cellcolor{lightgreen!25}0.45 & 0.59 & \cellcolor{lightgreen!25}0.77 & \cellcolor{lightgreen!25}9.24 & 0.13 \\
\midrule

\multirow{2}{*}{Wan2.2-Fun-5B-Control~\cite{wan22fun}}
 & Edge  & \cellcolor{lightgreen!25}0.61 & \cellcolor{lightgreen!25}0.27 & \cellcolor{lightgreen!25}1.01 & \cellcolor{lightgreen!25}0.71 & 8.79 & 0.40 \\
 & Depth & 0.56 & 0.11 & 1.82 & 0.62 & \cellcolor{lightgreen!25}9.32 & \cellcolor{lightgreen!25}0.48 \\
\midrule

\multirow{4}{*}{Wan2.2-Fun-A14B-Control~\cite{wan22fun}}
 & Blur  & 0.57 & 0.09 & 2.11 & 0.50 & 8.81 & \cellcolor{lightgreen!25}0.53 \\
 & Edge  & \cellcolor{lightgreen!25}0.68 & \cellcolor{lightgreen!25}0.37 & \cellcolor{lightgreen!25}0.84 & \cellcolor{lightgreen!25}0.74 & 9.00 & 0.38 \\
 & Depth & 0.56 & 0.11 & 2.10 & 0.58 & \cellcolor{lightgreen!25}9.22 & 0.52 \\
 & Seg   & 0.47 & 0.10 & 1.60 & 0.66 & 7.79 & 0.36 \\
\bottomrule
\end{tabular}}
\vspace{-6pt}
\label{tab:transfer_results}
\end{table*}

\subsection{PAI-Bench-U} \label{subsec:exp-u}

\thinparagraph{Models.} We evaluate 21 MLLMs, covering both proprietary and open-source models. The proprietary models include Claude-3.5-Sonnet~\cite{claude35sonnet}, GPT-4o~\cite{openai2024gpt4ocard}, and GPT-5~\cite{openai2025gpt5}. For GPT-5, we use minimal reasoning effort. The open-source group spans several major families, including Qwen2.5-VL~\cite{bai2025qwen25vltechnicalreport}, Qwen3-VL~\cite{qwen_vl_3}, Cosmos-Reason1~\cite{reason1}, InternVL3.5~\cite{wang2025internvl3_5}, and GLM-4.5V~\cite{vteam2025glm45vglm41vthinkingversatilemultimodal}. This collection covers a broad spectrum of model scales, from 7B to over 200B. To analyze model performance, we also establish a human baseline derived from an annotator team. \\

\thinparagraph{Evaluation Suites.} We conduct our evaluation using the LMMs-Eval framework~\cite{zhang2024lmmsevalrealitycheckevaluation}. All models are evaluated using 16 video frames as input, except for InternVL3.5, which can accommodate only 8 frames due to its context length. We follow the method in~\cite{wang2024qwen2vlenhancingvisionlanguagemodels} to adaptively resize video frames. For all model outputs, we apply the lightweight LLM Qwen3-8B~\cite{qwen3} as a post-processing model to extract the final letter option from the full responses.
\vspace{0.1cm}

\begin{table*}[t]
\caption{\small \textbf{Evaluation of 16 MLLMs on PAI-Bench-U}. Embodied reasoning domains: BridgeData~(BD), RoboVQA~(RV), RoboFail~(RF), Agibot~(AB), HoloAssist~(HA), Autonomous Vehicle~(AV). \colorbox{lightred!40}{Red} denotes the best result across either proprietary or open-source models.}
\vspace{-6pt}
\centering
\small
\setlength{\tabcolsep}{8pt}
\resizebox{1.0\textwidth}{!}{%
\begin{tabular}{@{}l|c|cccc|*{7}{c}@{}}
\toprule
\multirow{2}{*}{\; Models} & \multirow{2}{*}{Overall} &
\multicolumn{4}{c|}{\makecell{Common Sense}} & \multicolumn{7}{c}{\makecell{Embodied Reasoning}} \\
\cmidrule(lr){3-6} \cmidrule(lr){7-13}
& & Space & Time & Physics & Avg. & BD & RV & RF & AB & HA & AV & Avg. \\
\midrule
  \; Human Performance & 93.2 & 92.1 & 94.3 & 95.0 & 93.6 & 94.0 & 95.5 & 97.0 & 90.0 & 89.0 & 100.0 & 88.9 \\
  \; Random Guess & 37.0 & 41.5 & 38.8 & 38.3 & 38.9 & 25.0 & 50.0 & 50.0 & 26.9 & 25.0 & 32.5 & 35.2 \\
\midrule
\multicolumn{13}{l}{\cellcolor{gray!10}\textit{Proprietary Models}} \\
  \; Claude-3.5-Sonnet~\cite{claude35sonnet} & 46.0 & 55.0 & 46.6 & 46.9 & 47.8 & 29.0 & 74.5 & 58.0 & 28.0 & 38.0 & 34.0 & 44.1 \\
  \; GPT-4o~\cite{openai2024gpt4ocard} & 56.2 & 61.2 & 57.0 & 59.7 & 58.6 & 44.0 & 68.2 & 71.0 & \cellcolor{lightred!40}45.0 & 55.0 & 38.0 & 53.8 \\
  \; GPT-5~\cite{openai2025gpt5} & \cellcolor{lightred!40}61.8 & \cellcolor{lightred!40}63.8 & \cellcolor{lightred!40}65.8 & \cellcolor{lightred!40}61.5 & \cellcolor{lightred!40}63.9 & \cellcolor{lightred!40}38.0 & \cellcolor{lightred!40}85.5 & \cellcolor{lightred!40}72.0 & \cellcolor{lightred!40}45.0 & \cellcolor{lightred!40}72.0 & \cellcolor{lightred!40}43.0 & \cellcolor{lightred!40}59.7 \\
\midrule
\multicolumn{13}{l}{\cellcolor{gray!10}\textit{Open-source Models}} \\
  \; InternVL3.5-14B~\cite{wang2025internvl3_5} & 48.8 & 50.0 & 51.3 & 47.3 & 49.7 & 23.0 & 80.0 & 67.0 & 27.0 & 56.0 & 31.0 & 47.9 \\
  \; InternVL3.5-30B-A3B~\cite{wang2025internvl3_5} & 49.4 & 48.8 & 55.7 & 46.0 & 51.2 & 37.0 & 74.5 & 60.0 & 23.0 & 55.0 & 34.0 & 47.7 \\
  \; InternVL3.5-8B~\cite{wang2025internvl3_5} & 49.7 & 50.0 & 55.0 & 44.7 & 50.5 & 29.0 & 75.5 & 63.0 & 39.0 & 49.0 & 35.0 & 48.9 \\
  \; Qwen2.5-VL-7B~\cite{bai2025qwen25vltechnicalreport} & 51.0 & 51.2 & 48.7 & 39.8 & 45.7 & 35.0 & 87.3 & 63.0 & \cellcolor{lightred!40}53.0 & 60.0 & 36.0 & 56.2 \\
  \; Qwen2.5-VL-32B~\cite{bai2025qwen25vltechnicalreport} & 55.3 & 50.0 & 62.1 & 48.7 & 55.5 & 35.0 & \cellcolor{lightred!40}93.6 & 65.0 & 45.0 & 56.0 & 32.0 & 55.1 \\
  \; Cosmos-Reason1-7B~\cite{reason1} & 55.7 & 63.8 & 55.7 & 46.0 & 53.1 & 41.0 & 91.8 & 66.0 & 41.0 & 59.0 & 47.0 & 58.2 \\
  \; InternVL3.5-38B~\cite{wang2025internvl3_5} & 55.8 & 58.8 & 59.7 & 49.6 & 55.8 & 36.0 & 82.7 & 66.0 & 44.0 & 69.0 & 34.0 & 55.7 \\
  \; InternVL3.5-241B-A28B~\cite{wang2025internvl3_5} & 56.3 & 60.0 & 57.4 & 53.5 & 56.3 & 34.0 & 78.2 & 66.0 & 43.0 & 75.0 & 40.0 & 56.4 \\
  \; Qwen3-VL-8B~\cite{qwen_vl_3} & 56.8 & 53.8 & 58.4 & 50.9 & 55.0 & 35.0 & 89.1 & 61.0 & 49.0 & 75.0 & 40.0 & 58.7 \\
  \; Qwen3-VL-30B-A3B~\cite{qwen_vl_3} & 59.5 & 52.5 & 60.4 & 58.4 & 58.6 & 38.0 & 90.9 & 69.0 & 41.0 & 73.0 & 47.0 & 60.3 \\
  \; Qwen2.5-VL-72B~\cite{bai2025qwen25vltechnicalreport} & 60.8 & \cellcolor{lightred!40}65.0 & 57.7 & 57.5 & 58.6 & \cellcolor{lightred!40}50.0 & 91.8 & 68.0 & 52.0 & 70.0 & 43.0 & 63.0 \\
  \; Qwen3-VL-32B~\cite{qwen_vl_3} & 62.0 & 53.8 & 67.8 & 59.7 & 62.9 & 42.0 & 90.9 & \cellcolor{lightred!40}71.0 & 50.0 & 72.0 & 38.0 & 61.1 \\
  \; Qwen3-VL-235B-A22B~\cite{qwen_vl_3} & \cellcolor{lightred!40}64.7 & 56.2 & \cellcolor{lightred!40}69.5 & \cellcolor{lightred!40}61.9 & \cellcolor{lightred!40}64.9 & 42.0 & \cellcolor{lightred!40}93.6 & \cellcolor{lightred!40}71.0 & 45.0 & \cellcolor{lightred!40}76.0 & \cellcolor{lightred!40}56.0 & \cellcolor{lightred!40}64.4 \\
\bottomrule
\end{tabular}%
}
\vspace{-6pt}
\label{tab:reason_results}
\end{table*}

\begin{figure}[t]
    \centering
    \begin{minipage}[b]{0.48\linewidth}
        \makeatletter\def\@captype{table}\makeatother 
        
        \caption{Ablation study on thinking mode across different MLLMs on PAI-Bench-U. Performance gains and degradations are highlighted.}
        \vspace{-6pt}
        \centering
        \resizebox{\linewidth}{!}{%
            \begin{tabular}{@{}l|c|c|c|c@{}}
            \toprule
            {\; Models} & \makecell{Thinking Mode} & {Overall} &
            {\makecell{Common Sense}} & {\makecell{Embodied Reasoning}} \\
            \midrule
            \; \multirow{2}{*}{Qwen3-VL-8B~\cite{qwen_vl_3}} & \rederror & 56.8 & 55.0 & 58.7 \\
            \; & \greencheck & 57.3~\gain{0.5} & 57.0~\gain{2.0} & 57.7~\loss{1.0} \\
            \midrule
            \; \multirow{2}{*}{Qwen3-VL-30B-A3B~\cite{qwen_vl_3}} & \rederror & 59.5 & 58.6 & 60.3 \\
            \; & \greencheck & 57.3~\loss{2.2} & 56.1~\loss{2.5} & 58.5~\loss{1.8} \\
            \midrule
            \; \multirow{2}{*}{Qwen3-VL-32B~\cite{qwen_vl_3}} & \rederror & 62.0 & 62.9 & 61.1 \\
            \; & \greencheck & 61.0~\loss{1.0} & 63.7~\gain{0.8} & 58.4~\loss{2.7} \\
            \midrule
            \; \multirow{2}{*}{Qwen3-VL-235B-A22B~\cite{qwen_vl_3}} & \rederror & 64.7 & 64.9 & 64.4 \\
            \; & \greencheck & 63.7~\loss{1.0} & 66.4~\gain{1.5} & 61.0~\loss{3.4} \\
            \midrule
            \; \multirow{2}{*}{GPT-5~\cite{openai2025gpt5}} & \rederror & 61.8 & 63.9 & 59.7 \\
            \; & \greencheck & 69.8~\gain{8.0} & 71.4~\gain{7.5} & 68.2~\gain{8.5} \\
            \bottomrule
            \end{tabular}%
        }
        \label{tab:thinking}
    \end{minipage}
    \hfill 
    \begin{minipage}[b]{0.48\linewidth}
        \centering
        \includegraphics[width=\linewidth]{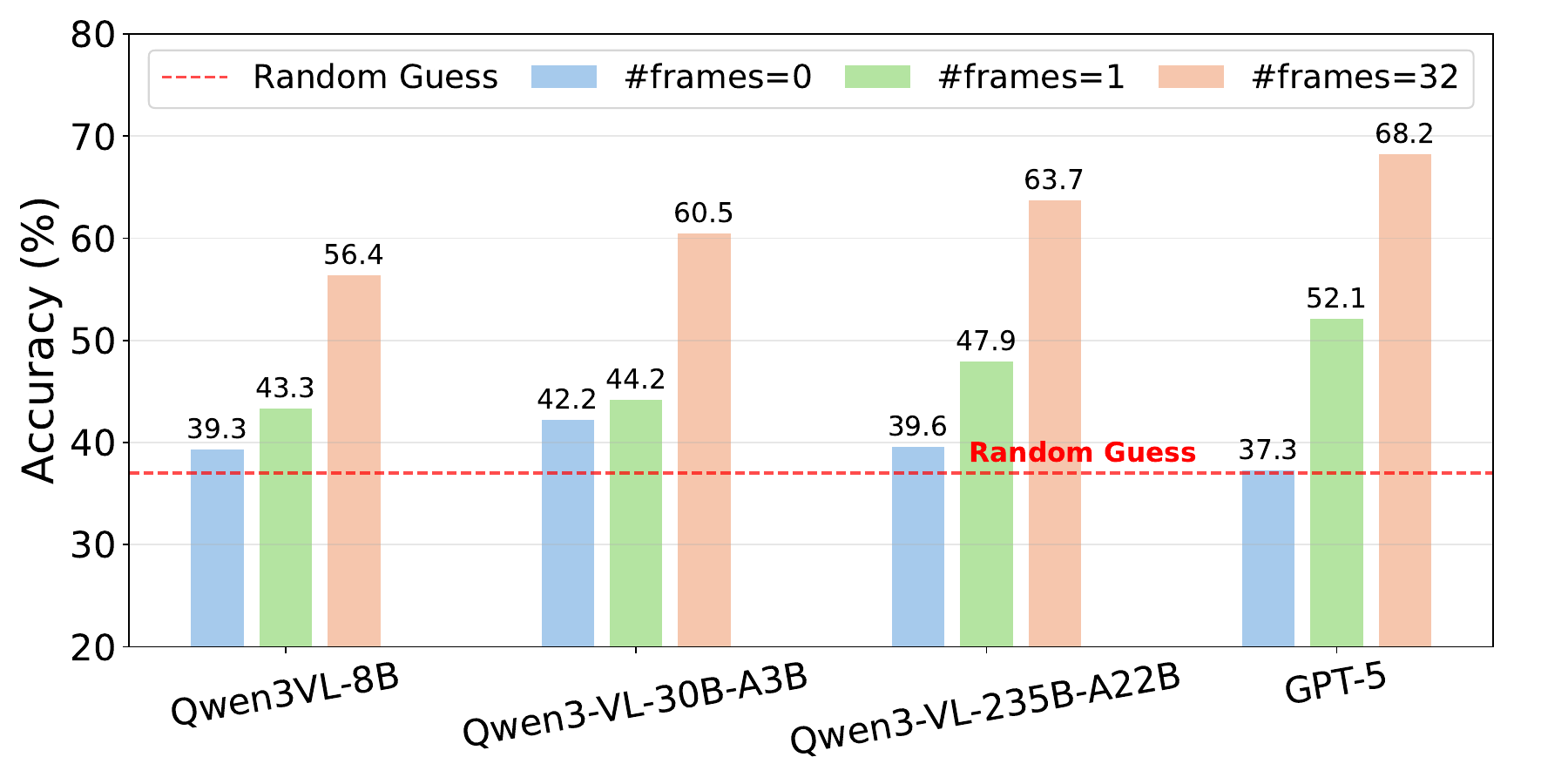}
        \vspace{-18pt} 
        \caption{Performance comparison across different frame counts on PAI-Bench-U.}
        \label{fig:reason-frame-analysis}
    \end{minipage}
\end{figure}

\thinparagraph{PAI-Bench-U is unbiased.}
Prevailing video understanding benchmarks often suffer from significant biases~\cite{goyal2017makingvvqamatter, tong2024cambrian, ramakrishnan2018overcominglanguagepriorsvisual, lei2022revealingsingleframebias}. Specifically, models frequently 1) exploit strong language priors to infer answers without grounding in visual data, and 2) leverage a \textit{static bias}, where questions are answerable from a single frame while obviating the need for temporal context. We analyzed PAI-Bench-U for these issues by evaluating SOTA models with varying input frame counts in Figure~\ref{fig:reason-frame-analysis}. First, with zero frames (\ie, text-only input), model performance degraded to the level of random guessing, which confirms that PAI-Bench-U effectively neutralizes language priors, ensuring that performance gains are sourced from visual understanding. Furthermore, we observed a substantial performance gap between the 1-frame and 32-frame inputs. This indicates that the tasks are not solvable with static information alone, validating PAI-Bench-U's requirement for temporal context comprehension.
\vspace{0.1cm}

\thinparagraph{MLLMs lag significantly behind human performance, with no clear proprietary model advantage.} The results in Table~\ref{tab:reason_results} highlight a substantial performance gap between all models and human accuracy (93.2). This high performance validates PAI-Bench-U as a well-posed benchmark; yet, the stark discrepancy underscores the significant limitations of current models in understanding capabilities in the Physical AI domain. Surprisingly, proprietary models do not uniformly outperform open-source models; for instance, Qwen3-VL-235B~\cite{qwen_vl_3} achieves a higher accuracy than GPT-5~\cite{openai2025gpt5}. This finding suggests that the Physical AI domain has not yet been a primary data collection or optimization priority for the wider field.

\thinparagraph{Textual thinking does not guarantee performance gains.} We investigated whether introducing an additional thinking process during inference enhances performance. As presented in Table~\ref{tab:thinking}, the Qwen3-VL~\cite{qwen_vl_3} series, which employs textual-only thinking, unexpectedly exhibited a slight performance degradation, particularly in embodied reasoning tasks. We attribute this to the high demands of these tasks, which require perceiving fine-grained visual details, such as minute regions or subtle temporal changes. When vision modules fail to capture these nuances, subsequent textual reasoning operates without the necessary grounding, rendering it ineffective. This hypothesis is supported by GPT-5~\cite{openai2025gpt5}’s results. We use medium reasoning effort for GPT-5~\cite{openai2025gpt5}, and the model shows clear performance gains under this setting. Crucially, GPT-5's~\cite{openai2025gpt5} reasoning involves both textual and visual thinking. This integrated visual reasoning appears to be an effective solution for capturing fine-grained information. Therefore, this finding underscores the necessity of advancing visual thinking capabilities within the Physical AI domain.

\section{Further Analysis and Discussion}
\label{sec:further_analysise}

Based on an in-depth analysis and cross-comparison of three tracks of PAI-Bench, we extract several common insights and subsequently synthesize a central conclusion:

\vspace{3pt}
\thinparagraph{Physical AI remains a largely unsolved domain.}
\noindent In PAI-Bench-G, leading open-source VGMs, including Wan2.2-I2V-A14B~\cite{wan2025} (81.6) and Cosmos-Predict2.5-2B~\cite{predict1} (81.4), achieve performance comparable to the proprietary Veo3~\cite{deepmind2025veo} (82.1), as shown in Table~\ref{tab:predict_results}. A similar trend is observed for MLLMs, where the open-source Qwen3-VL-235B-A22B~\cite{qwen_vl_3} (64.7) narrows the performance gap to proprietary GPT-5~\cite{openai2025gpt5} (61.8) to just 2.9 points in Table~\ref{tab:reason_results}. This lack of differentiation, coupled with generally low absolute performance, suggests one of two systemic challenges. 1) A community-wide data gap may exist, as Physical AI has not been a primary target for data collection or optimization. This would necessitate a focused, large-scale data collection effort to advance the domain. 2) Alternatively, the data may exist, but both proprietary and open-source models are failing to learn the specific capabilities required by Physical AI. For instance, VGMs may not be learning the underlying principles of the physical world~\cite{motamed2025generative, chen2024far, wiedemer2025videomodelszeroshotlearners}, while MLLMs may not be learning to leverage common sense to guide agent policies. Either scenario strongly indicates that the field of Physical AI remains in a nascent stage.

\section{Conclusion}
The evolution of AI toward embodied agency necessitates a mastery of physical world dynamics. To this end, we introduced PAI-Bench, a comprehensive benchmark evaluating the perception and prediction capabilities critical for this transition. By assessing performance across video generation (PAI-Bench-G), conditional control (PAI-Bench-C), and physical understanding (PAI-Bench-U), we established a rigorous baseline for Physical AI. Our evaluation reveals that while VGMs achieve high visual fidelity, they frequently fail to maintain physical consistency. Conversely, MLLMs demonstrate a significant performance deficit compared to human baselines in physical common sense reasoning and embodied forecasting. These findings confirm that Physical AI remains a nascent and largely unsolved domain. We hope PAI-Bench serves as a catalyst for future research to bridge this gap, shifting the focus from models that merely replicate the appearance of reality to those that internalize its underlying physics, enabling true understanding and robust action.

\section*{Acknowledgements}

We would like to thank NVIDIA Research, especially the Cosmos team, for their support, which led to the creation of PAI-Bench. We also thank Yin Cui, Jinwei Gu, Heng Wang, Prithvijit Chattopadhyay, Andrew Z. Wang, Imad El Hanafi, and Ming-Yu Liu for their valuable feedback and collaboration that helped shape the project. This research was supported in part by the National Science Foundation under Award \#2427478 - CAREER Program and by the National Science Foundation and the Institute of Education Sciences, U.S. Department of Education, under Award \#2229873 - National AI Institute for Exceptional Education. This project was also partially supported by cyberinfrastructure resources and services provided by the Georgia Institute of Technology. Also, we would like to thank the Cambrian~\cite{tong2024cambrian} Team for their awesome template; our template is based on their work.

\newpage
\printbibliography[heading=bibintoc]

\newpage
\appendix

\section{More Details about PAI-Bench-G}

\subsection{Data Source} \label{appen:source}
We curated data by integrating established open-source benchmarks with web-sourced content. The open-source component comprises EgoExo4D~\cite{grauman2024egoexo4dunderstandingskilledhuman}, HowTo100M~\cite{miech19howto100m}, Physics-IQ~\cite{motamed2025generative}, WISA-80K~\cite{wang2025wisa}, Agibot~\cite{bu2025agibot}, BridgeDatav2~\cite{walke2024bridgedatav2datasetrobot}, and Open X-Embodiment~\cite{open_x_embodiment_rt_x_2023}. Complementing these, we also collected raw video data from YouTube and stock footage platforms, including Pexels, Pixabay, Free-video, MixKit, and Free-stock Video.

\subsection{Details of Quality Score Calculation} \label{appen:quality_score}
For Quality Score, we adopt the evaluation protocol from VBench and VBench++~\cite{zheng2025vbench2,huang2024vbench++,huang2023vbench}. The specific components are defined as follows:

\noindent\textbf{Subject Consistency.} Evaluates the stability of the main subject's identity across frames. We extract frame-level features using DINO~\cite{caron2021emerging}, which offers robust identity-sensitive representations~\cite{ruiz2022dreambooth}. For a video with $T$ frames, the score is computed as:
\vspace{-4pt}
\begin{equation}
\label{eq:subject_consistency}
    S_{\textrm{subject}} = \frac{1}{T-1}\sum_{t=2}^{T} \frac{1}{2}(\langle d_1, d_t \rangle + \langle d_{t-1}, d_t\rangle),
    \vspace{-4pt}
\end{equation}
where $d_{i}$ represents the unit-normalized DINO feature of the $i$-th frame, and $\langle \cdot, \cdot \rangle$ denotes the dot product. This metric averages the similarity of each frame with both the initial frame and its immediate predecessor.

\noindent\textbf{Background Consistency.} Assesses the temporal stability of the background scene, independent of foreground motion. We utilize the CLIP image encoder~\cite{radford2021clip} to extract frame-level features. The score is calculated as:
\vspace{-4pt}
\begin{equation}
    S_{\textrm{background}} = \frac{1}{T-1}\sum_{t=2}^{T} \frac{1}{2}(\langle c_1, c_t \rangle + \langle c_{t-1}, c_t \rangle),
\label{eq:background_consistency}
\vspace{-4pt}
\end{equation}
where $c_{i}$ denotes the unit-normalized CLIP feature of the $i$-th frame. Similar to subject consistency, this metric aggregates similarities relative to the first and preceding frames.

\noindent\textbf{Motion Smoothness.} Quantifies the physical plausibility and temporal coherence of motion. Following standard motion priors, we assume real-world motion is locally linear or quadratic. We subsample the generated video $[f_0, f_1, \dots, f_{2n}]$ by dropping odd-indexed frames, reconstruct them using a frame interpolation model~\cite{licvpr23amt} to obtain $[\hat{f}_1, \hat{f}_3, \dots]$, and compute the Mean Absolute Error (MAE) between the original and reconstructed frames:
\vspace{-4pt}
\begin{equation}
S_{\textrm{smoothness}}
= \frac{1}{T/2}
\sum_{t=1}^{T/2}
\lVert f_{2t-1} - \hat{f}_{2t-1} \rVert_{1}.
\label{eq:smoothness_degree}
\vspace{-4pt}
\end{equation}
The final score is normalized to $[0, 1]$:
\begin{equation}
S_{\textrm{smoothness-norm}} = 1 - \frac{S_{\textrm{smoothness}}}{255},
\label{eq:smoothness_score}
\vspace{-4pt}
\end{equation}
where higher values indicate smoother motion dynamics.

\noindent\textbf{Aesthetic Quality.} Measures visual appeal, encompassing composition, color harmony, and artistic quality. We utilize the LAION aesthetic predictor~\cite{LAIONaes} to score each frame on a scale of $[0, 10]$. These scores are linearly normalized to $[0, 1]$ and averaged across all frames to derive the video-level metric.

\noindent\textbf{Imaging Quality.} Evaluates low-level image fidelity, specifically distortions such as overexposure, noise, and blur. We employ the MUSIQ predictor~\cite{Ke2021MUSIQ} (trained on SPAQ~\cite{Fang2020spaq}) to obtain frame-level scores in $[0, 100]$. The final score is the average of these normalized values.

\noindent\textbf{Overall Consistency.} Measures the semantic and stylistic alignment between the generated video and the textual prompt. We employ the video-text consistency score from ViCLIP~\cite{wang2023internvid}, which directly assesses the correspondence between the video content and the input description.

\begin{table*}[t]
\centering
\caption{\centering Generation parameters for the evaluated VGMs.}
\label{tab:model_video_specs}
\resizebox{0.80\linewidth}{!}{%
    \begin{tabular}{lc|cccc}
    \toprule
    Model & Version & \#Frames & FPS & Width & Height \\
    \midrule
    CogVideoX~\cite{yang2025cogvideoxtexttovideodiffusionmodels} & - & 49 & 16 & 720 & 480 \\
    CogVideoX1.5~\cite{yang2025cogvideoxtexttovideodiffusionmodels} & - & 81 & 16 & 1360 & 768 \\
    Cosmos-Predict2-2B~\cite{liu2025world} & - & 93 & 16 & 1280 & 704 \\
    Cosmos-Predict2-14B~\cite{liu2025world} & - & 93 & 16 & 1280 & 704 \\
    Cosmos-Predict2.5-2B~\cite{liu2025world} & \texttt{base/post-trained} & 93 & 16 & 1280 & 704 \\
    DynamicCrafter~\cite{xing2023dynamicrafter} & \texttt{DynamiCrafter\_1024} & 16 & 8 & 1024 & 576 \\
    HunyuanVideo-I2V~\cite{kong2024hunyuanvideo} & - & 129 & 24 & 1184 & 768 \\
    LTX-Video-2B~\cite{HaCohen2024LTXVideo} & - & 121 & 30 & 1216 & 704 \\
    LTX-Video-13B~\cite{HaCohen2024LTXVideo} & - & 121 & 30 & 1216 & 704 \\
    MAGI-1-4.5B~\cite{ai2025magi1autoregressivevideogeneration} & - & 120 & 24 & 720 & 720 \\
    MAGI-1-24B~\cite{ai2025magi1autoregressivevideogeneration} & - & 120 & 24 & 1280 & 720 \\
    Veo3~\cite{deepmind2025veo} & \texttt{veo-3.0-generate-001} & 192 & 24 & 1280 & 720 \\
    Wan2.1-I2V-14B~\cite{wan2025} & \texttt{Wan2.1-I2V-14B-720P} & 81 & 16 & 1280 & 720 \\
    Wan2.2-TI2V-5B~\cite{wan2025} & - & 121 & 24 & 1248 & 704 \\
    Wan2.2-I2V-A14B~\cite{wan2025} & - & 81 & 16 & 1280 & 720 \\
    \bottomrule
    \end{tabular}%
}
\end{table*}

\noindent\textbf{Image-to-Video Subject.} Ensures the subject in the generated video remains faithful to the input reference image. Using DINO~\cite{caron2021emerging} features, we compute the similarity between the input image ($s_{\textrm{img}}$) and video frames ($s_t$):
\vspace{-4pt}
\begin{equation}
    S_{\textrm{i2v\_subject}} = \frac{1}{T-1}\sum_{t=2}^{T} \frac{1}{2}\left(\langle s_{\textrm{img}}, s_t \rangle + \langle s_{t-1}, s_t \rangle\right).
    \label{eq:i2v_subject}
    \vspace{-4pt}
\end{equation}
The final score is averaged across all image-conditioned samples.

\noindent\textbf{Image-to-Video Background.} Verifies consistency between the generated background and the input image environment. This is critical for inputs emphasizing scene layout. We extract features using DreamSim~\cite{fu2023learning}, which is sensitive to layout variations, and compute:
\vspace{-4pt}
\begin{equation}
    S_{\textrm{i2v\_bg}} = \frac{1}{T-1}\sum_{t=2}^{T} \frac{1}{2}\left(\langle b_{\textrm{img}}, b_t \rangle + \langle b_{t-1}, b_t \rangle\right),
    \label{eq:i2v_background}
    \vspace{-4pt}
\end{equation}
where $b_{\textrm{img}}$ and $b_t$ are the unit-normalized DreamSim features of the reference image and the $t$-th frame, respectively.

\subsection{Details of Domain Score Calculation}
The Domain Score quantifies the adherence of generated videos to domain-specific physical and semantic constraints. We employ Qwen3-VL-235B-A22B-Instruct~\cite{qwen_vl_3} as an automated evaluator, querying the model with a curated set of verification questions derived from the ground truth. The score is computed as the accuracy of the model's binary responses against the expected answers. For inference, we uniformly sample frames at 2 fps and utilize greedy decoding to ensure deterministic evaluation. The specific prompt template used for evaluation is illustrated in Figure~\ref{fig:judger-system-prompt}.

\begin{figure}[h]
    \begin{tcolorbox}[
        title=\textbf{Prompt Template for PAI-Bench-G Domain Score Evaluation},
        colback=gray!3,
        colframe=black!60,
        fonttitle=\bfseries,
        sharp corners,
        boxrule=0.6pt
    ]
    You are a helpful AI assistant that answers questions about videos. Answer with just YES or NO.
    I'll show you a video with several frames. Please look carefully at all frames to understand what's happening in the video, then answer the question about the video with either YES or NO.

    \vspace{0.2cm}
    \textbf{Input}\\
    \textit{\{Sampled Frames 0\}}\\
    \textit{\{Sampled Frames 1\}}\\
    \ldots\\
    \textit{\{Sampled Frames $N$\}}\\

    \vspace{0.2cm}
    \textbf{Question:}\\
    Question: \textit{\{Question\}}\\[3pt]
    Please answer with YES or NO and explain your reasoning.

    \end{tcolorbox}
    \vspace{-6pt}
    \caption{\textbf{Prompt template utilized for Domain Score evaluation.} The model receives uniformly sampled frames and a specific constraint question to verify physical compliance.}
    \vspace{-12pt}
    \label{fig:judger-system-prompt}
\end{figure}

\subsection{Details about Experiments}

\noindent \textbf{Model Configurations.} To ensure reproducibility, we detail the inference specifications for all evaluated VGMs in Table~\ref{tab:model_video_specs}. We report the specific checkpoint versions alongside their corresponding spatial resolutions ($W \times H$) and temporal settings (frame count and FPS) used during the generation process.

\begin{figure*}[tbp]
    \centering
    \includegraphics[width=0.85\textwidth]{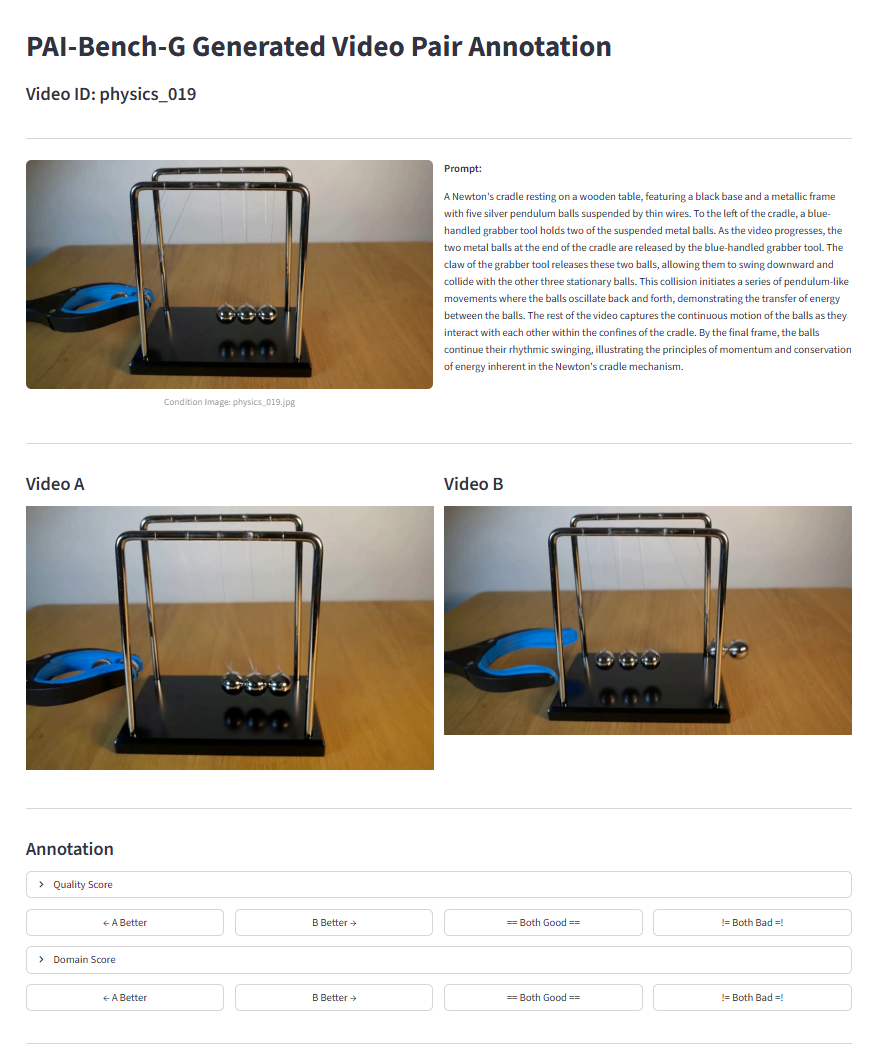}
    \caption{User study interface for preferences on PAI-Bench-G.}
    \label{fig:user_study_interface}
\end{figure*}

\noindent\textbf{Human Evaluation Protocol.} To validate our automated metrics against human preferences, we conducted a pairwise comparison study using a web-based interface, as shown in Fig.~\ref{fig:user_study_interface}. Annotators were presented with a text prompt, a reference image, and two generated videos (labeled A and B). Evaluation focused on two distinct dimensions: (1) \textit{Video Quality}, which assesses visual coherence, motion smoothness, and clarity (correlating with our Quality Score); and (2) \textit{Physical Plausibility}, which evaluates adherence to physical laws and domain-specific realism (correlating with our Domain Score). For each dimension, participants selected one of four outcomes: \textit{A Better}, \textit{B Better}, \textit{Both Good}, or \textit{Both Bad}.

\noindent \textbf{ELO Rating Formulation.} We quantified model rankings using an ELO rating system initialized at 1000 with a $K$ of 32. Pairwise comparisons are converted into numerical scores: a definitive preference is assigned 1.0 to the chosen model and 0.0 to the other, while ties expressed as \textit{Both Good} or \textit{Both Bad} are assigned 0.5 to each. We maintained independent ELO trackers for video quality and physical plausibility. For the Overall score, we computed a separate ELO rating by averaging the quality and plausibility preference scores for each comparison and using these averaged preference values as the inputs to the ELO update process. We present the detailed ELO rating results in Table~\ref{tab:predict_elo_summary}.

\begin{table*}[t]
  \centering

  \caption{\centering Human-study (ELO) vs. MLLM evaluation on PAI-Bench-G.}
  \vspace{-6pt}

  \resizebox{0.8\textwidth}{!}{%
  \begin{tabular}{@{}l|ccc|ccc@{}}
  \toprule
  \multirow{2}{*}{Models} & \multicolumn{3}{c|}{ELO Scores} & \multicolumn{3}{c}{Evaluation Scores} \\
  \cmidrule(lr){2-4} \cmidrule(lr){5-7}
  & Overall & Domain & Quality & Overall & Domain & Quality \\
  \midrule
  Source Videos & 1225.0 & 1278.1 & 1176.0 & 83.9 & 89.8 & 78.0 \\
  Veo3~\cite{deepmind2025veo} & 1112.2 & 1122.8 & 1102.9 & 82.2 & 86.8 & 77.6 \\
  DynamicCrafter~\cite{xing2023dynamicrafter} & 734.9 & 749.9 & 718.9 & 68.3 & 63.0 & 73.7 \\
  HunyuanVideo-I2V~\cite{kong2024hunyuanvideo} & 964.3 & 946.6 & 981.1 & 77.6 & 77.1 & 78.0 \\
  CogVideoX1.5~\cite{yang2025cogvideoxtexttovideodiffusionmodels} & 907.9 & 896.4 & 918.1 & 78.4 & 80.3 & 76.6 \\
  MAGI-1-24B~\cite{ai2025magi1autoregressivevideogeneration} & 988.9 & 969.3 & 1007.7 & 79.4 & 82.4 & 76.5 \\
  Wan2.1-I2V-14B~\cite{wan2025} & 976.0 & 961.4 & 988.7 & 80.8 & 84.3 & 77.3 \\
  Cosmos-Predict2.5-2B~\cite{liu2025world} & 1008.1 & 968.5 & 1047.8 & 81.4 & 84.9 & 78.0 \\
  Wan2.2-I2V-A14B~\cite{wan2025} & 1082.6 & 1106.9 & 1058.7 & 82.3 & 87.1 & 77.5 \\
  \bottomrule
  \end{tabular}%
  }
  \label{tab:predict_elo_summary}
\end{table*}

\subsection{Qualitative Visualization}

\begin{figure*}[p]
    \begin{tcolorbox}[colback=white, colframe=black, arc=4mm, boxrule=0.7pt, width=0.98\linewidth, left=3pt, right=3pt, top=3pt, bottom=3pt]
        \centering
        \begin{minipage}{0.98\linewidth}

            \centering
            \begin{subfigure}{\linewidth}
                \begin{minipage}[t]{0.24\linewidth}
                    \centering
                    \vspace{0pt}
                    \includegraphics[width=\linewidth]{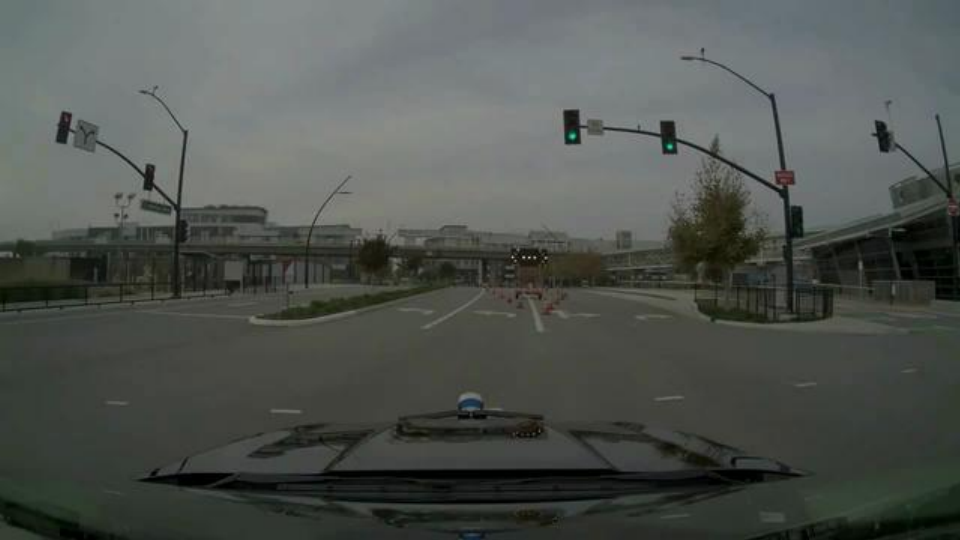}
                \end{minipage}%
                \hfill
                \begin{minipage}[t]{0.72\linewidth}
                    \vspace{0pt}
                    \scriptsize
                    The video begins with a view from inside a vehicle, likely captured by a dashboard camera, showing a wide, open road ahead under an overcast sky. The road is marked with white directional arrows indicating left or straight-right turns. Two lines of orange traffic cones are placed along the center of the road, suggesting some form of construction or roadwork. Within the area enclosed by the traffic cones, \textcolor{red}{an arrow board displays flashing arrows pointing left and right} to guide traffic around the detour. On either side of the road, there are grassy areas with small trees and shrubs. In the background, modern buildings, possibly part of an urban or suburban area, are visible, with a pedestrian bridge crossing above the road. The overall atmosphere is calm and quiet, with no other vehicles or pedestrians in sight.As the video progresses, \textcolor{red}{the car changes lanes to the left and continues along the road}. The camera angle shifts slightly to show the car moving further down the road, \textcolor{red}{passing more traffic cones, and approaching the pedestrian bridge}, which remains visible above the road.
                \end{minipage}
                \caption{Input condition signals}
            \end{subfigure}

            \centering

            \begin{subfigure}{\linewidth}
                \centering
                \includegraphics[width=\linewidth]{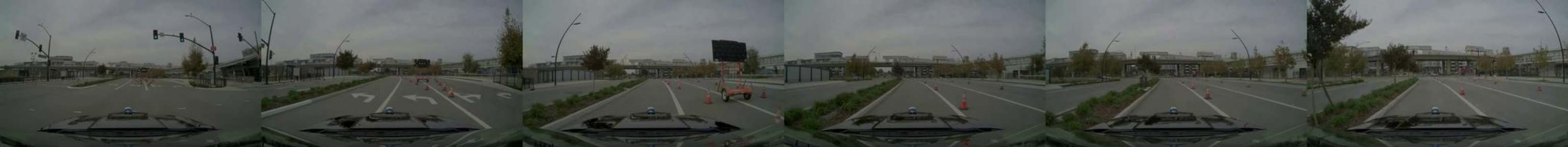}
                \caption{Source video}
            \end{subfigure}

            \begin{subfigure}{\linewidth}
                \centering
                \includegraphics[width=\linewidth]{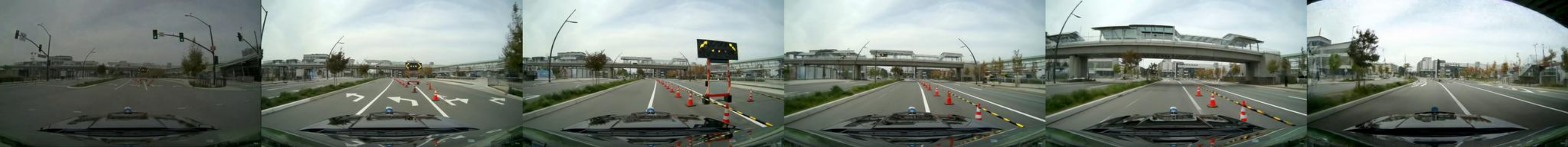}
                \caption{Veo3}
            \end{subfigure}

            \begin{subfigure}{\linewidth}
                \centering
                \includegraphics[width=\linewidth]{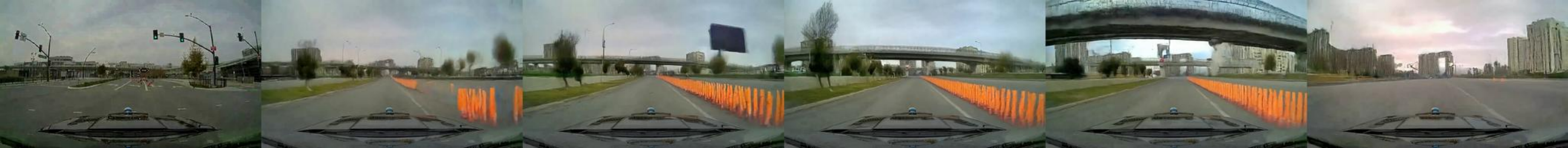}
                \caption{CogVideoX1.5}
            \end{subfigure}

            \begin{subfigure}{\linewidth}
                \centering
                \includegraphics[width=\linewidth]{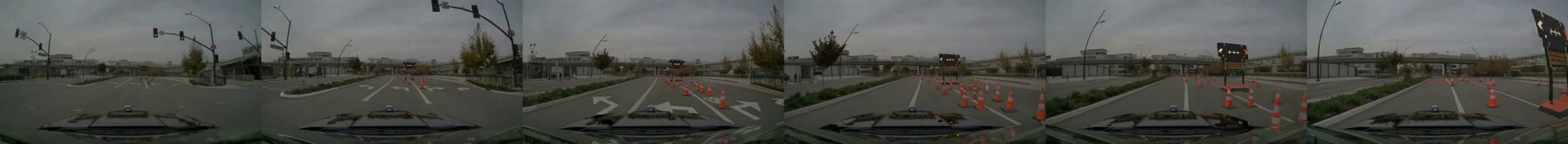}
                \caption{Cosmos-Predict2.5-2B}
            \end{subfigure}

            \begin{subfigure}{\linewidth}
                \centering
                \includegraphics[width=\linewidth]{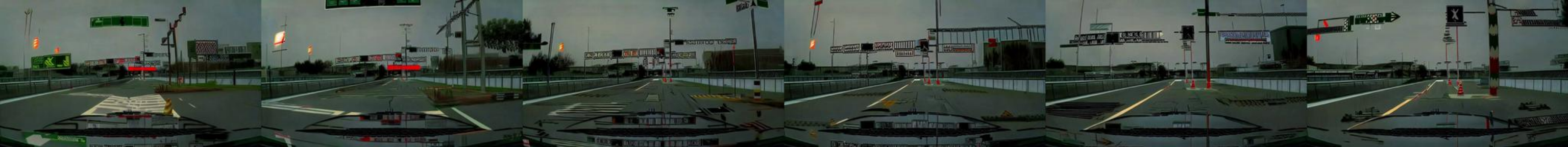}
                \caption{DynamicCrafter}
            \end{subfigure}

            \begin{subfigure}{\linewidth}
                \centering
                \includegraphics[width=\linewidth]{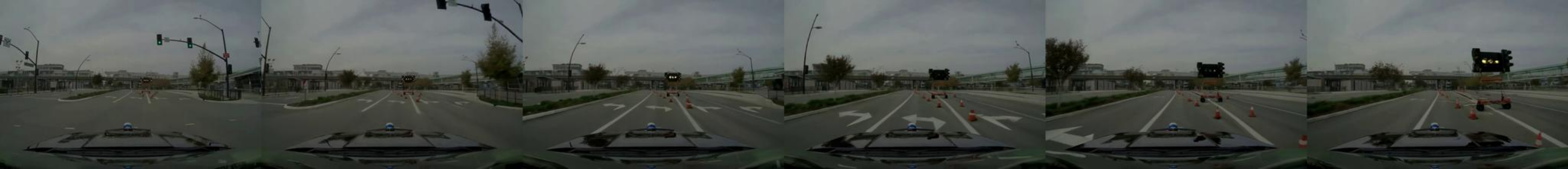}
                \caption{HunyuanVideo-I2V}
            \end{subfigure}

            \begin{subfigure}{\linewidth}
                \centering
                \includegraphics[width=\linewidth]{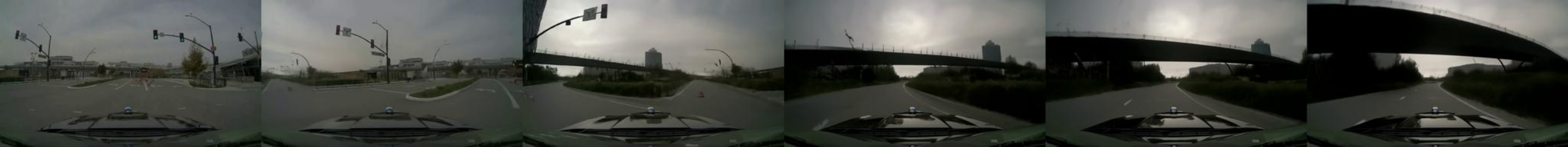}
                \caption{MAGI-1-24B}
            \end{subfigure}

            \begin{subfigure}{\linewidth}
                \centering
                \includegraphics[width=\linewidth]{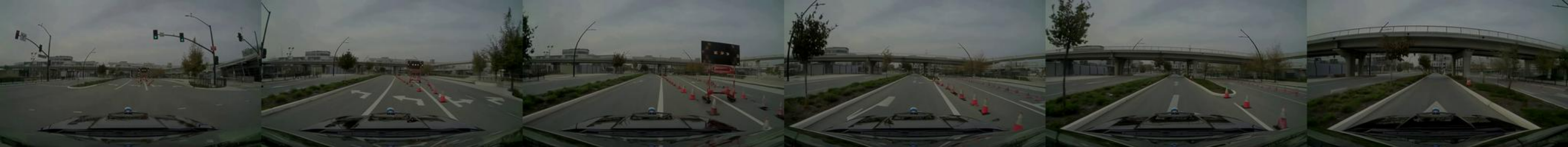}
                \caption{Wan2.2-I2V-A14B}
            \end{subfigure}

        \end{minipage}
    \end{tcolorbox}

    \caption{
        \textbf{Example of autonomous vehicles domain and model generations from PAI-Bench-G.} \textit{Best viewed with zoom.}
    }
    \label{fig:predict_examples_av}
\end{figure*}

\begin{figure*}[p]
    \begin{tcolorbox}[colback=white, colframe=black, arc=4mm, boxrule=0.7pt, width=0.98\linewidth, left=3pt, right=3pt, top=3pt, bottom=3pt]
        \centering
        \begin{minipage}{0.98\linewidth}

            \centering
            \begin{subfigure}{\linewidth}
                \begin{minipage}[t]{0.24\linewidth}
                    \centering
                    \vspace{0pt}
                    \includegraphics[width=\linewidth]{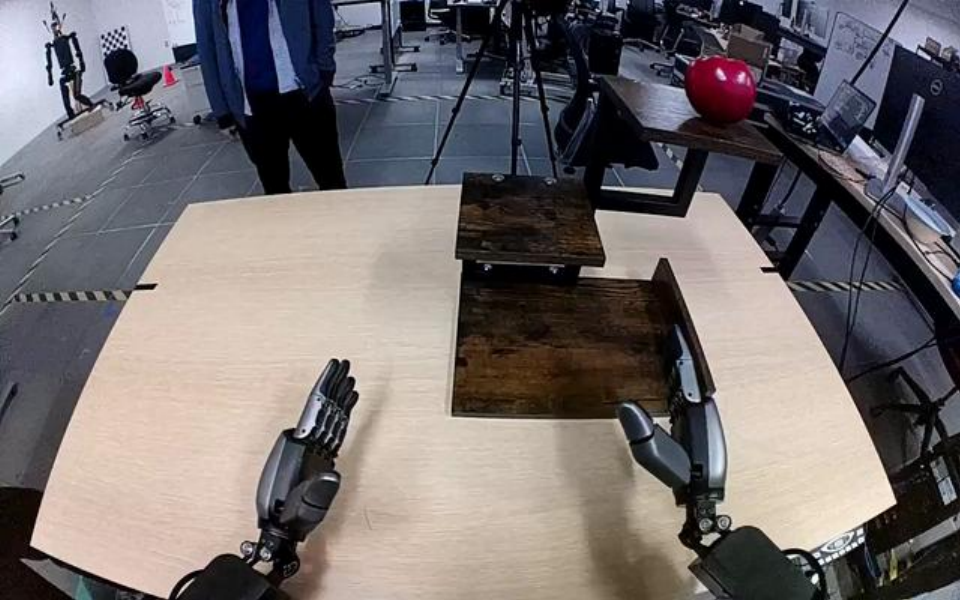}
                \end{minipage}%
                \hfill
                \begin{minipage}[t]{0.72\linewidth}
                    \vspace{0pt}
                    \scriptsize
                    The video opens with a view of a testing environment, characterized by a large wooden table at the center. On this table, two robot arms are positioned at opposite ends, with the left arm closer to the camera and the right arm further away. Between the hands lies a dark wooden shelf with a red spherical object on its top rack, likely serving as a platform or obstacle. In the background, various pieces of equipment, including a tripod, a chair, are visible. \textcolor{gray}{[TRUNCATED]}  As the video progresses, \textcolor{red}{the right robotic hand extends outward}, moving from its initial position towards the red spherical object on the shelf. The hand then \textcolor{red}{picks up the object and places it on the lowest rack of the shelf}, completing a smooth, deliberate manipulation. The left robotic hand remains stationary throughout the sequence. No new objects appear in the video; all existing elements maintain their positions except for the movement of the right robotic hand. The scene concludes with the right robotic hand returning to its initial position, while the left hand continues to rest on the table. The overall environment remains unchanged, with the focus remaining on the interaction between the robotic hands and the wooden block, highlighting precise control during the demonstration.
                \end{minipage}
                \caption{Input condition signals}
            \end{subfigure}

            \centering

            \begin{subfigure}{\linewidth}
                \centering
                \includegraphics[width=\linewidth]{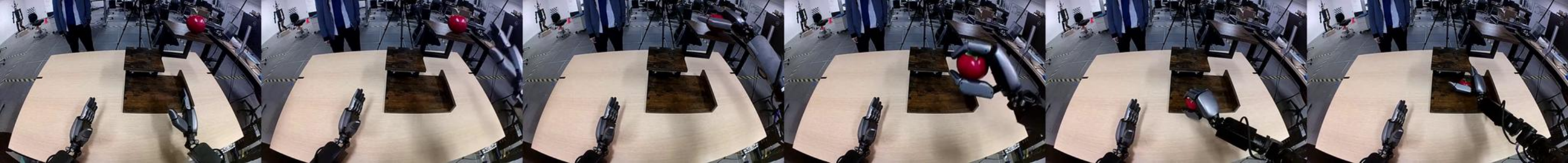}
                \caption{Source video}
            \end{subfigure}

            \begin{subfigure}{\linewidth}
                \centering
                \includegraphics[width=\linewidth]{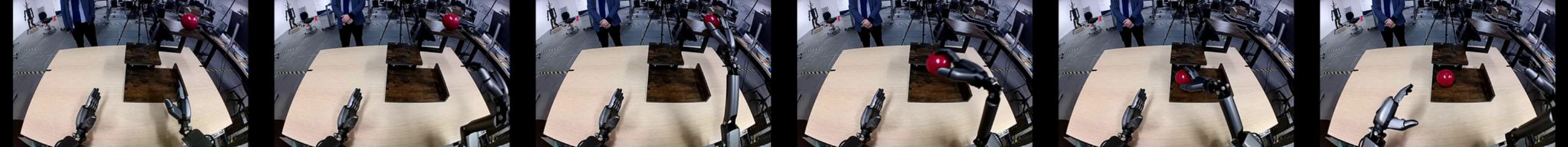}
                \caption{Veo3}
            \end{subfigure}

            \begin{subfigure}{\linewidth}
                \centering
                \includegraphics[width=\linewidth]{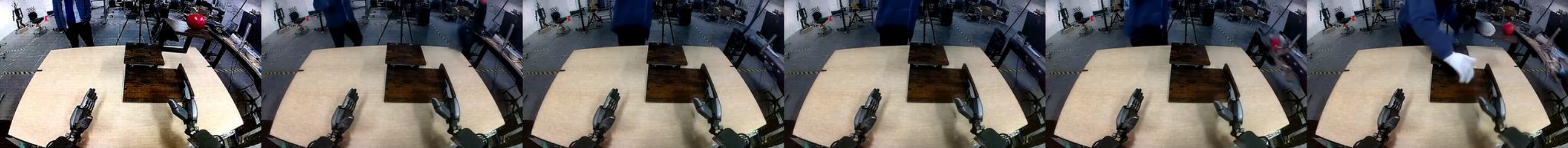}
                \caption{CogVideoX1.5}
            \end{subfigure}

            \begin{subfigure}{\linewidth}
                \centering
                \includegraphics[width=\linewidth]{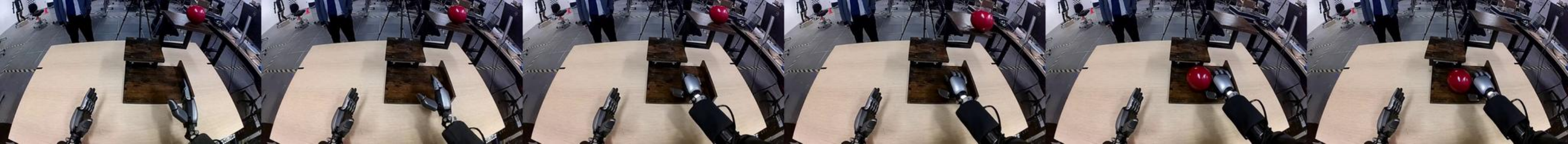}
                \caption{Cosmos-Predict2.5-2B}
            \end{subfigure}

            \begin{subfigure}{\linewidth}
                \centering
                \includegraphics[width=\linewidth]{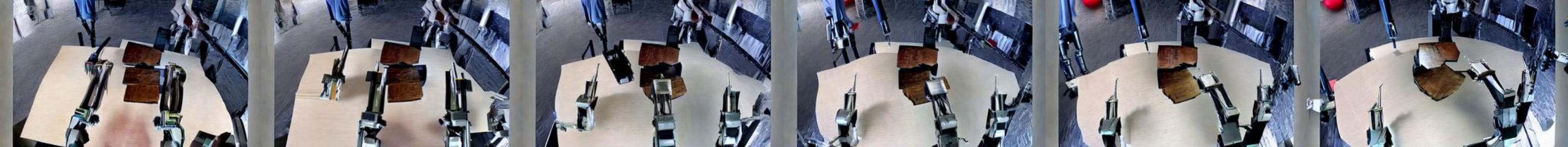}
                \caption{DynamicCrafter}
            \end{subfigure}

            \begin{subfigure}{\linewidth}
                \centering
                \includegraphics[width=\linewidth]{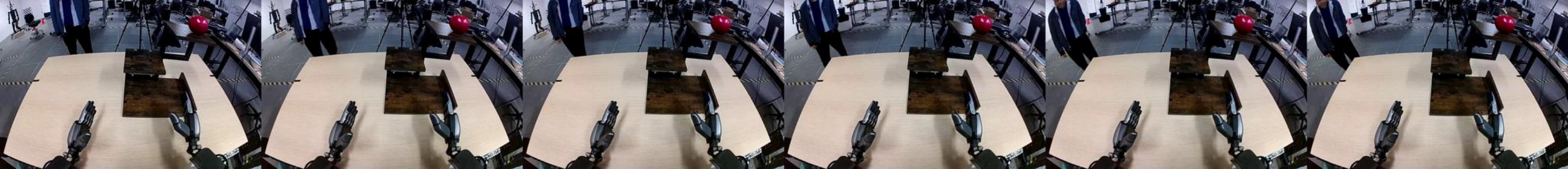}
                \caption{HunyuanVideo-I2V}
            \end{subfigure}

            \begin{subfigure}{\linewidth}
                \centering
                \includegraphics[width=\linewidth]{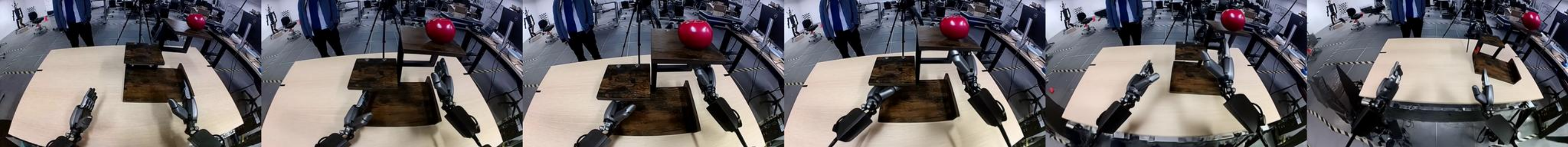}
                \caption{MAGI-1-24B}
            \end{subfigure}

            \begin{subfigure}{\linewidth}
                \centering
                \includegraphics[width=\linewidth]{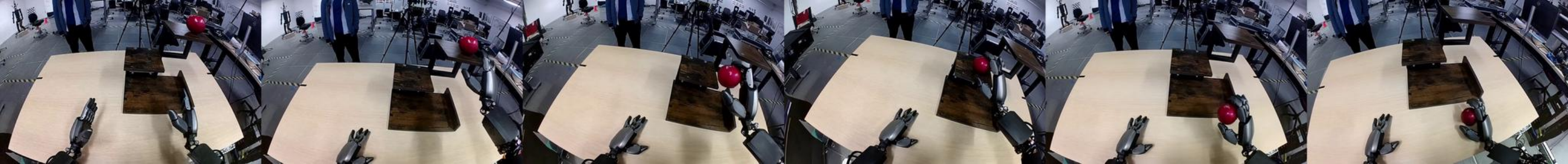}
                \caption{Wan2.2-I2V-A14B}
            \end{subfigure}

        \end{minipage}
    \end{tcolorbox}

    \caption{
        \textbf{Example of robotics domain and model generations from PAI-Bench-G.} \textit{Best viewed with zoom.}
    }
    \label{fig:predict_examples_robotics}
\end{figure*}

\begin{figure*}[p]

    \begin{tcolorbox}[colback=white, colframe=black, arc=4mm, boxrule=0.7pt, width=0.98\linewidth, left=3pt, right=3pt, top=3pt, bottom=3pt]
        \centering
        \begin{minipage}{0.98\linewidth}

            \centering
            \begin{subfigure}{\linewidth}
                \begin{minipage}[t]{0.24\linewidth}
                    \centering
                    \vspace{0pt}
                    \includegraphics[width=\linewidth]{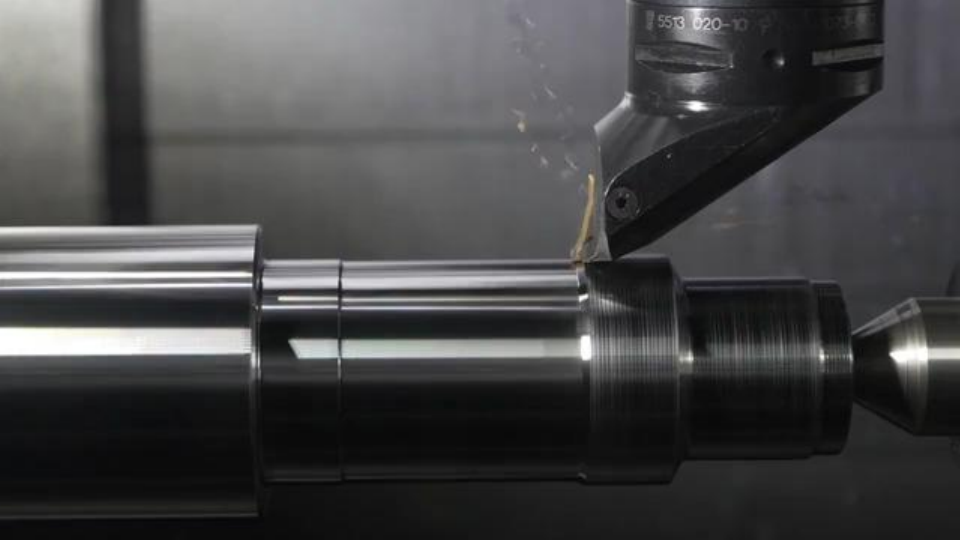}
                \end{minipage}%
                \hfill
                \begin{minipage}[t]{0.72\linewidth}
                    \vspace{0pt}
                    \scriptsize
                    A close-up of a precision metalworking process in a controlled industrial setting. The first frame captures a cylindrical metal workpiece securely mounted on a lathe, rotating smoothly as a cutting machine, held by a black, angular fixture, approaches from above. The cutting machine, marked with numerical identifiers (5513 020-10), engages with the workpiece, shaving off thin metal shavings that are visibly ejected into the air, creating a fine mist around the machining area. \textcolor{gray}{[TRUNCATED]} As the video progresses, \textcolor{red}{the cutting machine continues its linear motion along the length of the workpiece}, maintaining a steady pace. The tool's engagement with the material results in \textcolor{red}{consistent metal shaving, producing a continuous stream of shavings that are dispersed into the surrounding space}. The workpiece remains stationary relative to the camera's perspective, ensuring a clear view of the cutting metal process. The environment suggests a well-lit workshop, emphasizing the precision and efficiency of the operation. By the final frame, the cutting machine has almost completed its pass along the workpiece, \textcolor{red}{leaving behind a smooth, polished surface}. The metal shavings continue to be ejected, and the overall scene maintains a focused and industrious atmosphere, underscoring the meticulous nature of the metalworking process.
                \end{minipage}
                \caption{Input condition signals}
            \end{subfigure}

            \centering

            \begin{subfigure}{\linewidth}
                \centering
                \includegraphics[width=\linewidth]{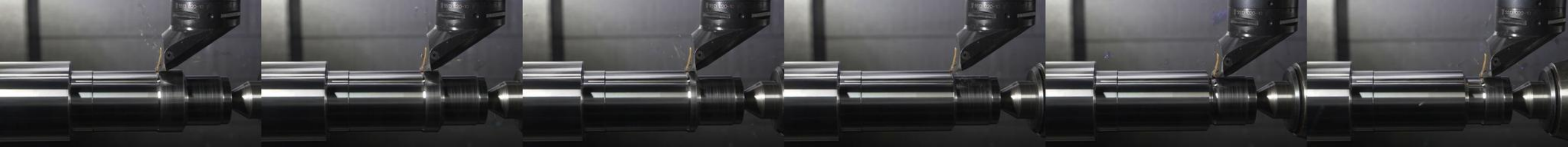}
                \caption{Source video}
            \end{subfigure}

            \begin{subfigure}{\linewidth}
                \centering
                \includegraphics[width=\linewidth]{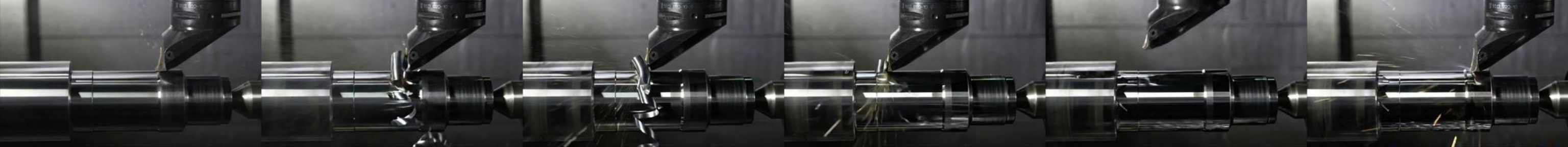}
                \caption{Veo3}
            \end{subfigure}

            \begin{subfigure}{\linewidth}
                \centering
                \includegraphics[width=\linewidth]{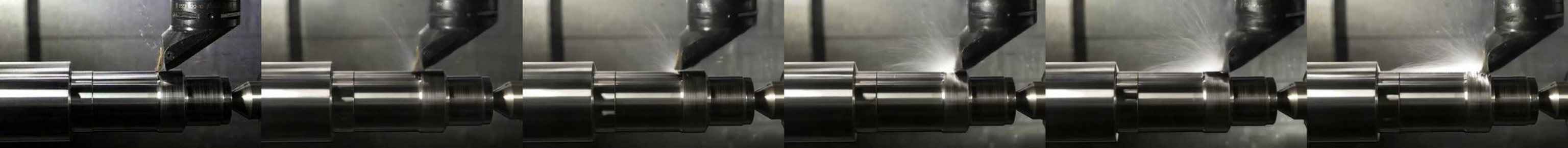}
                \caption{CogVideoX1.5}
            \end{subfigure}

            \begin{subfigure}{\linewidth}
                \centering
                \includegraphics[width=\linewidth]{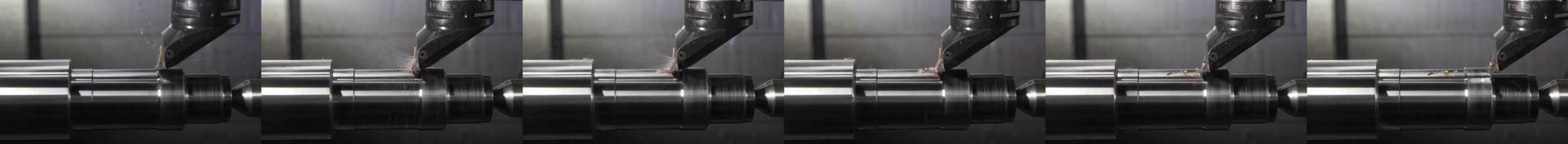}
                \caption{Cosmos-Predict2.5-2B}
            \end{subfigure}

            \begin{subfigure}{\linewidth}
                \centering
                \includegraphics[width=\linewidth]{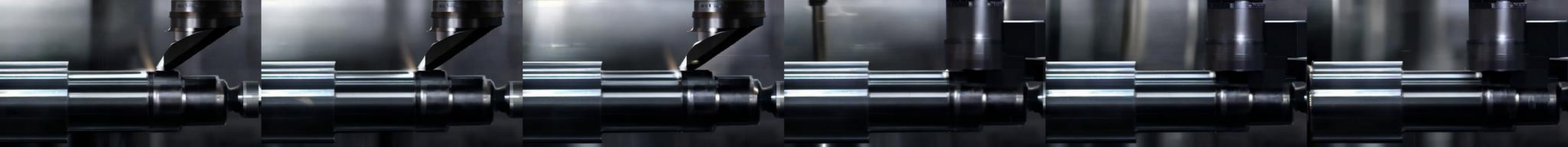}
                \caption{DynamicCrafter}
            \end{subfigure}

            \begin{subfigure}{\linewidth}
                \centering
                \includegraphics[width=\linewidth]{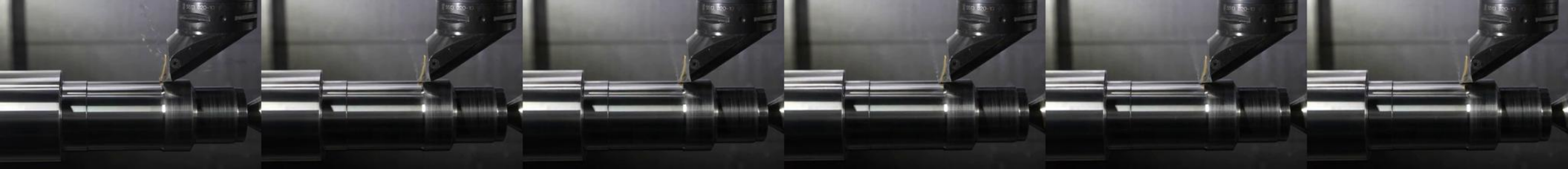}
                \caption{HunyuanVideo-I2V}
            \end{subfigure}

            \begin{subfigure}{\linewidth}
                \centering
                \includegraphics[width=\linewidth]{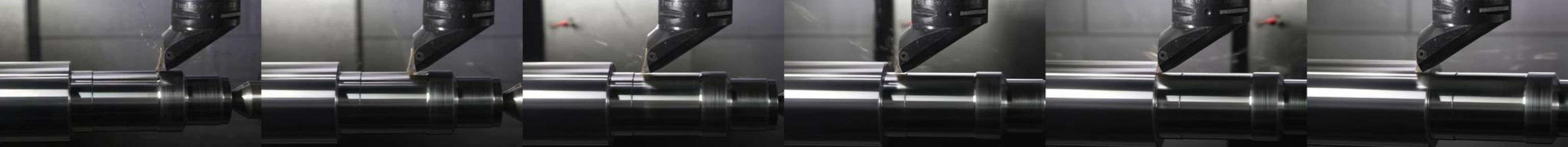}
                \caption{MAGI-1-24B}
            \end{subfigure}

            \begin{subfigure}{\linewidth}
                \centering
                \includegraphics[width=\linewidth]{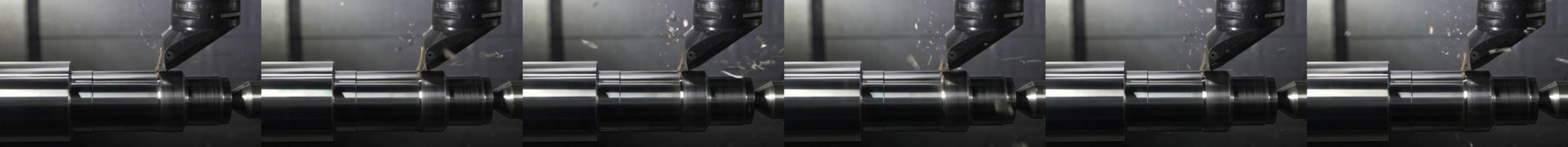}
                \caption{Wan2.2-I2V-A14B}
            \end{subfigure}

        \end{minipage}
    \end{tcolorbox}

    \caption{
        \textbf{Example of industry domain and model generations from PAI-Bench-G.} \textit{Best viewed with zoom.}
    }
    \label{fig:predict_examples_industry}
\end{figure*}

\begin{figure*}[p]

    \begin{tcolorbox}[colback=white, colframe=black, arc=4mm, boxrule=0.7pt, width=0.98\linewidth, left=3pt, right=3pt, top=3pt, bottom=3pt]
        \centering
        \begin{minipage}{0.98\linewidth}

            \centering
            \begin{subfigure}{\linewidth}
                \begin{minipage}[t]{0.24\linewidth}
                    \centering
                    \vspace{0pt}
                    \includegraphics[width=\linewidth]{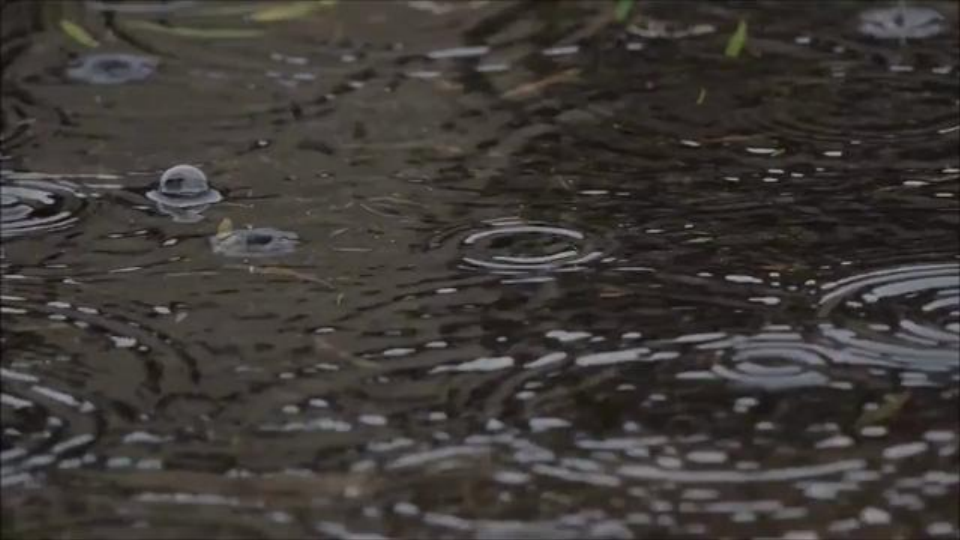}
                \end{minipage}%
                \hfill
                \begin{minipage}[t]{0.72\linewidth}
                    \vspace{0pt}
                    \scriptsize
                    A close-up shot captures a wet, dark surface, likely a puddle or damp ground, under an overcast sky. The surface is covered with small ripples and bubbles, indicating recent rain. Several droplets of water form small, round bubbles that float on the surface, while others create concentric circles as they spread outwards. Tiny bits of debris, including leaves and twigs, are scattered across the surface, adding texture. As the video progresses, \textcolor{red}{the rain continues to fall, creating more ripples and bubbles}. Each raindrop impacts the water, causing a series of concentric circles that expand outward from the point of impact. \textcolor{red}{The bubbles grow slightly larger} as they float on the surface, reflecting the light and adding a shimmering effect. The scattered debris remains stationary, but the ripples and bubbles continuously shift and evolve, capturing the dynamic nature of the rainfall. The scene remains focused on the wet, reflective surface, with the ongoing rain creating a continuous pattern of ripples and bubbles.
                \end{minipage}
                \caption{Input condition signals}
            \end{subfigure}

            \centering

            \begin{subfigure}{\linewidth}
                \centering
                \includegraphics[width=\linewidth]{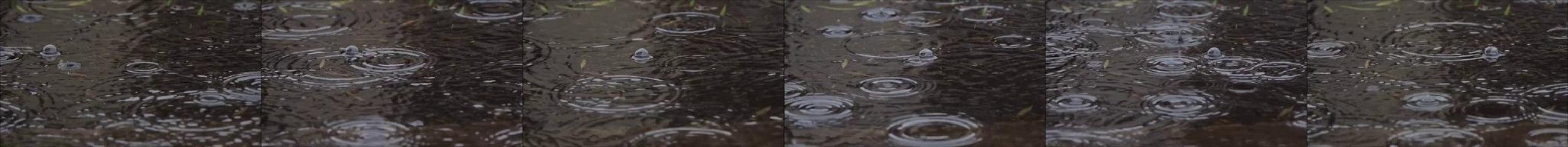}
                \caption{Source video}
            \end{subfigure}

            \begin{subfigure}{\linewidth}
                \centering
                \includegraphics[width=\linewidth]{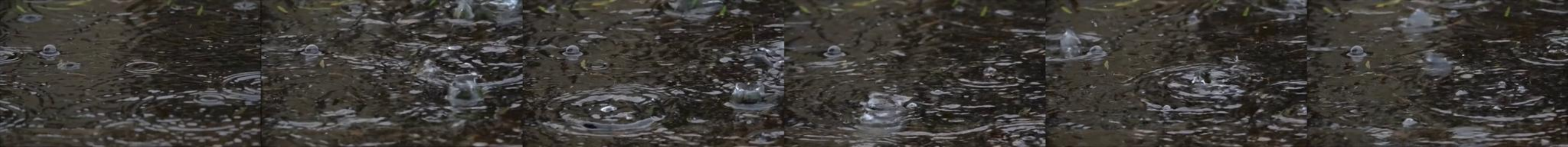}
                \caption{Veo3}
            \end{subfigure}

            \begin{subfigure}{\linewidth}
                \centering
                \includegraphics[width=\linewidth]{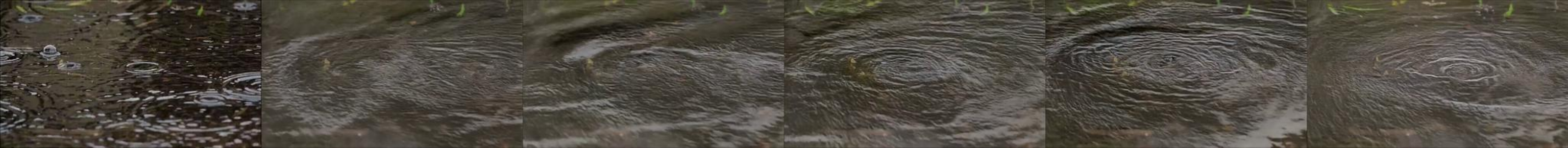}
                \caption{CogVideoX1.5}
            \end{subfigure}

            \begin{subfigure}{\linewidth}
                \centering
                \includegraphics[width=\linewidth]{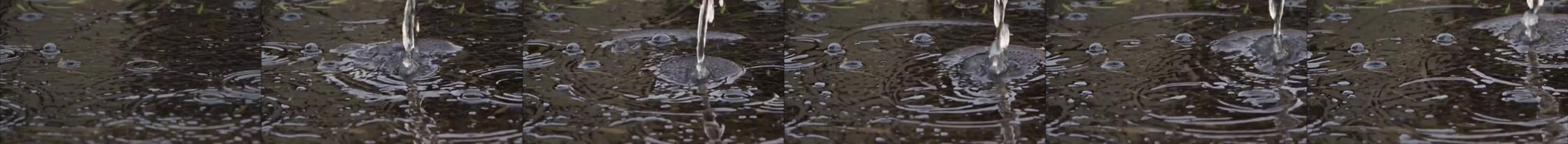}
                \caption{Cosmos-Predict2.5-2B}
            \end{subfigure}

            \begin{subfigure}{\linewidth}
                \centering
                \includegraphics[width=\linewidth]{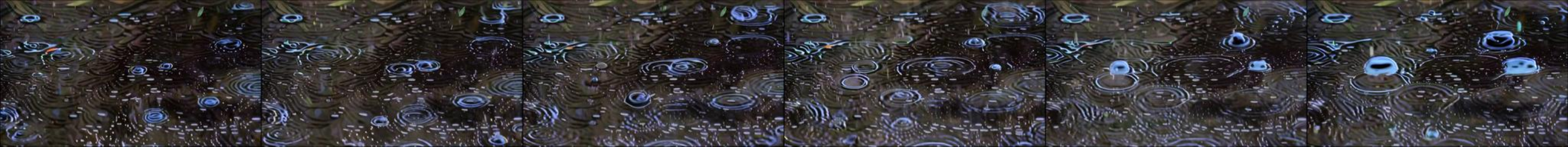}
                \caption{DynamicCrafter}
            \end{subfigure}

            \begin{subfigure}{\linewidth}
                \centering
                \includegraphics[width=\linewidth]{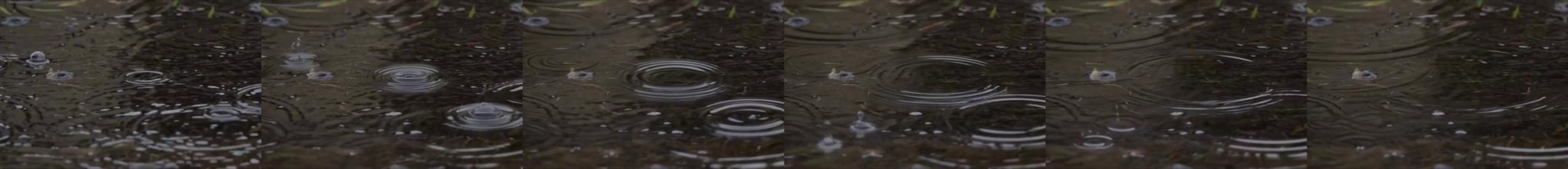}
                \caption{HunyuanVideo-I2V}
            \end{subfigure}

            \begin{subfigure}{\linewidth}
                \centering
                \includegraphics[width=\linewidth]{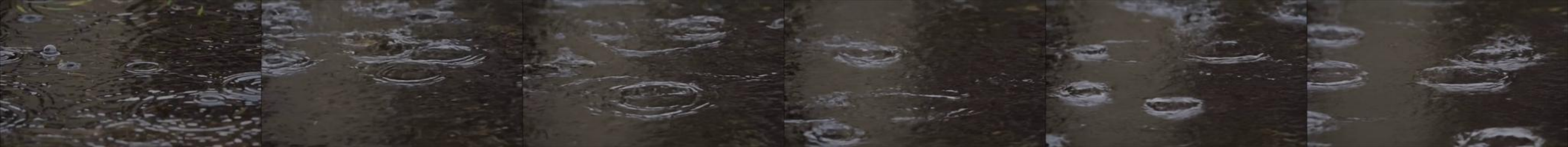}
                \caption{MAGI-1-24B}
            \end{subfigure}

            \begin{subfigure}{\linewidth}
                \centering
                \includegraphics[width=\linewidth]{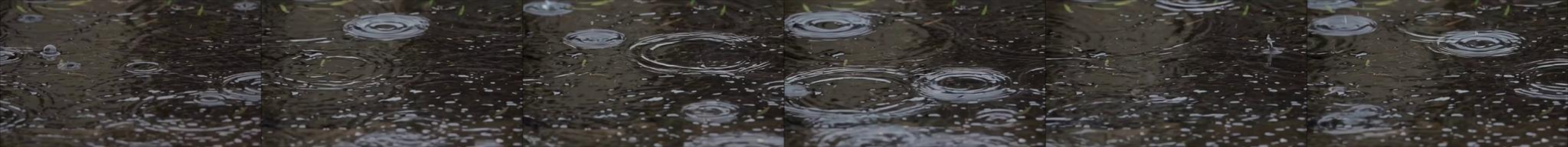}
                \caption{Wan2.2-I2V-A14B}
            \end{subfigure}

        \end{minipage}
    \end{tcolorbox}

    \caption{
        \textbf{Example of common sense domain and model generations from PAI-Bench-G.} \textit{Best viewed with zoom.}
    }
    \label{fig:predict_examples_cs}
\end{figure*}

\begin{figure*}[p]

    \begin{tcolorbox}[colback=white, colframe=black, arc=4mm, boxrule=0.7pt, width=0.98\linewidth, left=3pt, right=3pt, top=3pt, bottom=3pt]
        \centering
        \begin{minipage}{0.98\linewidth}

            \centering
            \begin{subfigure}{\linewidth}
                \begin{minipage}[t]{0.24\linewidth}
                    \centering
                    \vspace{0pt}
                    \includegraphics[width=\linewidth]{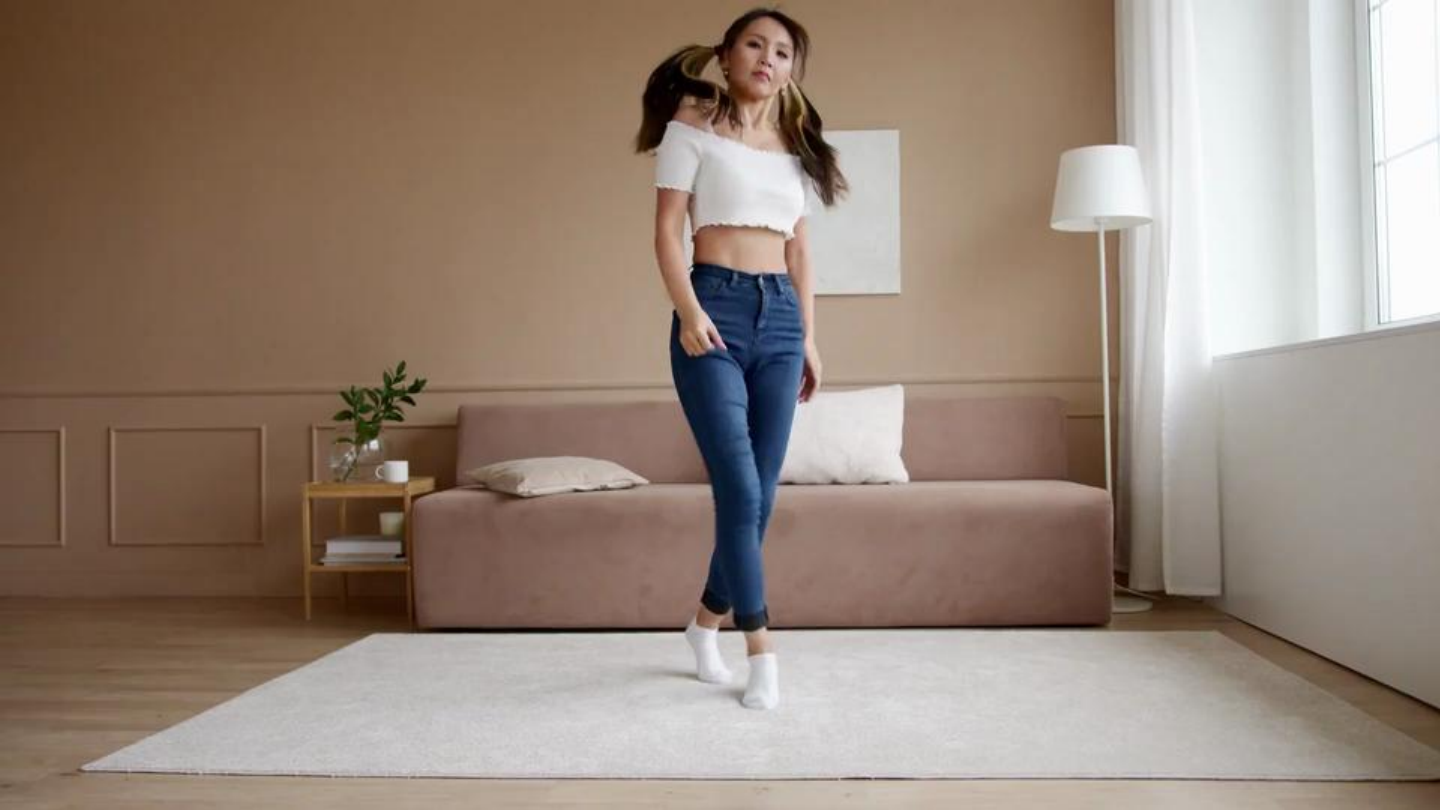}
                \end{minipage}%
                \hfill
                \begin{minipage}[t]{0.72\linewidth}
                    \vspace{0pt}
                    \scriptsize
                    A modern, minimalist living room serves as the backdrop for a dynamic dance performance. A woman, positioned centrally on a light beige rug, executes a series of graceful dance moves. She wears a white cropped top, high-waisted blue jeans, and white sneakers, with her long hair tied up into two low pig tails that flow freely as she dances. \textcolor{gray}{[TRUNCATED]} The video begins with the woman in a poised stance, her right leg crossed over her left, with both her arms to her sides. As she starts her dance routine, \textcolor{red}{she lifts her right leg high into the air}, balancing on her left foot, and \textcolor{red}{extends her right arm gracefully above her head}. She \textcolor{red}{shifts her weight back down}, landing smoothly and continuing with fluid, rhythmic movements that include \textcolor{red}{hip sways and arm gestures}. Her hair flows dynamically with her movements, adding a sense of energy and liveliness to the scene. By the final frame, the woman is caught mid-step in her dance routine, standing upright with her arms slightly bent in mid-air at the elbows. Her hair continues to flow, indicating the dynamic nature of her movements throughout the sequence. \textcolor{gray}{[TRUNCATED]}
                \end{minipage}
                \caption{Input condition signals}
            \end{subfigure}

            \centering

            \begin{subfigure}{\linewidth}
                \centering
                \includegraphics[width=\linewidth]{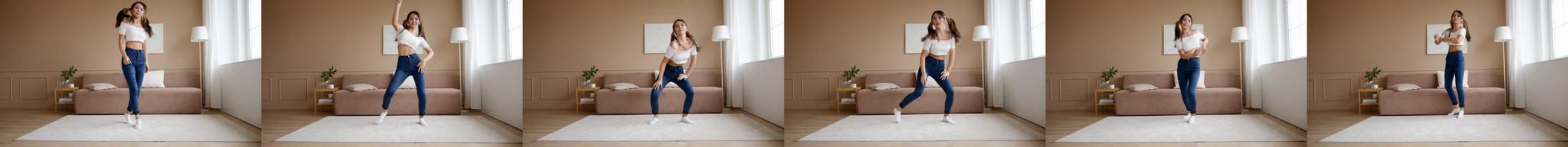}
                \caption{Source video}
            \end{subfigure}

            \begin{subfigure}{\linewidth}
                \centering
                \includegraphics[width=\linewidth]{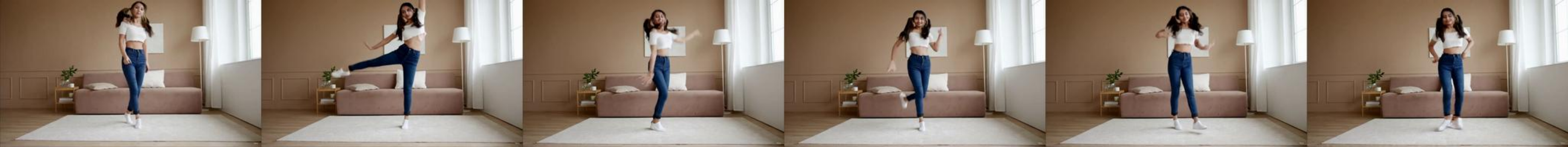}
                \caption{Veo3}
            \end{subfigure}

            \begin{subfigure}{\linewidth}
                \centering
                \includegraphics[width=\linewidth]{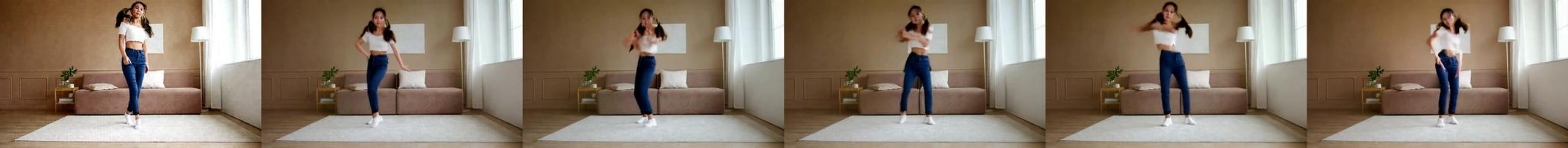}
                \caption{CogVideoX1.5}
            \end{subfigure}

            \begin{subfigure}{\linewidth}
                \centering
                \includegraphics[width=\linewidth]{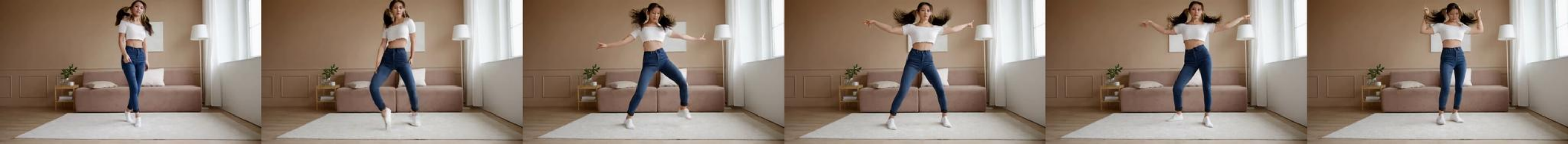}
                \caption{Cosmos-Predict2.5-2B}
            \end{subfigure}

            \begin{subfigure}{\linewidth}
                \centering
                \includegraphics[width=\linewidth]{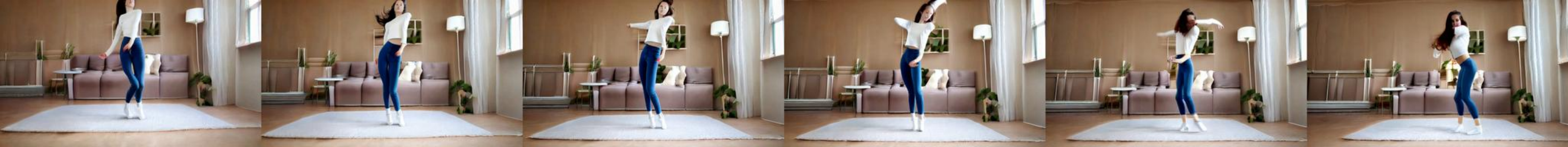}
                \caption{DynamicCrafter}
            \end{subfigure}

            \begin{subfigure}{\linewidth}
                \centering
                \includegraphics[width=\linewidth]{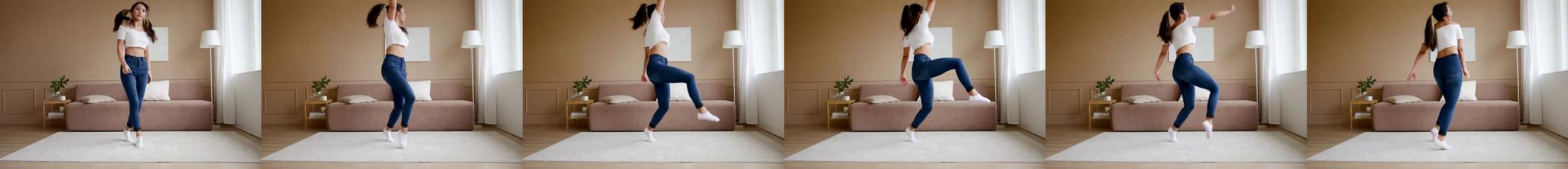}
                \caption{HunyuanVideo-I2V}
            \end{subfigure}

            \begin{subfigure}{\linewidth}
                \centering
                \includegraphics[width=\linewidth]{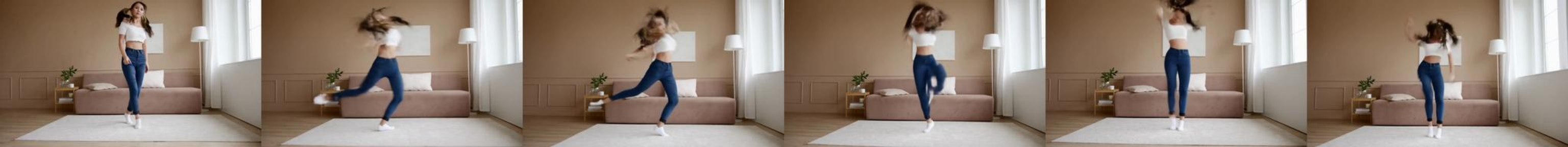}
                \caption{MAGI-1-24B}
            \end{subfigure}

            \begin{subfigure}{\linewidth}
                \centering
                \includegraphics[width=\linewidth]{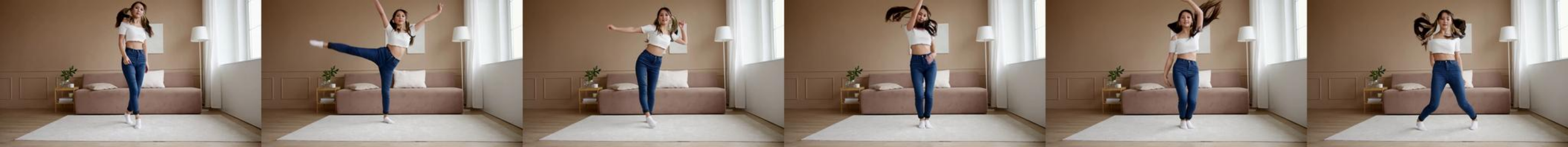}
                \caption{Wan2.2-I2V-A14B}
            \end{subfigure}

        \end{minipage}
    \end{tcolorbox}

    \caption{
        \textbf{Example of human domain and model generations from PAI-Bench-G.} \textit{Best viewed with zoom.}
    }
    \label{fig:predict_examples_human}
\end{figure*}

\begin{figure*}[p]

    \begin{tcolorbox}[colback=white, colframe=black, arc=4mm, boxrule=0.7pt, width=0.98\linewidth, left=3pt, right=3pt, top=3pt, bottom=3pt]
        \centering
        \begin{minipage}{0.98\linewidth}

            \centering
            \begin{subfigure}{\linewidth}
                \begin{minipage}[t]{0.24\linewidth}
                    \centering
                    \vspace{0pt}
                    \includegraphics[width=\linewidth]{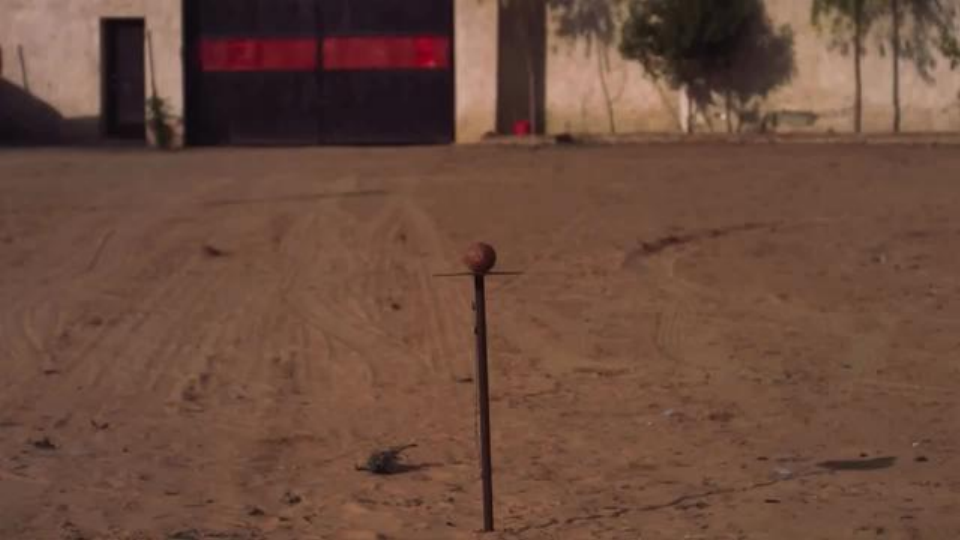}
                \end{minipage}%
                \hfill
                \begin{minipage}[t]{0.72\linewidth}
                    \vspace{0pt}
                    \scriptsize
                    A close-up shot captures a small, round object mounted on a metal pole in a sandy area. The background features a building with a red stripe across its facade and some trees to the right side. Tire tracks mark the ground, indicating recent vehicle activity. The object remains stationary until \textcolor{red}{it suddenly ignites, producing a burst of bright light and smoke}. The explosion is intense, \textcolor{red}{sending sparks flying outward and creating a large cloud of white smoke} that billows into the air. \textcolor{red}{The smoke gradually disperses, revealing scattered debris around the pole.} The scene remains static after the explosion, focusing on the remnants of the blast and the surrounding environment.
                \end{minipage}
                \caption{Input condition signals}
            \end{subfigure}

            \centering

            \begin{subfigure}{\linewidth}
                \centering
                \includegraphics[width=\linewidth]{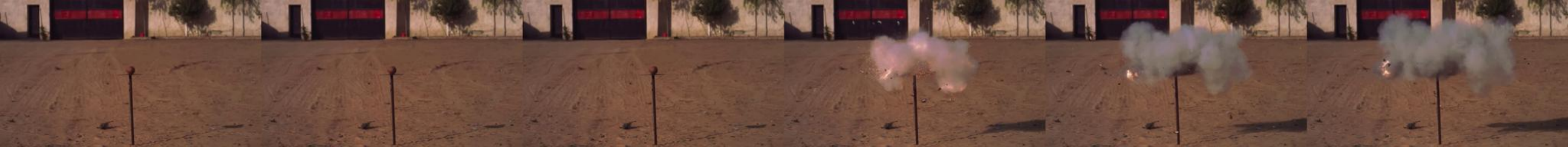}
                \caption{Source video}
            \end{subfigure}

            \begin{subfigure}{\linewidth}
                \centering
                \includegraphics[width=\linewidth]{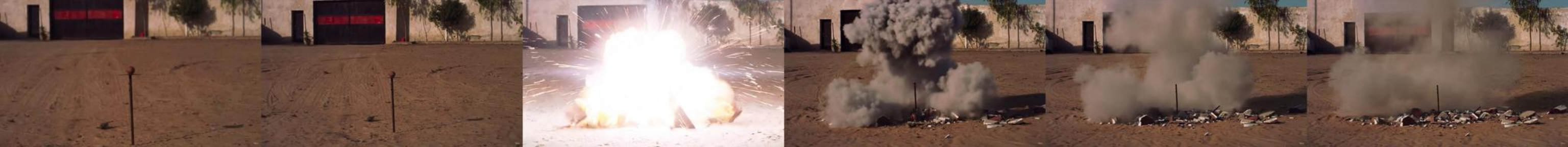}
                \caption{Veo3}
            \end{subfigure}

            \begin{subfigure}{\linewidth}
                \centering
                \includegraphics[width=\linewidth]{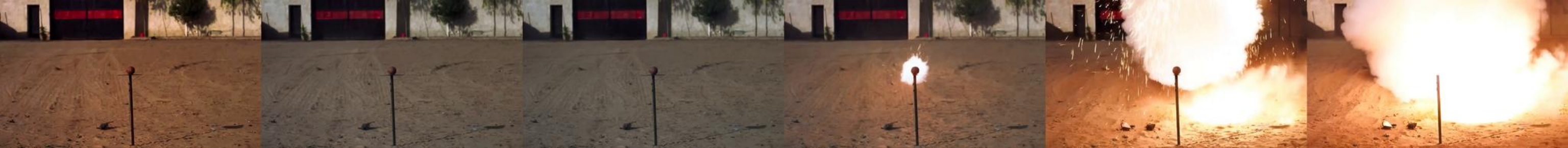}
                \caption{CogVideoX1.5}
            \end{subfigure}

            \begin{subfigure}{\linewidth}
                \centering
                \includegraphics[width=\linewidth]{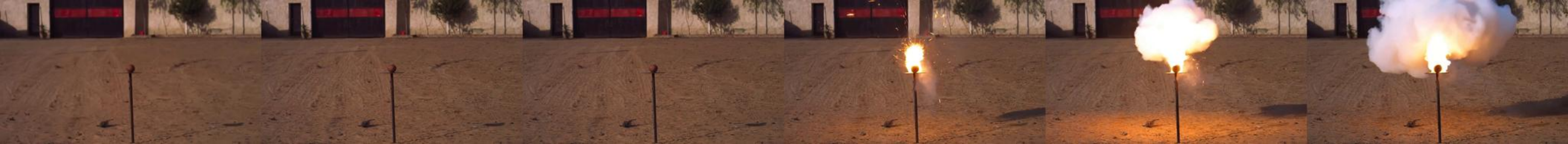}
                \caption{Cosmos-Predict2.5-2B}
            \end{subfigure}

            \begin{subfigure}{\linewidth}
                \centering
                \includegraphics[width=\linewidth]{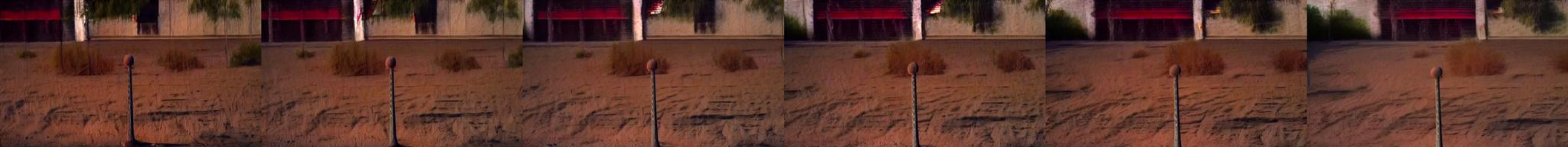}
                \caption{DynamicCrafter}
            \end{subfigure}

            \begin{subfigure}{\linewidth}
                \centering
                \includegraphics[width=\linewidth]{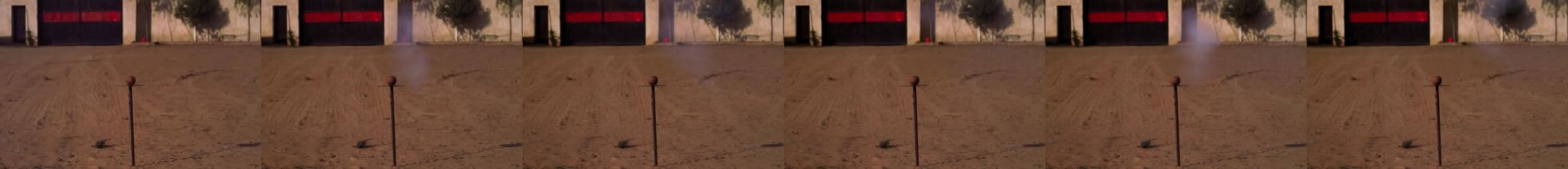}
                \caption{HunyuanVideo-I2V}
            \end{subfigure}

            \begin{subfigure}{\linewidth}
                \centering
                \includegraphics[width=\linewidth]{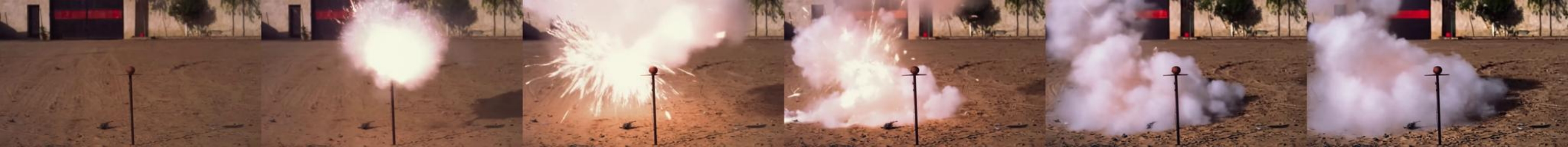}
                \caption{MAGI-1-24B}
            \end{subfigure}

            \begin{subfigure}{\linewidth}
                \centering
                \includegraphics[width=\linewidth]{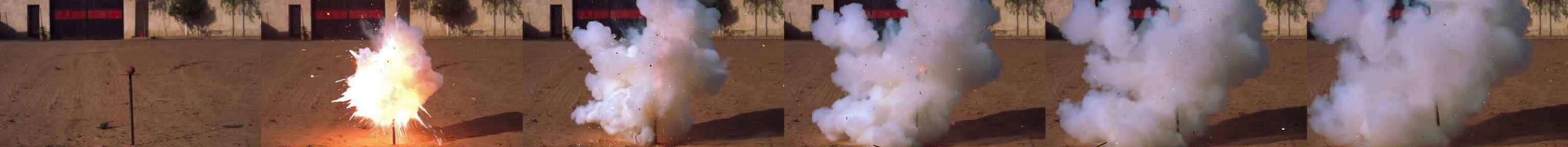}
                \caption{Wan2.2-I2V-A14B}
            \end{subfigure}

        \end{minipage}
    \end{tcolorbox}

    \caption{
        \textbf{Example of physics domain and model generations from PAI-Bench-G.} \textit{Best viewed with zoom.}
    }
    \label{fig:predict_examples_physics}
\end{figure*}

We present qualitative visualizations of generated samples to illustrate model performance across the diverse domains of PAI-Bench-G. Representative examples corresponding to autonomous vehicles, robotics, industrial settings, common sense reasoning, human activity, and physical dynamics are provided in assets~\ref{fig:predict_examples_av}, \ref{fig:predict_examples_robotics}, \ref{fig:predict_examples_industry}, \ref{fig:predict_examples_cs}, \ref{fig:predict_examples_human}, and \ref{fig:predict_examples_physics}, respectively.

\begin{algorithm}[t]
\caption{PAI-Bench-C Evaluation Pipeline}
\label{alg:pai_bench_c}
\small
\begin{algorithmic}[1]

\Require
    \Statex Conditional video $X$, generated video $\hat{X}$, \# prompts $K$.
    \Statex \textbf{Modalities} $\mathcal{M}=\{\text{Blur, Edge, Depth, Seg}\}$.
    \Statex \textbf{Operators} defined as pairs $(E_m, F_m)$ for $m \in \mathcal{M}$:
    \Statex \quad $\bullet$ \textbf{Blur}: Kernel $\rightarrow$ SSIM
    \Statex \quad $\bullet$ \textbf{Edge}: Canny $\rightarrow$ F1 Score
    \Statex \quad $\bullet$ \textbf{Depth}: VideoDepthAnything $\rightarrow$ si-RMSE
    \Statex \quad $\bullet$ \textbf{Seg}: GroundingDINO+SAM2 $\rightarrow$ mIoU

\vspace{1em}
\hrule
\vspace{1em}

\For{each modality $m \in \mathcal{M}$} \Comment{Stage 1: Condition Alignment}
    \State Extract features: $X_m \leftarrow E_m(X)$; \quad $\hat{X}_m \leftarrow E_m(\hat{X})$
    \State Compute fidelity: $s_m \leftarrow F_m(X_m,\,\hat{X}_m)$
\EndFor
\State $S_{\text{Align}} \leftarrow \{s_{\text{Blur}}, s_{\text{Edge}}, s_{\text{Depth}}, s_{\text{Seg}}\}$

\vspace{1em}
\For{each condition $c$} \Comment{Stage 2: Generation Diversity}
    \State Generate $K$ samples $\{\hat{X}^{(1)},\ldots,\hat{X}^{(K)}\}$
    \State $\text{Div}_c \leftarrow \frac{2}{K(K-1)}\sum_{i<j} \mathrm{LPIPS}(\hat{X}^{(i)},\hat{X}^{(j)})$
\EndFor
\State $S_{\text{Div}} \leftarrow \frac{1}{|\mathcal{C}|}\sum_c \text{Div}_c$

\vspace{1em}
\State $S_{\text{Qual}} \leftarrow \mathrm{DOVER}(\hat{X})$ \Comment{Stage 3: Visual Quality}

\vspace{1em}
\Ensure $S_{\text{Align}}$, $S_{\text{Div}}$, $S_{\text{Qual}}$

\end{algorithmic}
\end{algorithm}

\section{PAI-Bench-C Score Calculation Details} \label{appen:pai-c-score}
To comprehensively evaluate conditional VGMs, we propose a metric suite assessing three critical dimensions: 1) control fidelity (faithfulness to input signals), 2) visual quality, and 3) generative diversity under identical conditions. A unified formulation of the extraction and fidelity operators, alongside the complete scoring pipeline is in Algorithm~\ref{alg:pai_bench_c}.

\subsection{Control Fidelity}
We quantify the alignment between generated videos ($V_{gen}$) and reference control signals ($V_{ref}$) by projecting both into shared representation spaces, specifically varying levels of abstraction including low-frequency structure, edges, depth geometry, and semantic segmentation.

\noindent\textbf{Vis Alignment:} We apply the same blurring operation to both $V_{gen}$ and $V_{ref}$. We then compute the Structural Similarity Index Measure (SSIM)~\cite{Wang2004ImageQA} between the blurred representations, averaging scores across the dataset.

\noindent\textbf{Edge Consistency:} We evaluate boundary alignment by extracting binary edge maps from $V_{gen}$ and $V_{ref}$ using the Canny edge detector. The alignment is measured via the classification F1 score~\cite{van1979information} on the pixel-wise binary maps.

\noindent \textbf{Geometric Fidelity:} To measure 3D geometric consistency, we extract depth maps using DepthAnythingV2~\cite{yang2024depthv2}. We calculate the scale-invariant Root Mean Squared Error (si-RMSE)~\cite{eigen2014depthmappredictionsingle} between depth estimations of $V_{gen}$ and $V_{ref}$.

\noindent \textbf{Semantic Alignment:} We assess semantic layout consistency using a segmentation pipeline powered by GroundingDINO~\cite{liu2023grounding} and SAM2~\cite{ravi2024sam2}. To mitigate redundancy from open-set detection, instance masks are aggregated by caption phrase to form class-level masks. We establish correspondence between $V_{gen}$ and $V_{ref}$ masks via an Intersection over Union (IoU) matching algorithm. Matches with an $\text{IoU} < 0.1$ are filtered out, and the final score is reported as the mean IoU (mIoU) over valid correspondences.

\subsection{Visual Quality}
We assess the aesthetic and technical fidelity of generated videos using DOVER-technical~\cite{wu2022fastvqa,wu2023dover,end2endvideoqualitytool}, a reference-free video quality assessment metric. We compute the mean score across the entire dataset. Higher values correspond to superior visual clarity and stability.

\subsection{Generation Diversity}

To quantify the diversity of model's outputs under identical conditioning, we employ the Learned Perceptual Image Patch Similarity (LPIPS) metric~\cite{zhang2018perceptual}. For a fixed condition, we generate a set of $N$ videos corresponding to $N$ distinct text prompts. We calculate the pairwise LPIPS distance between all $\frac{N(N-1)}{2}$ unique pairs within this set. The final \textit{Diversity-LPIPS} score is derived by averaging these pairwise distances across all samples in the dataset. Higher values indicate greater diversity and reduced mode collapse.

\subsection{Qualitative Visualization}

\begin{figure*}[p]
    \begin{tcolorbox}[colback=white, colframe=black, arc=4mm, boxrule=0.7pt, width=0.98\linewidth, left=3pt, right=3pt, top=3pt, bottom=3pt]
        \centering
        \begin{minipage}{0.98\linewidth}
            \centering
            \begin{tcolorbox}[colback=white, colframe=gray, arc=4mm, boxrule=0.7pt, width=0.98\linewidth, left=3pt, right=3pt, top=3pt, bottom=3pt]
                \begin{minipage}{\linewidth}
                    \centering
                    \begin{subfigure}{0.48\linewidth}
                        \centering
                        \includegraphics[width=\linewidth]{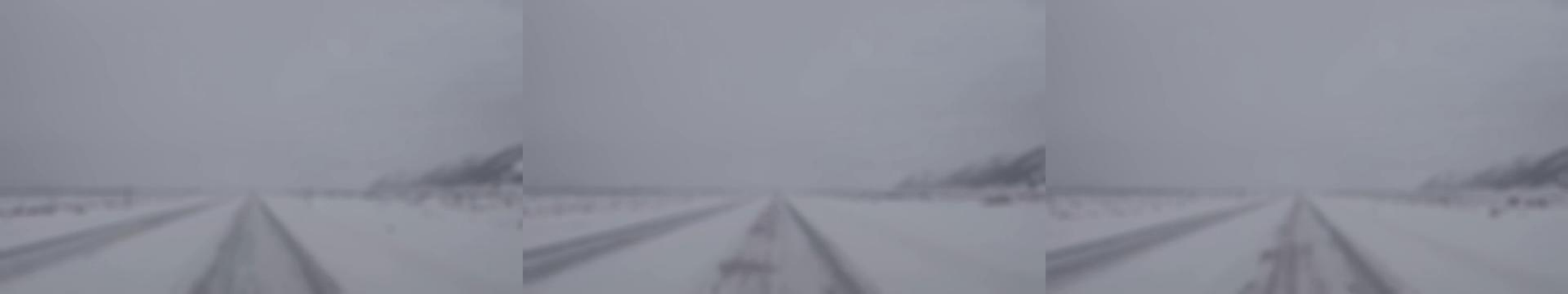}
                        \caption{Blur control signal (input)}
                    \end{subfigure}
                    \begin{subfigure}{0.48\linewidth}
                        \centering
                        \includegraphics[width=\linewidth]{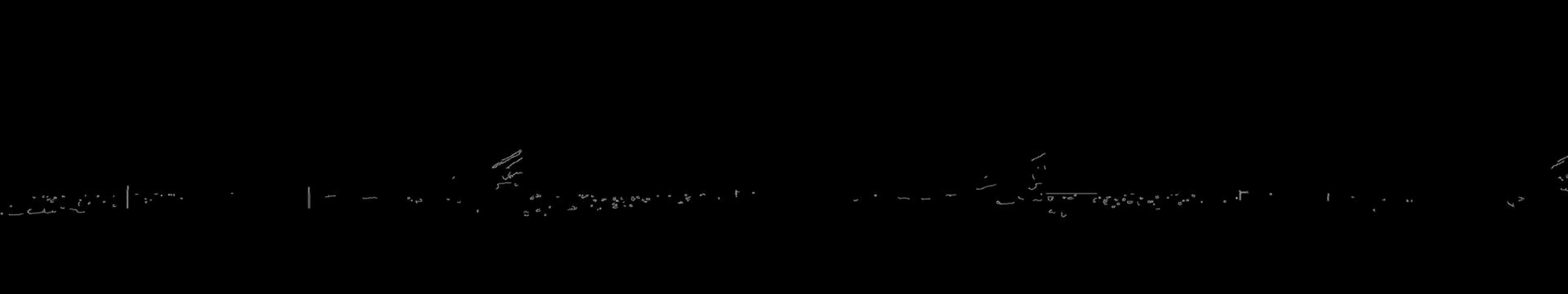}
                        \caption{Edge control signal (input)}
                    \end{subfigure}

                    \begin{subfigure}{0.48\linewidth}
                        \centering
                        \includegraphics[width=\linewidth]{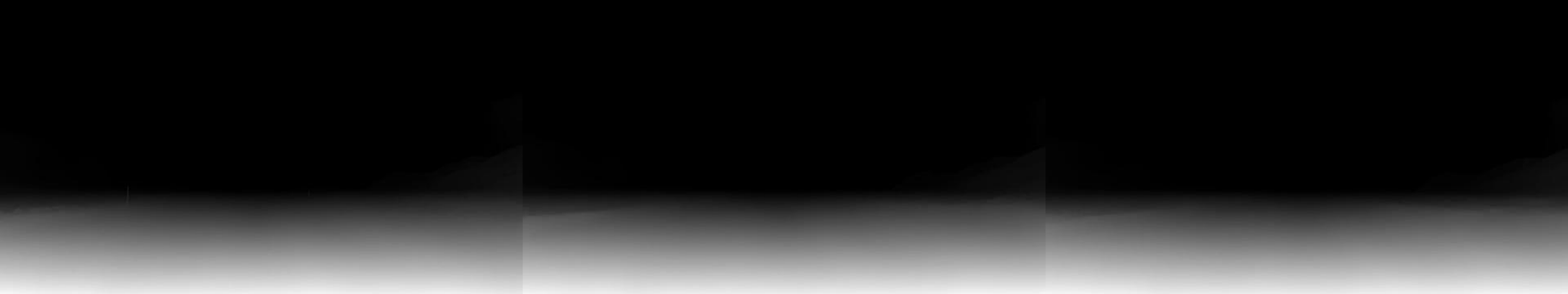}
                        \caption{Depth control signal (input)}
                    \end{subfigure}
                    \begin{subfigure}{0.48\linewidth}
                        \centering
                        \includegraphics[width=\linewidth]{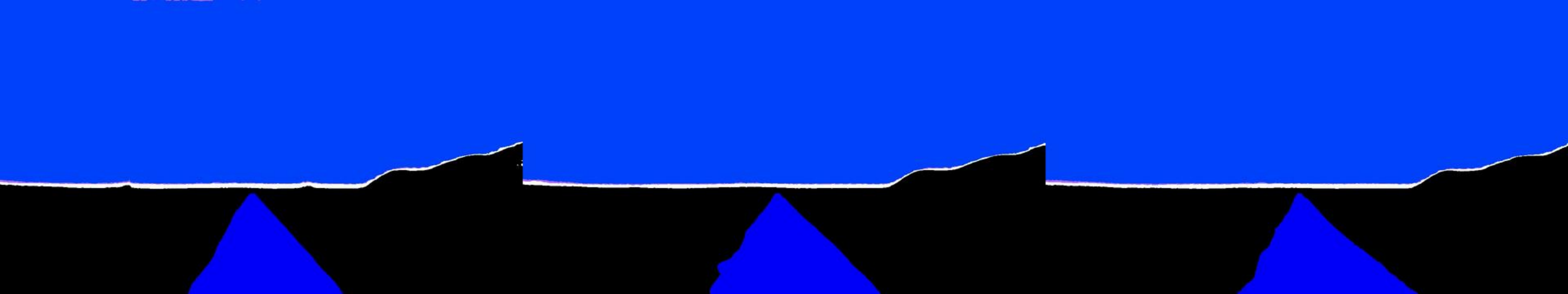}
                        \caption{Segmentation control signal (input)}
                    \end{subfigure}

                    \begin{subfigure}{\linewidth}
                        \centering
                        \begin{minipage}[t]{0.48\linewidth}
                            \centering
                            \vspace{0pt}
                            \includegraphics[width=\linewidth]{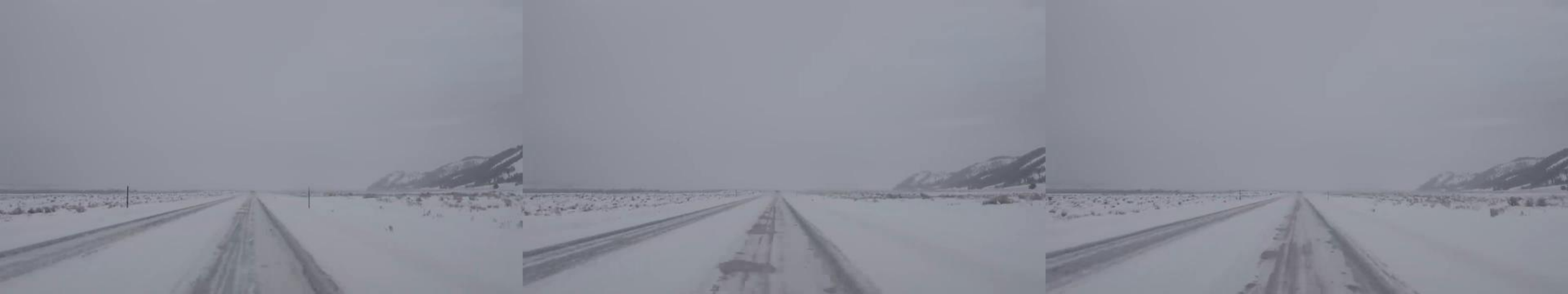}
                            \caption{Source video (reference)}
                        \end{minipage}
                        \begin{minipage}[t]{0.48\linewidth}
                            \vspace{0pt}
                            \scriptsize
                            	\textcolor{red}{A tranquil autumn landscape unfolds} throughout the video, showcasing a long, straight road stretching into the horizon. The road is covered in fallen leaves, with tire tracks visible, indicating recent vehicular passage. The surrounding terrain is flat and vast, blanketed in a mix of orange and brown leaves, creating a stark contrast against the clear blue sky. On either side of the road, sparse trees with bare branches add texture to the otherwise smooth surface. \textcolor{gray}{[TRUNCATED]}
                            \caption{Text control signal (input)}
                        \end{minipage}
                    \end{subfigure}
                \end{minipage}
            \end{tcolorbox}

            \begin{tcolorbox}[colback=white, colframe=gray, arc=4mm, boxrule=0.7pt, width=0.98\linewidth, left=3pt, right=3pt, top=3pt, bottom=3pt]
                \begin{minipage}{0.98\linewidth}
                    \centering

                    \begin{subfigure}{0.48\linewidth}
                        \centering
                        \includegraphics[width=\linewidth]{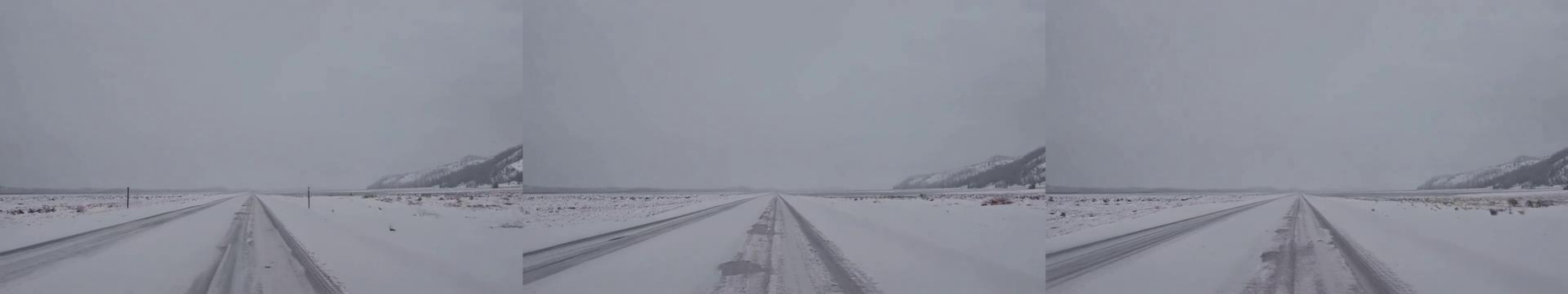}
                        \caption{Cosmos-Transfer2.5-2B (blur)}
                    \end{subfigure}
                    \begin{subfigure}{0.48\linewidth}
                        \centering
                        \includegraphics[width=\linewidth]{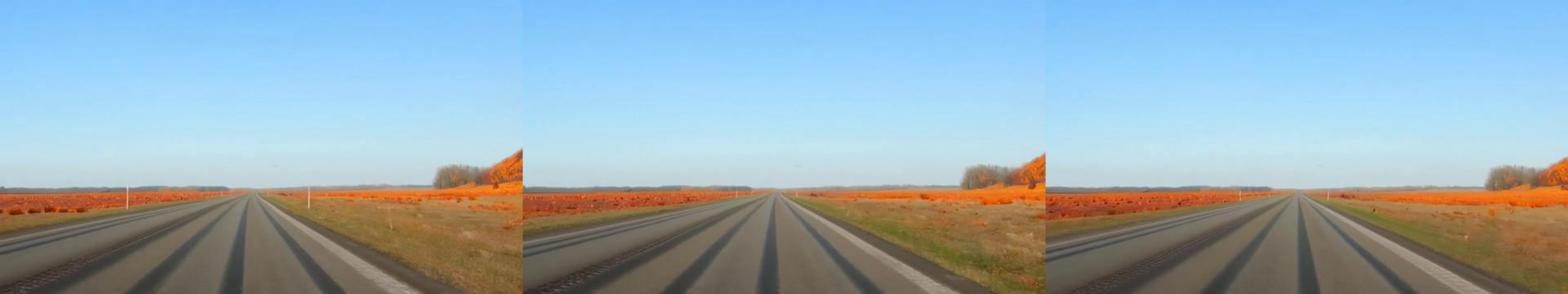}
                        \caption{Cosmos-Transfer2.5-2B (edge)}
                    \end{subfigure}

                    \begin{subfigure}{0.48\linewidth}
                        \centering
                        \includegraphics[width=\linewidth]{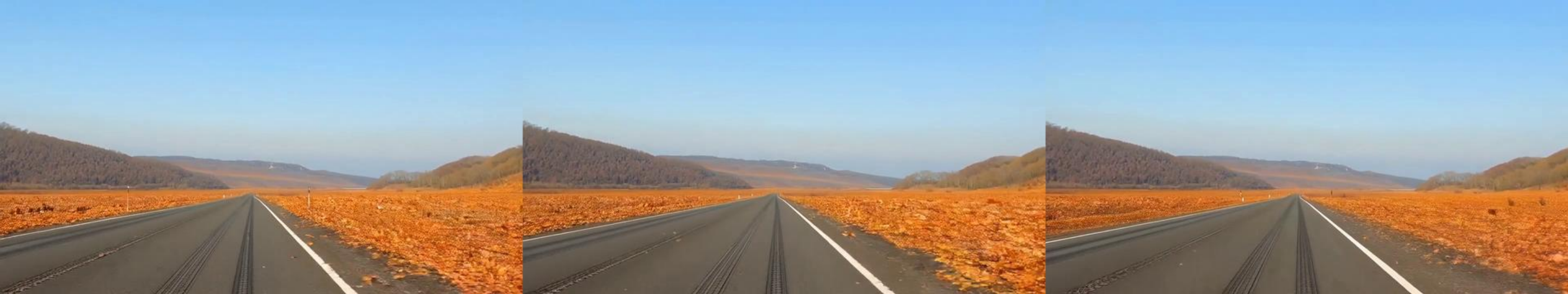}
                        \caption{Cosmos-Transfer2.5-2B (depth)}
                    \end{subfigure}
                    \begin{subfigure}{0.48\linewidth}
                        \centering
                        \includegraphics[width=\linewidth]{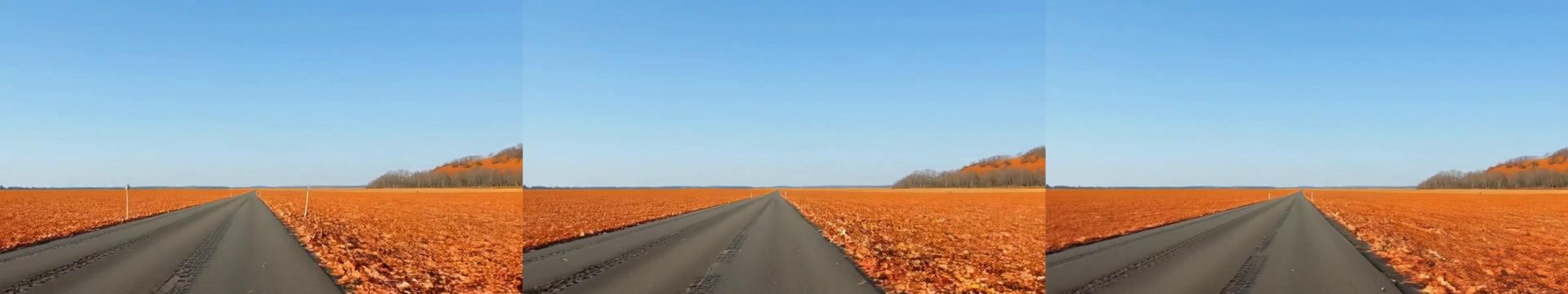}
                        \caption{Cosmos-Transfer2.5-2B (segmentation)}
                    \end{subfigure}

                    \begin{subfigure}{0.48\linewidth}
                        \centering
                        \includegraphics[width=\linewidth]{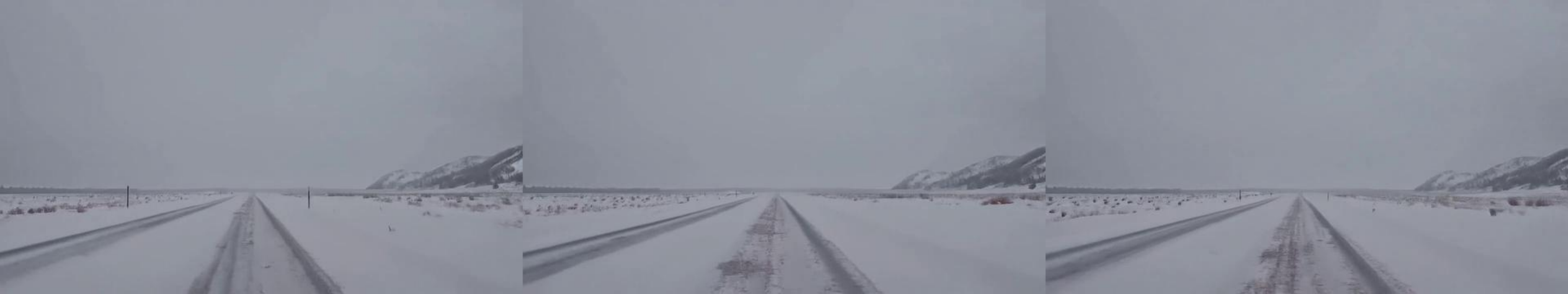}
                        \caption{Cosmos-Transfer2.5-2B (multi-control generation)}
                    \end{subfigure}

                \end{minipage}
            \end{tcolorbox}

            \begin{tcolorbox}[colback=white, colframe=gray, arc=4mm, boxrule=0.7pt, width=0.98\linewidth, left=3pt, right=3pt, top=3pt, bottom=3pt]
                \begin{minipage}{0.98\linewidth}
                    \centering

                    \begin{subfigure}{0.48\linewidth}
                        \centering
                        \includegraphics[width=\linewidth]{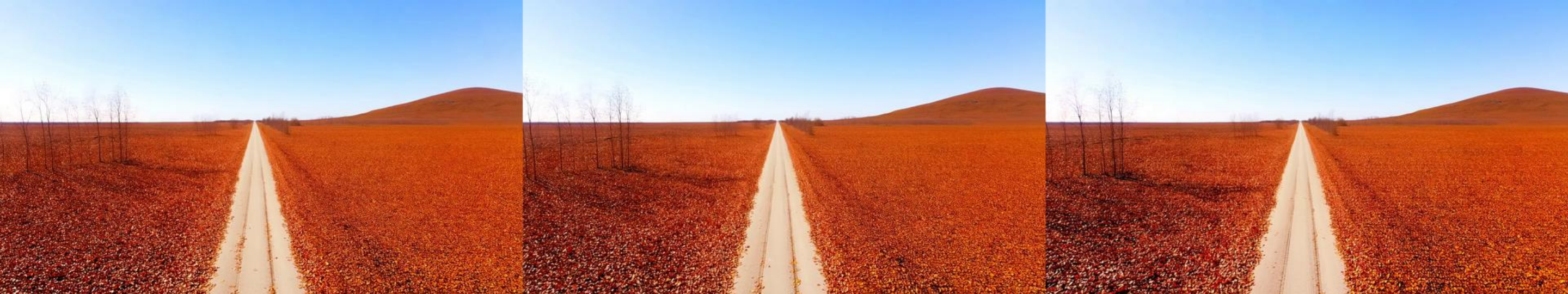}
                        \caption{Wan2.2-Fun-A14B-Control (blur)}
                    \end{subfigure}
                    \begin{subfigure}{0.48\linewidth}
                        \centering
                        \includegraphics[width=\linewidth]{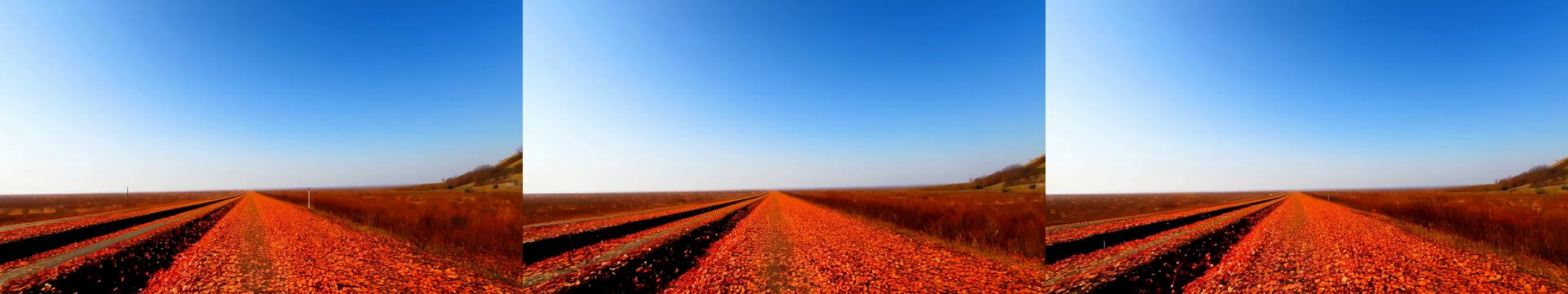}
                        \caption{Wan2.2-Fun-A14B-Control (edge)}
                    \end{subfigure}

                    \begin{subfigure}{0.48\linewidth}
                        \centering
                        \includegraphics[width=\linewidth]{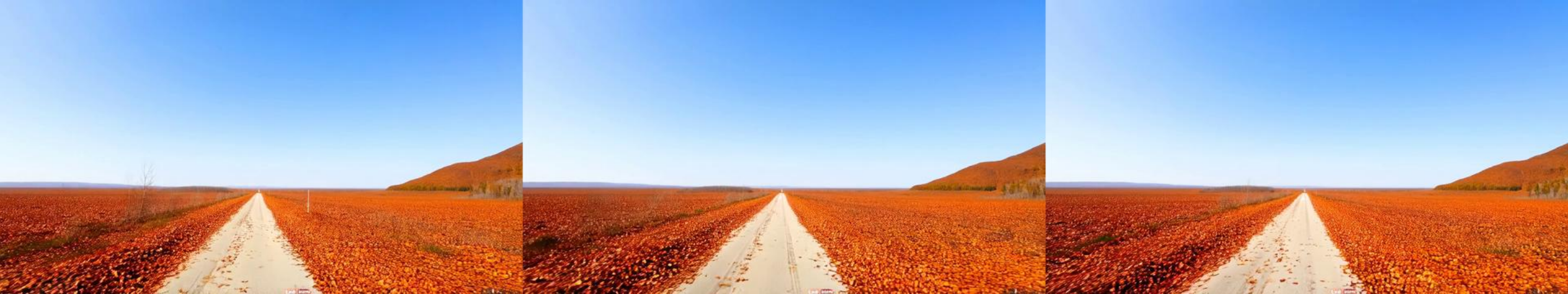}
                        \caption{Wan2.2-Fun-A14B-Control (depth)}
                    \end{subfigure}
                    \begin{subfigure}{0.48\linewidth}
                        \centering
                        \includegraphics[width=\linewidth]{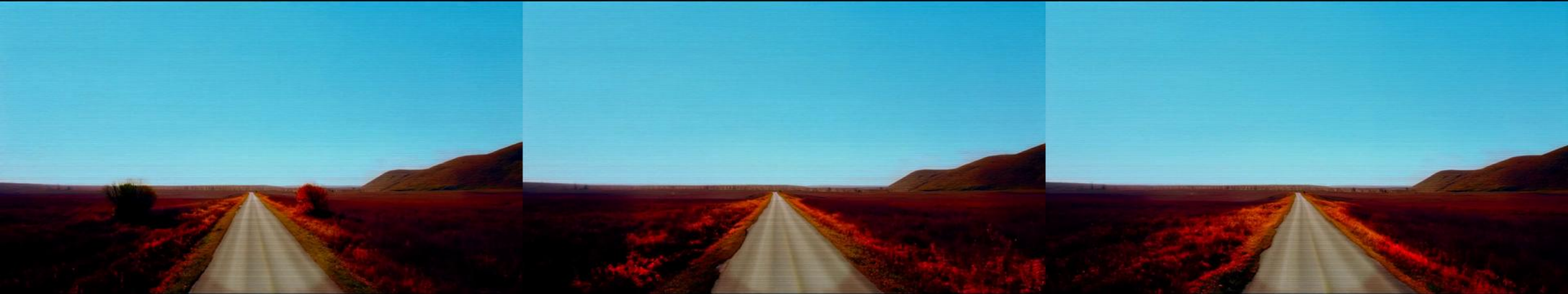}
                        \caption{Wan2.2-Fun-A14B-Control (segmentation)}
                    \end{subfigure}

                \end{minipage}
            \end{tcolorbox}

        \end{minipage}
    \end{tcolorbox}

    \caption{
        \textbf{Example of autonomous vehicle domain control signals and model generations from PAI-Bench-C.} \textit{Best viewed with zoom.}
    }
    \label{fig:transfer_examples_av}
\end{figure*}

\begin{figure*}[p]
    \begin{tcolorbox}[colback=white, colframe=black, arc=4mm, boxrule=0.7pt, width=0.98\linewidth, left=3pt, right=3pt, top=3pt, bottom=3pt]
        \centering
        \begin{minipage}{0.98\linewidth}
            \centering
            \begin{tcolorbox}[colback=white, colframe=gray, arc=4mm, boxrule=0.7pt, width=0.98\linewidth, left=3pt, right=3pt, top=3pt, bottom=3pt]
                \begin{minipage}{\linewidth}
                    \centering
                    \begin{subfigure}{0.48\linewidth}
                        \centering
                        \includegraphics[width=\linewidth]{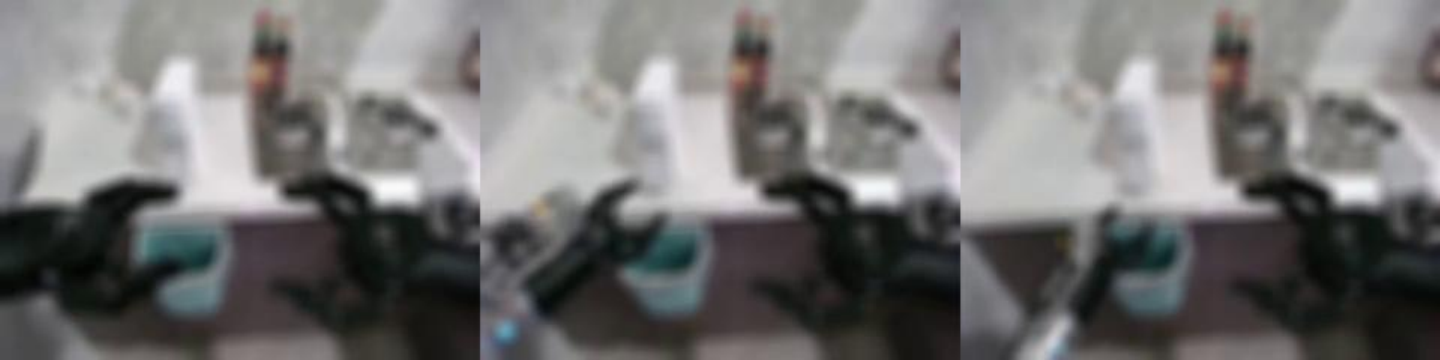}
                        \caption{Blur control signal (input)}
                    \end{subfigure}
                    \begin{subfigure}{0.48\linewidth}
                        \centering
                        \includegraphics[width=\linewidth]{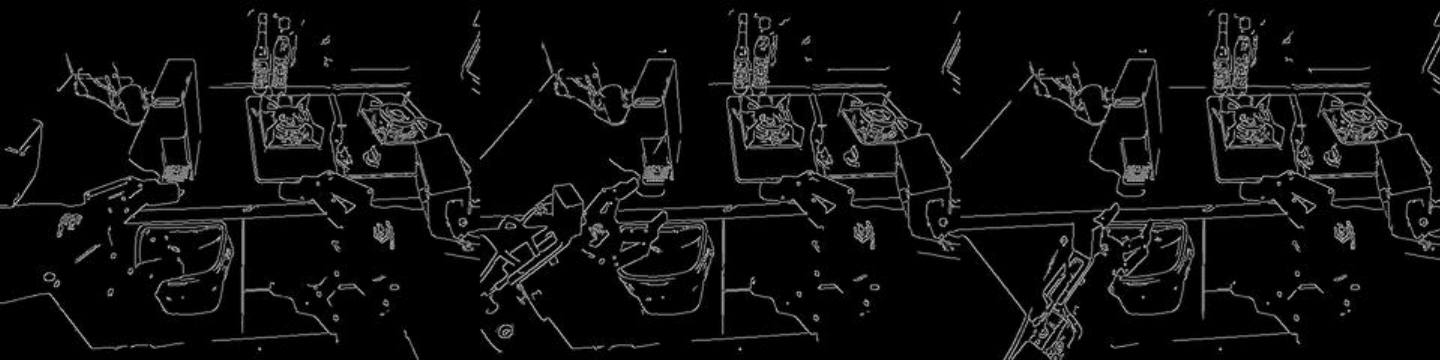}
                        \caption{Edge control signal (input)}
                    \end{subfigure}

                    \begin{subfigure}{0.48\linewidth}
                        \centering
                        \includegraphics[width=\linewidth]{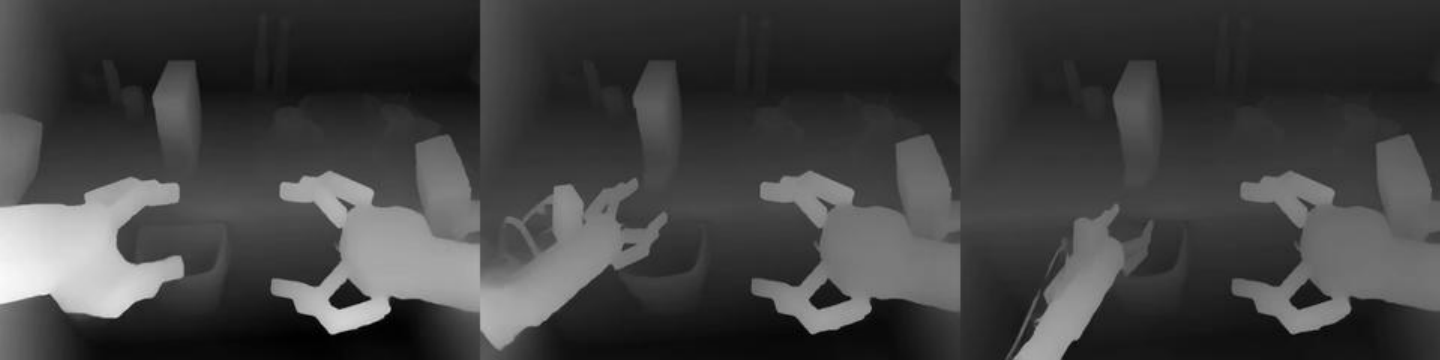}
                        \caption{Depth control signal (input)}
                    \end{subfigure}
                    \begin{subfigure}{0.48\linewidth}
                        \centering
                        \includegraphics[width=\linewidth]{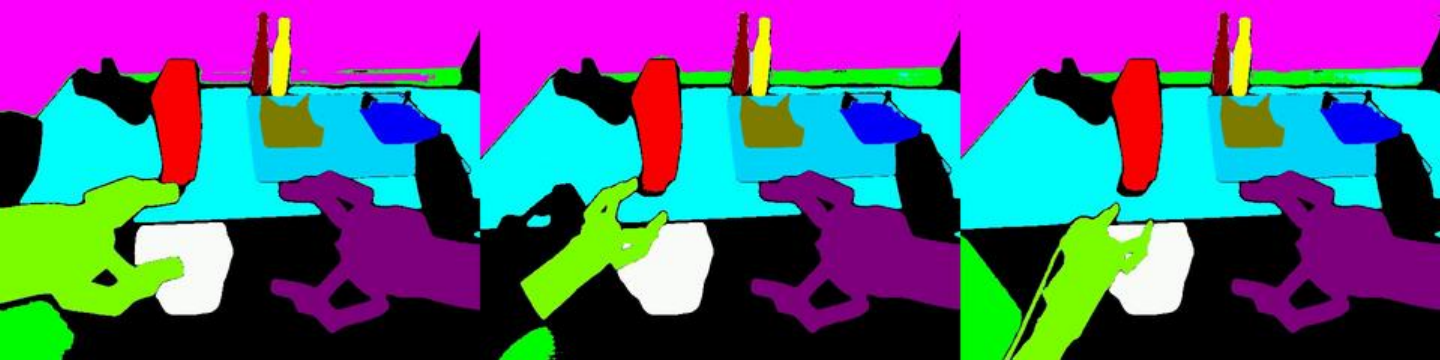}
                        \caption{Segmentation control signal (input)}
                    \end{subfigure}

                    \begin{subfigure}{\linewidth}
                        \centering
                        \begin{minipage}[t]{0.48\linewidth}
                            \centering
                            \vspace{0pt}
                            \includegraphics[width=\linewidth]{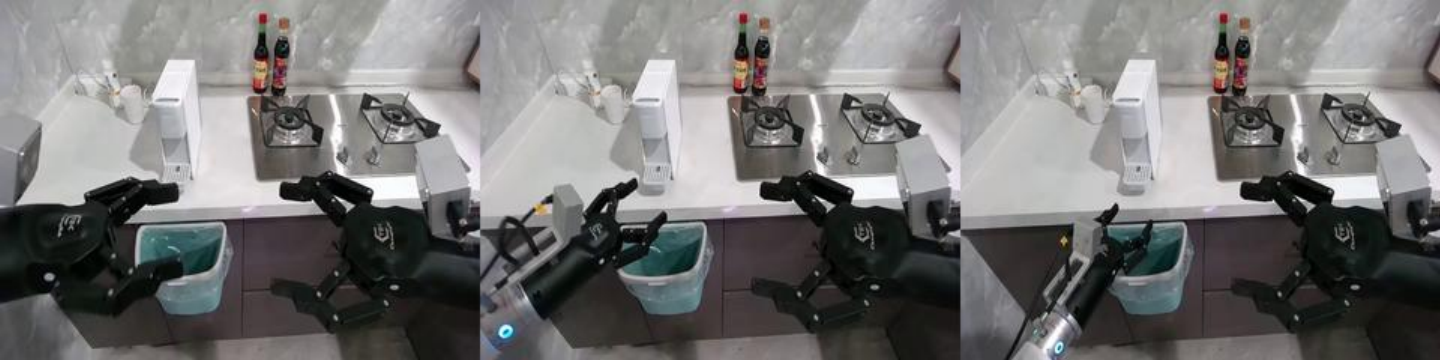}
                            \caption{Source video (reference)}
                        \end{minipage}
                        \begin{minipage}[t]{0.48\linewidth}
                            \vspace{0pt}
                            \scriptsize
                            In this video, robotic arms operate in \textcolor{red}{a contemporary kitchen featuring a wooden countertop and a glass-top gas stove} with two burners. Three ceramic jars are aligned on the counter. \textcolor{red}{A black appliance}, likely a water filter, is placed to the left. \textcolor{red}{The wall has a textured stone appearance}, adding to the kitchen's stylish design.
                            \caption{Text control signal (input)}
                        \end{minipage}
                    \end{subfigure}
                \end{minipage}
            \end{tcolorbox}

            \begin{tcolorbox}[colback=white, colframe=gray, arc=4mm, boxrule=0.7pt, width=0.98\linewidth, left=3pt, right=3pt, top=3pt, bottom=3pt]
                \begin{minipage}{\linewidth}
                    \centering
                    \begin{subfigure}{0.48\linewidth}
                        \centering
                        \includegraphics[width=\linewidth]{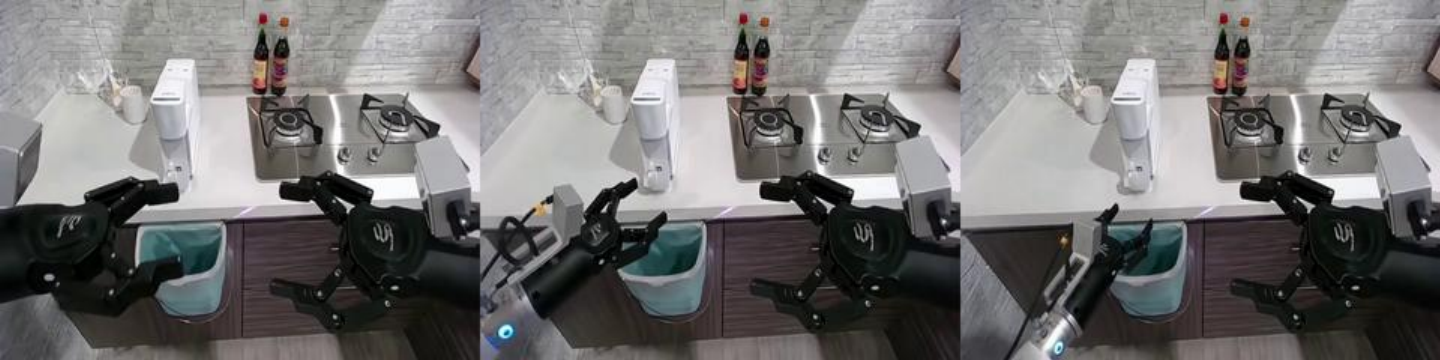}
                        \caption{Cosmos-Transfer2.5-2B (blur)}
                    \end{subfigure}
                    \begin{subfigure}{0.48\linewidth}
                        \centering
                        \includegraphics[width=\linewidth]{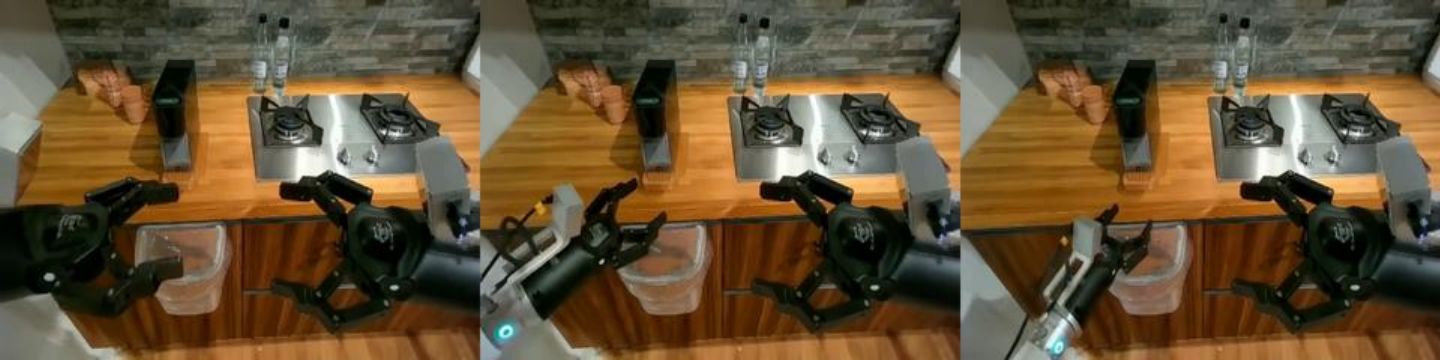}
                        \caption{Cosmos-Transfer2.5-2B (edge)}
                    \end{subfigure}

                    \begin{subfigure}{0.48\linewidth}
                        \centering
                        \includegraphics[width=\linewidth]{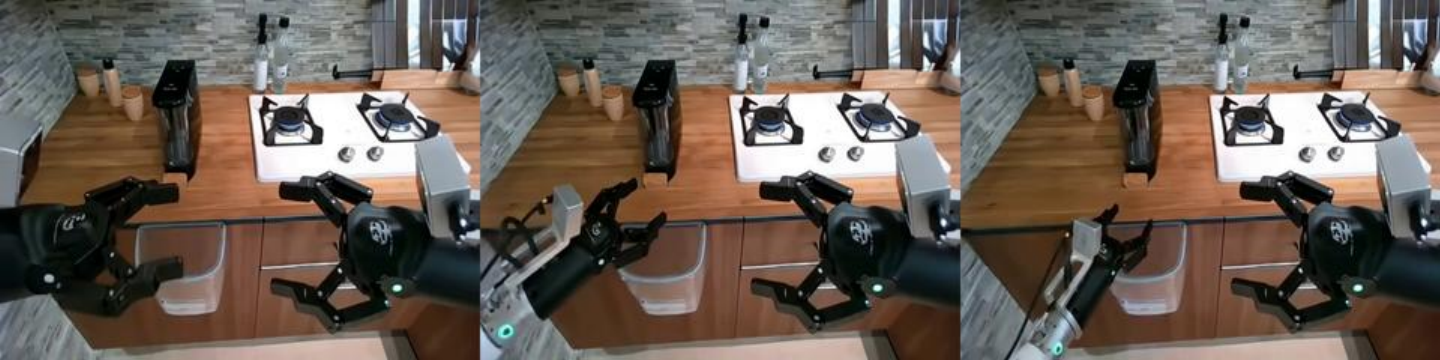}
                        \caption{Cosmos-Transfer2.5-2B (depth)}
                    \end{subfigure}
                    \begin{subfigure}{0.48\linewidth}
                        \centering
                        \includegraphics[width=\linewidth]{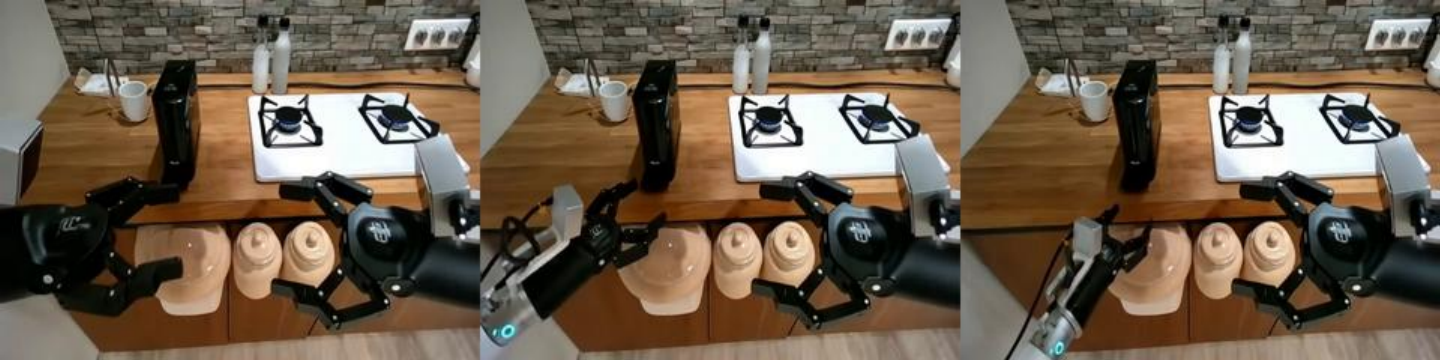}
                        \caption{Cosmos-Transfer2.5-2B (segmentation)}
                    \end{subfigure}

                    \begin{subfigure}{0.48\linewidth}
                        \centering
                        \includegraphics[width=\linewidth]{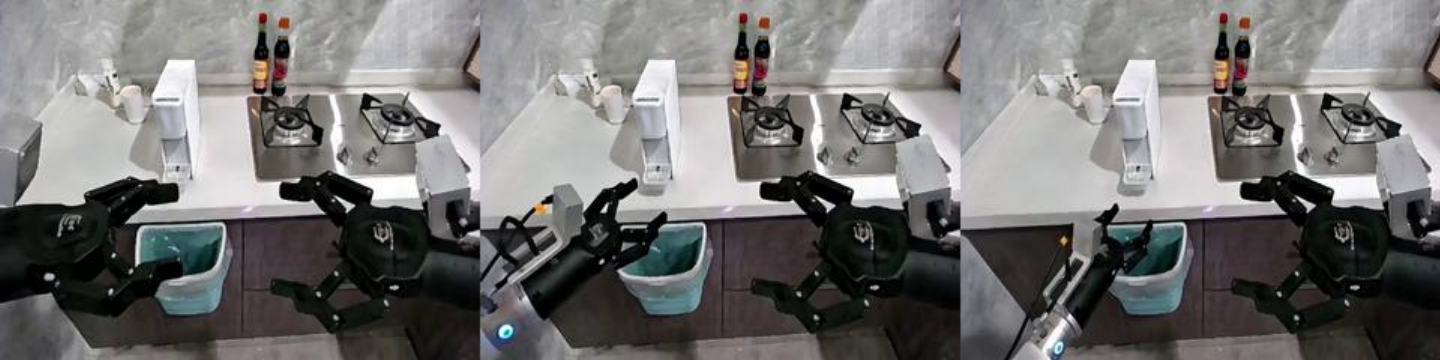}
                        \caption{Cosmos-Transfer2.5-2B (multi-control generation)}
                    \end{subfigure}
                \end{minipage}
            \end{tcolorbox}

            \begin{tcolorbox}[colback=white, colframe=gray, arc=4mm, boxrule=0.7pt, width=0.98\linewidth, left=3pt, right=3pt, top=3pt, bottom=3pt]
                \begin{minipage}{\linewidth}
                    \centering
                    \begin{subfigure}{0.48\linewidth}
                        \centering
                        \includegraphics[width=\linewidth]{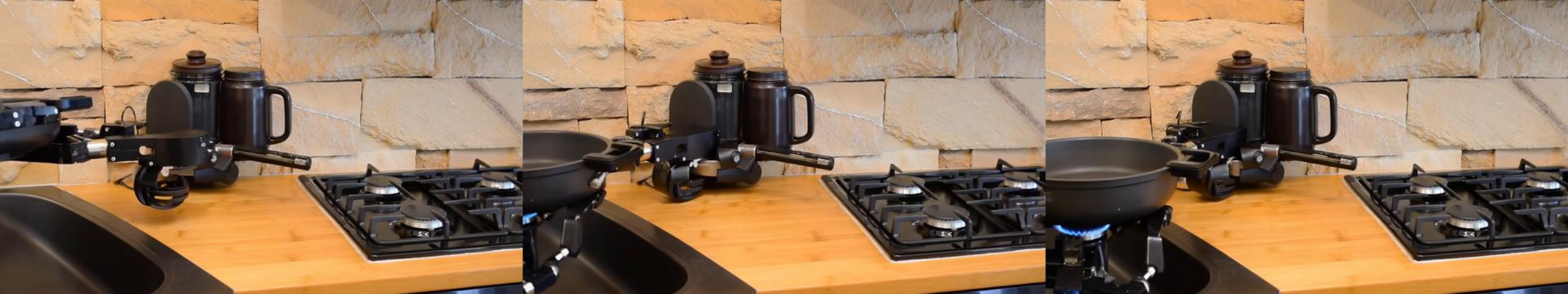}
                        \caption{Wan2.2-Fun-A14B-Control (blur)}
                    \end{subfigure}
                    \begin{subfigure}{0.48\linewidth}
                        \centering
                        \includegraphics[width=\linewidth]{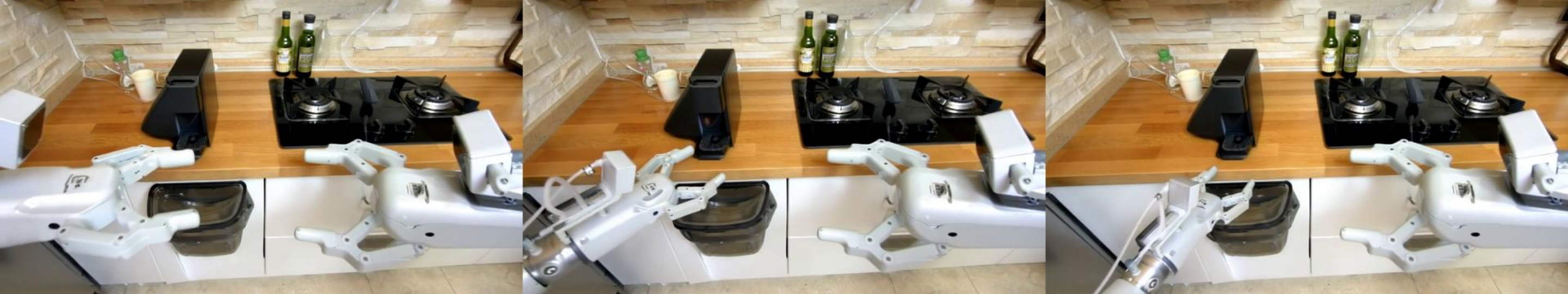}
                        \caption{Wan2.2-Fun-A14B-Control (edge)}
                    \end{subfigure}

                    \begin{subfigure}{0.48\linewidth}
                        \centering
                        \includegraphics[width=\linewidth]{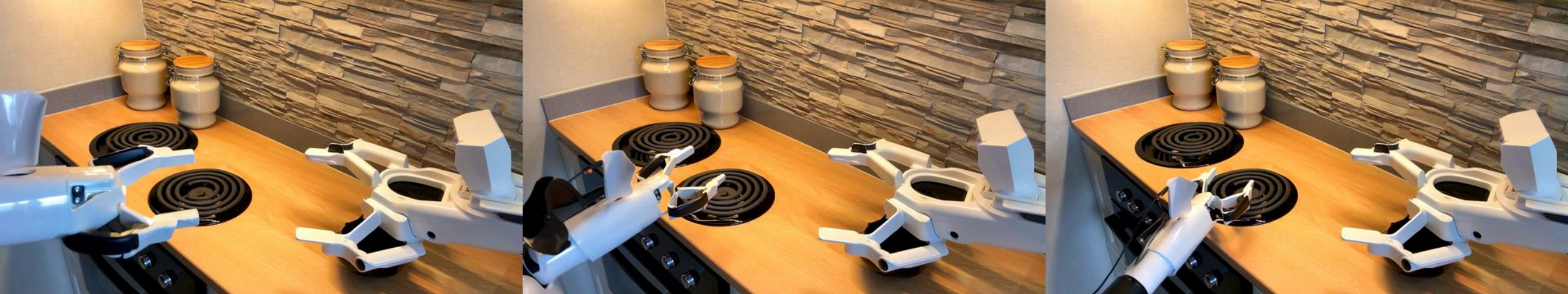}
                        \caption{Wan2.2-Fun-A14B-Control (depth)}
                    \end{subfigure}
                    \begin{subfigure}{0.48\linewidth}
                        \centering
                        \includegraphics[width=\linewidth]{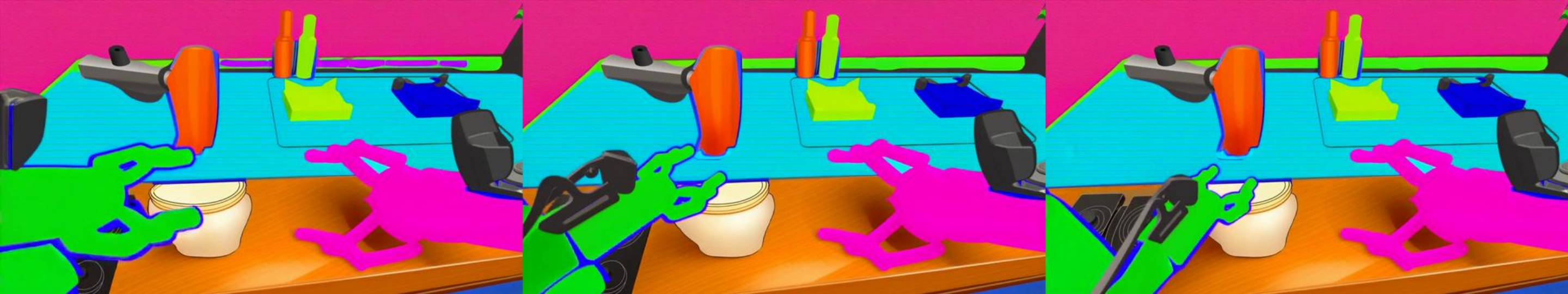}
                        \caption{Wan2.2-Fun-A14B-Control (segmentation)}
                    \end{subfigure}
                \end{minipage}
            \end{tcolorbox}
        \end{minipage}
    \end{tcolorbox}
    \caption{\textbf{Example of robotics domain control signals and model generations from PAI-Bench-C.} \textit{Best viewed with zoom.}}
    \label{fig:transfer_examples_robotics}
\end{figure*}

\begin{figure*}[p]
    \begin{tcolorbox}[colback=white, colframe=black, arc=4mm, boxrule=0.7pt, width=0.98\linewidth, left=3pt, right=3pt, top=3pt, bottom=3pt]
        \centering
        \begin{minipage}{0.98\linewidth}
            \centering
            \begin{tcolorbox}[colback=white, colframe=gray, arc=4mm, boxrule=0.7pt, width=0.98\linewidth, left=3pt, right=3pt, top=3pt, bottom=3pt]
                \begin{minipage}{\linewidth}
                    \centering
                    \begin{subfigure}{0.48\linewidth}
                        \centering
                        \includegraphics[width=\linewidth]{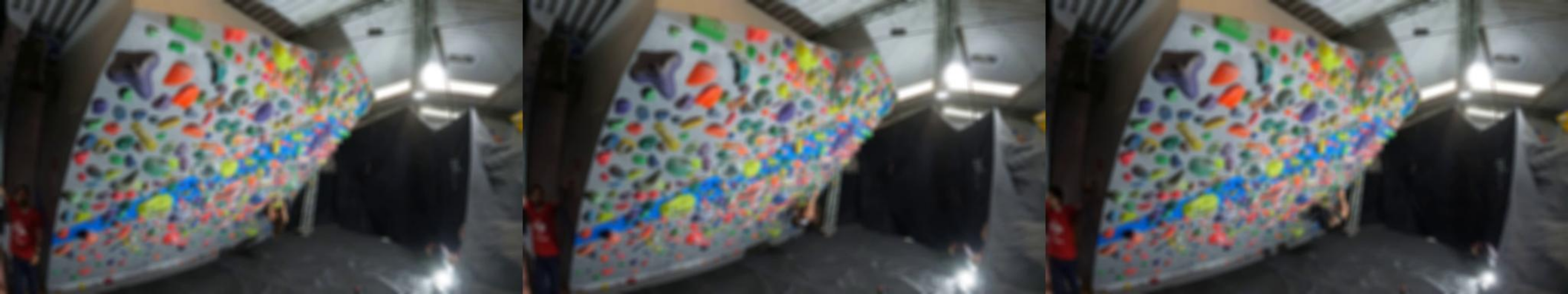}
                        \caption{Blur control signal (input)}
                    \end{subfigure}
                    \begin{subfigure}{0.48\linewidth}
                        \centering
                        \includegraphics[width=\linewidth]{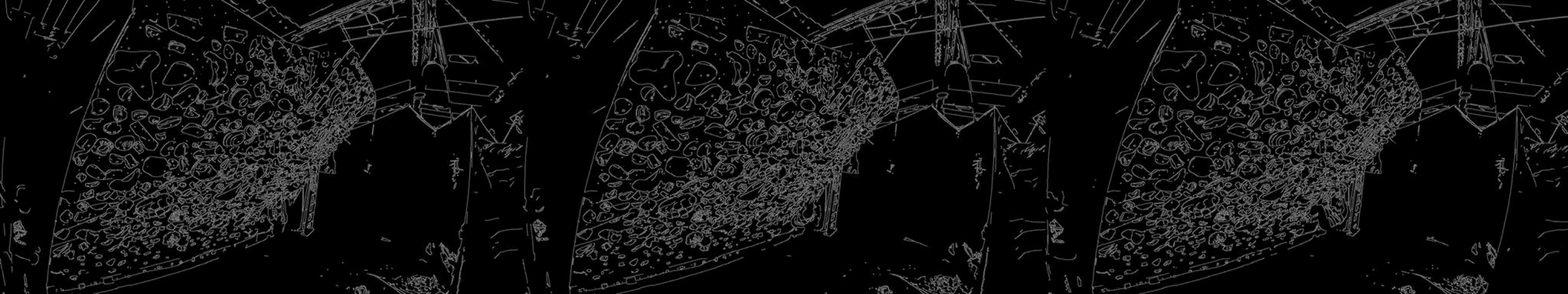}
                        \caption{Edge control signal (input)}
                    \end{subfigure}

                    \begin{subfigure}{0.48\linewidth}
                        \centering
                        \includegraphics[width=\linewidth]{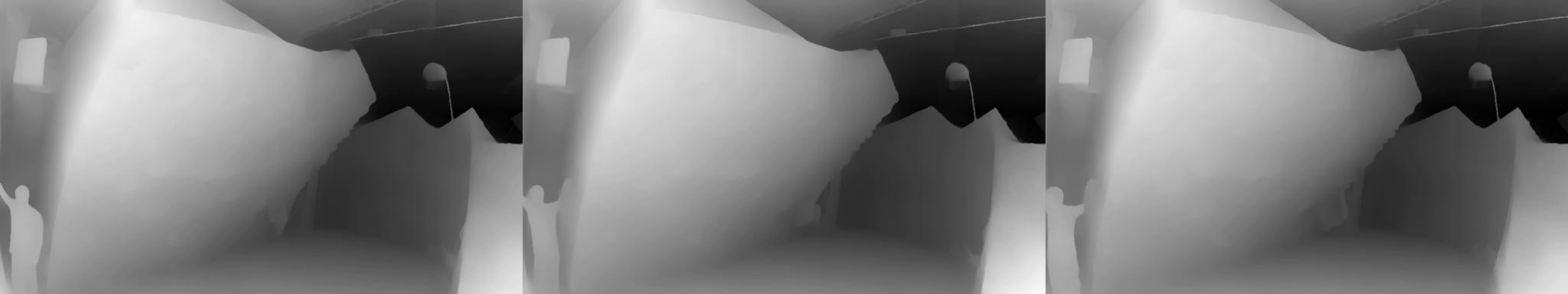}
                        \caption{Depth control signal (input)}
                    \end{subfigure}
                    \begin{subfigure}{0.48\linewidth}
                        \centering
                        \includegraphics[width=\linewidth]{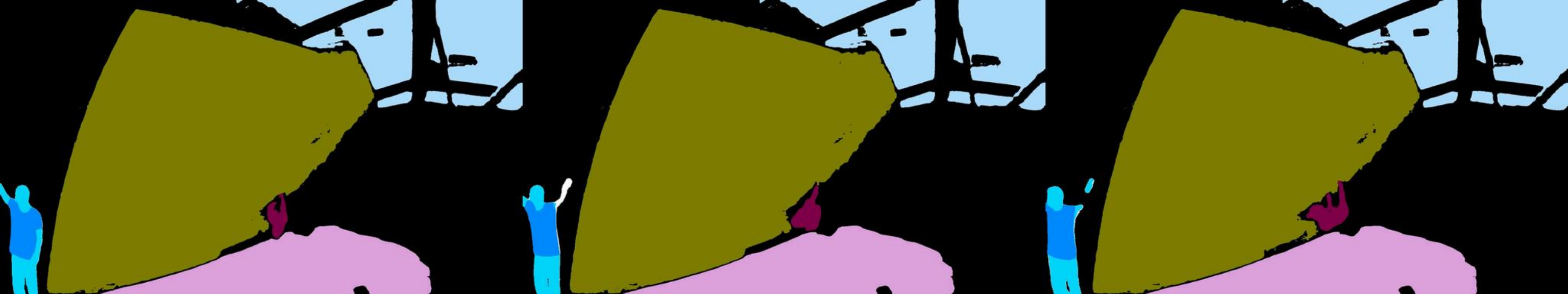}
                        \caption{Segmentation control signal (input)}
                    \end{subfigure}

                    \begin{subfigure}{\linewidth}
                        \centering
                        \begin{minipage}[t]{0.48\linewidth}
                            \centering
                            \vspace{0pt}
                            \includegraphics[width=\linewidth]{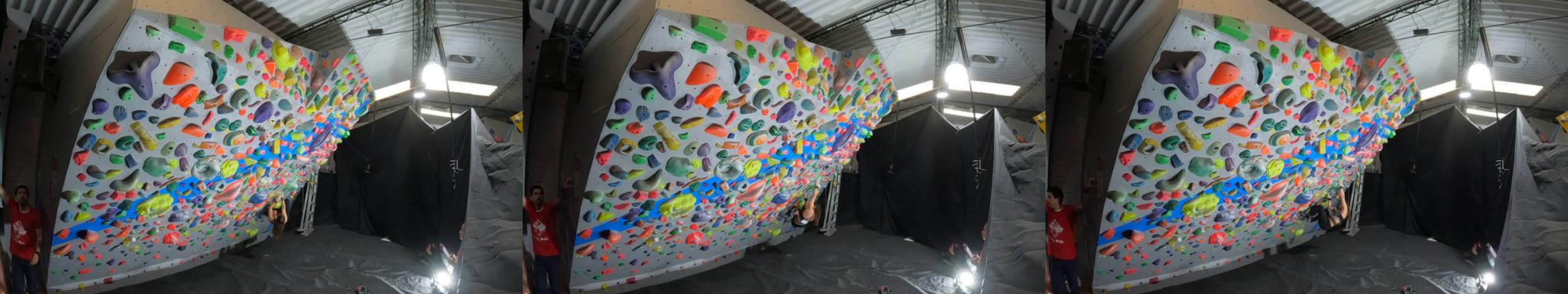}
                            \caption{Source video (reference)}
                        \end{minipage}
                        \begin{minipage}[t]{0.48\linewidth}
                            \vspace{0pt}
                            \scriptsize
                            The video is set in an indoor climbing gym, featuring a large, overhanging bouldering wall covered with \textcolor{red}{a variety of neon-colored climbing holds}. The holds are arranged in a complex pattern, with colors including \textcolor{red}{electric blue, neon pink, lime green, bright yellow, and vivid orange}, creating a visually dynamic backdrop. The wall is angled steeply, challenging the climber with its overhang. \textcolor{gray}{[TRUNCATED]}
                            \caption{Text control signal (input)}
                        \end{minipage}
                    \end{subfigure}
                \end{minipage}
            \end{tcolorbox}

            \begin{tcolorbox}[colback=white, colframe=gray, arc=4mm, boxrule=0.7pt, width=0.98\linewidth, left=3pt, right=3pt, top=3pt, bottom=3pt]
                \begin{minipage}{0.98\linewidth}
                    \centering
                    \begin{subfigure}{0.48\linewidth}
                        \centering
                        \includegraphics[width=\linewidth]{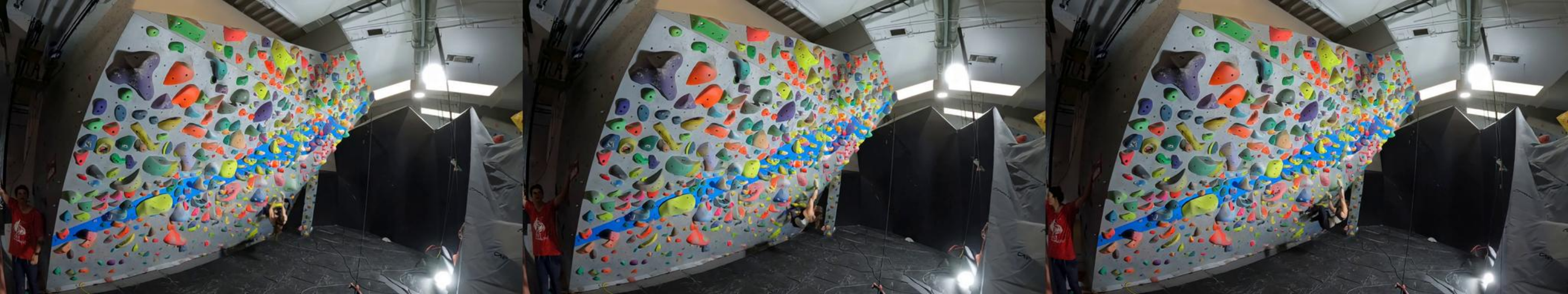}
                        \caption{Cosmos-Transfer2.5-2B (blur)}
                    \end{subfigure}
                    \begin{subfigure}{0.48\linewidth}
                        \centering
                        \includegraphics[width=\linewidth]{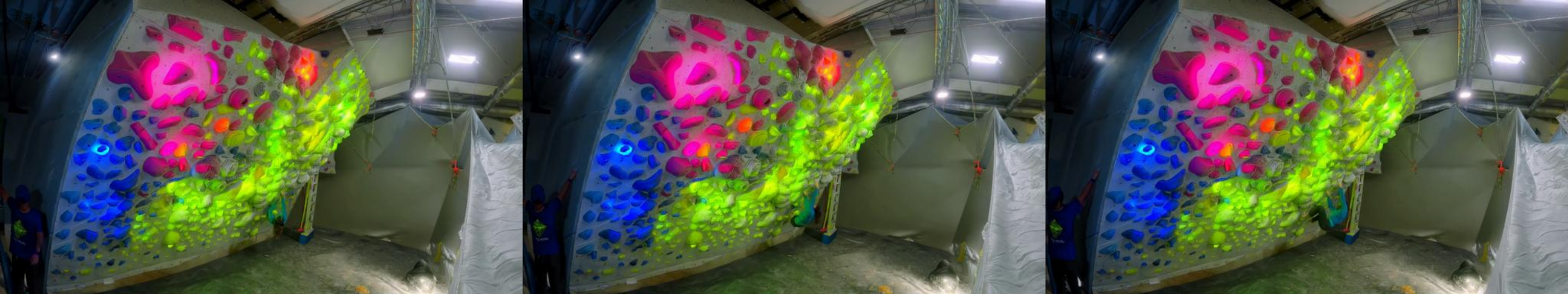}
                        \caption{Cosmos-Transfer2.5-2B (edge)}
                    \end{subfigure}

                    \begin{subfigure}{0.48\linewidth}
                        \centering
                        \includegraphics[width=\linewidth]{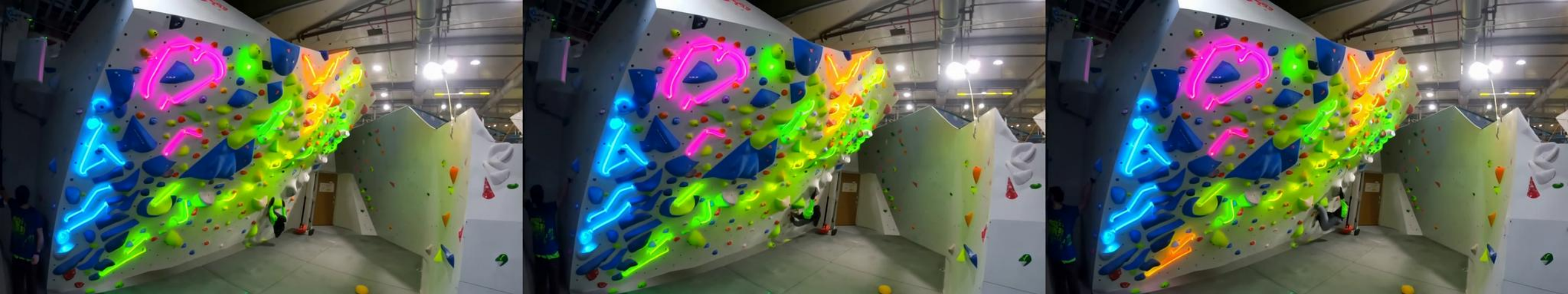}
                        \caption{Cosmos-Transfer2.5-2B (depth)}
                    \end{subfigure}
                    \begin{subfigure}{0.48\linewidth}
                        \centering
                        \includegraphics[width=\linewidth]{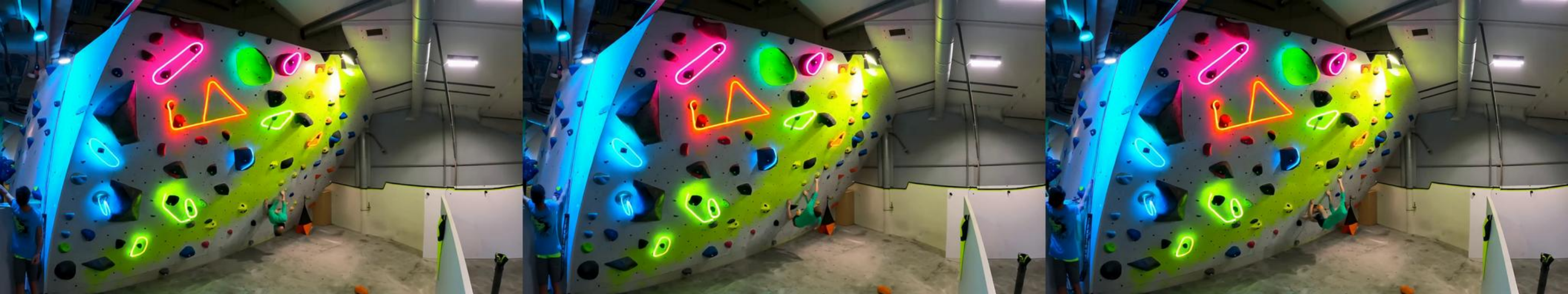}
                        \caption{Cosmos-Transfer2.5-2B (segmentation)}
                    \end{subfigure}

                    \begin{subfigure}{0.48\linewidth}
                        \centering
                        \includegraphics[width=\linewidth]{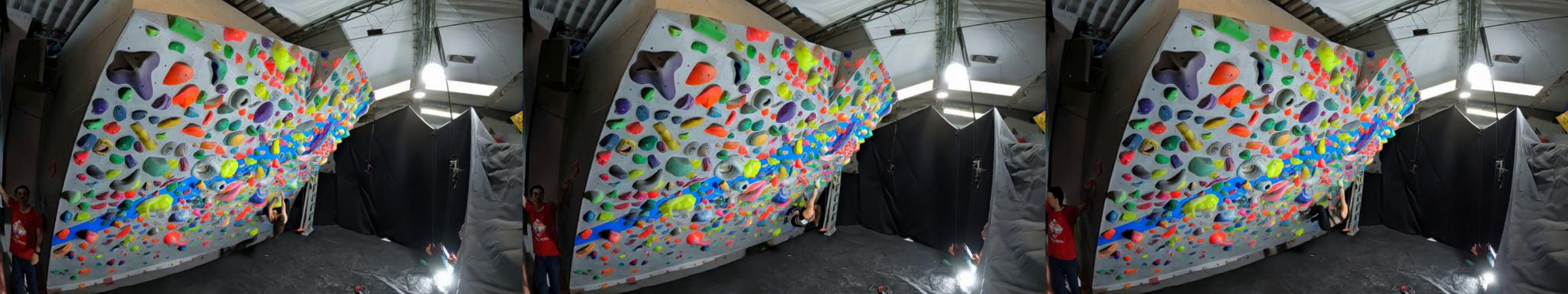}
                        \caption{Cosmos-Transfer2.5-2B (multi-control generation)}
                    \end{subfigure}
                \end{minipage}
            \end{tcolorbox}

            \begin{tcolorbox}[colback=white, colframe=gray, arc=4mm, boxrule=0.7pt, width=0.98\linewidth, left=3pt, right=3pt, top=3pt, bottom=3pt]
                \begin{minipage}{0.98\linewidth}
                    \centering
                    \begin{subfigure}{0.48\linewidth}
                        \centering
                        \includegraphics[width=\linewidth]{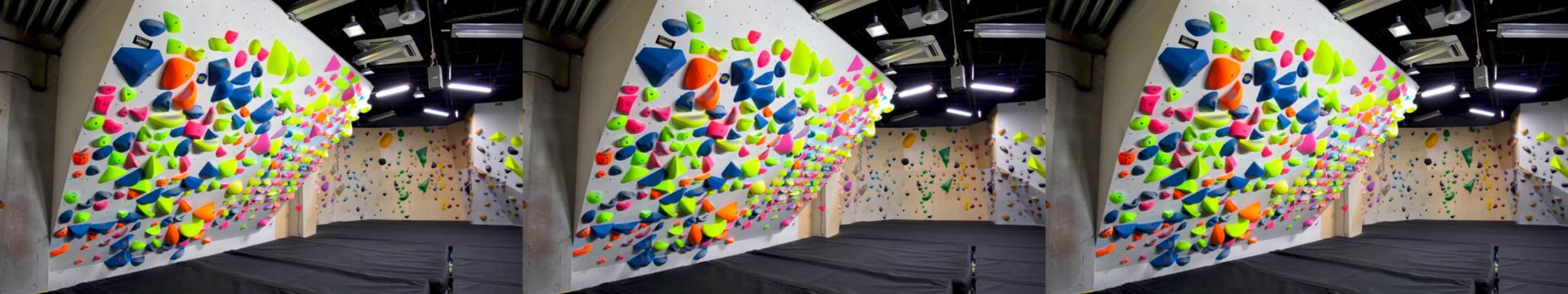}
                        \caption{Wan2.2-Fun-A14B-Control (blur)}
                    \end{subfigure}
                    \begin{subfigure}{0.48\linewidth}
                        \centering
                        \includegraphics[width=\linewidth]{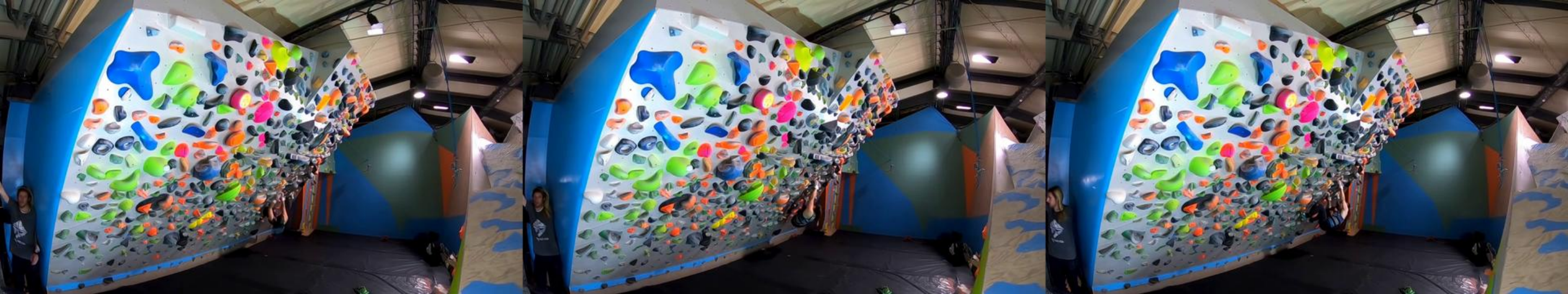}
                        \caption{Wan2.2-Fun-A14B-Control (edge)}
                    \end{subfigure}

                    \begin{subfigure}{0.48\linewidth}
                        \centering
                        \includegraphics[width=\linewidth]{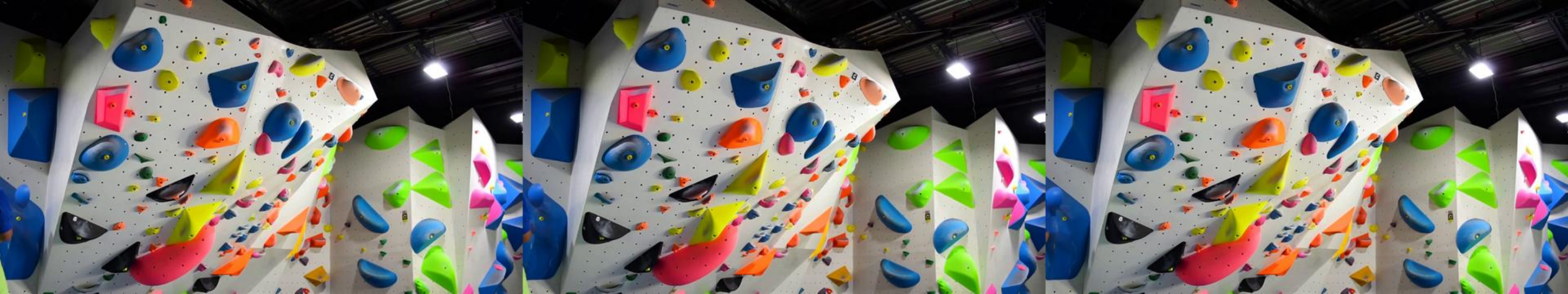}
                        \caption{Wan2.2-Fun-A14B-Control (depth)}
                    \end{subfigure}
                    \begin{subfigure}{0.48\linewidth}
                        \centering
                        \includegraphics[width=\linewidth]{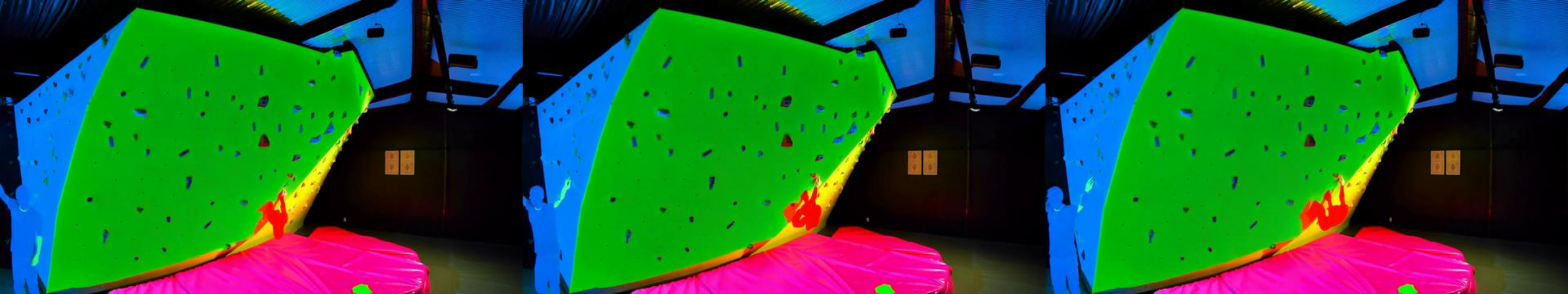}
                        \caption{Wan2.2-Fun-A14B-Control (segmentation)}
                    \end{subfigure}
                \end{minipage}
            \end{tcolorbox}
        \end{minipage}
    \end{tcolorbox}
    \caption{\textbf{Example of human domain control signals and model generations from PAI-Bench-C.} \textit{Best viewed with zoom.}}
    \label{fig:transfer_examples_action}
\end{figure*}

We present representative test cases and corresponding model-generated results from the autonomous vehicle, robotics, and human domains of PAI-Bench-C, as illustrated in Figure~\ref{fig:transfer_examples_av}, Figure~\ref{fig:transfer_examples_robotics}, and Figure~\ref{fig:transfer_examples_action}, respectively.
\section{More Details about PAI-Bench-U}

\begin{figure*}[ht]
    \centering
    \includegraphics[width=\linewidth]{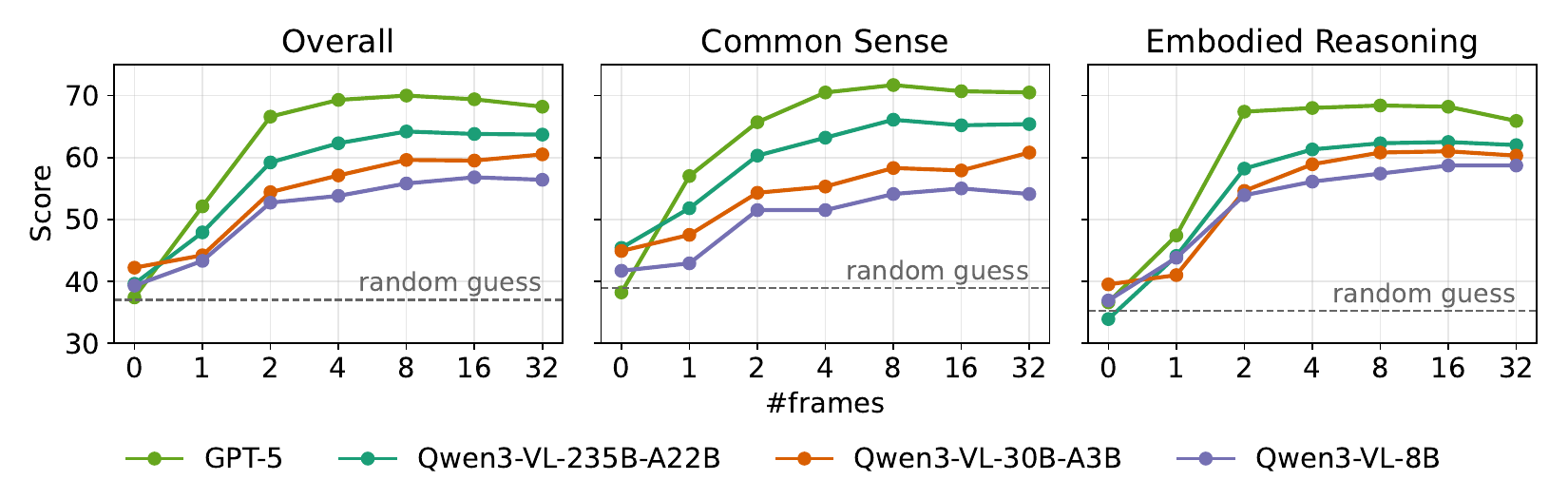}
    \vspace{-18pt}
    \caption{\textbf{Accuracy versus number of input frames on PAI-Bench-U.} The input frames are uniformly sampled from the video. The dashed horizontal lines denote the random-guess baselines.}
    \label{fig:metrics_three_col_multi_model}
\end{figure*}

\subsection{Detailed Data Curation Process} \label{sec:appendix-reason}
This section details the curation process for the embodied reasoning benchmark. To ensure rigorous evaluation, we adhere to two core design principles: 1) \textbf{Unified Question Templates:} We standardize question formulation to ensure reasoning is grounded in visual input rather than exploited through textual biases. 2) \textbf{Unified Action Granularity:} To resolve ambiguity in next-action prediction (where multiple granularities, e.g., ``grab can'' vs. ``move hand left'', may be valid), we adopt the hierarchical taxonomy proposed by~\cite{belkhale2024rthactionhierarchiesusing}. We categorize behaviors into atomic-level \textit{actions}, coarse-grained \textit{subtasks}, and dataset-specific \textit{goals}. The curation details for specific data sources are as follows:

\thinparagraph{RoboVQA.}
We curated 101 clips from the RoboVQA~\cite{sermanet2023robovqamultimodallonghorizonreasoning} validation split. We formulated binary multiple-choice questions (Yes/No) to evaluate two capabilities: \textit{task-completion verification} (determining if an instruction was successfully executed) and \textit{affordance} (assessing task feasibility given the visual context).

\thinparagraph{RoboFail.}
We manually annotated 100 examples from RoboFail~\cite{liu2023reflectsummarizingrobotexperiences} to assess reasoning under failure conditions. These samples evaluate action affordance and task completion verification in challenging scenarios characterized by: (1) complex temporal dynamics requiring observant perception, (2) physical constraints impeding execution, and (3) nuanced reasoning requirements beyond simple perception mismatches.

\thinparagraph{BridgeData V2.}
We processed the BridgeData V2~\cite{walke2024bridgedatav2datasetrobot} validation split to extract 100 clips for multiple-choice QA. Conditioned on the task instruction and the visual history, the model is queried to predict the most plausible immediate next action.

\thinparagraph{AgiBot.}
We derived 100 multiple-choice samples from AgiBot~\cite{bu2025agibot}. For each clip, we provide the high-level task information and ask the model to identify the correct next subtask. Distractors are randomly sampled from the subtask sequence within the clip's full trajectory.

\thinparagraph{HoloAssist.}
We constructed 100 QA pairs from HoloAssist~\cite{HoloAssist2023}. Providing the coarse-grained action annotation as the overall goal, we query the model for the most likely next subtask. Distractors are randomly sampled from other fine-grained action annotations associated with that coarse-grained goal.

\thinparagraph{AV (Proprietary).}
We curated 100 videos from a proprietary dataset to construct multiple-choice QA pairs. These videos exhibit diverse lateral and longitudinal behaviors with rich agent interactions. The questions are designed to: (1) predict the likely next immediate action of the ego vehicle, (2) verify the completion of a previously executed action, and (3) assess the affordance of specific actions within the given scenario.

\subsection{Detailed Experiment Setup}

\thinparagraph{Inference Prompt Template.} To ensure a rigorous and reproducible evaluation across all MLLMs, we employ a unified system prompt for PAI-Bench-U, as delineated in Figure~\ref{fig:reason-system-prompt}. Specifically, the model is instructed to generate an intermediate rationale within \texttt{<think>} tags prior to providing the final conclusion within \texttt{<answer>} tags. The input context consists of temporally sampled video frames ($Frames_{0} \dots Frames_{N}$) followed by the specific question, ensuring the model processes the visual narrative before addressing the query.

\begin{figure}[h]
    \begin{tcolorbox}[
        title=\textbf{PAI-Bench-U Inference Prompt Template},
        colback=gray!3,
        colframe=black!60,
        fonttitle=\bfseries,
        sharp corners,
        boxrule=0.6pt
    ]
    You are a helpful assistant. Answer in the format:
    <think>reasoning</think>

    <answer>answer</answer>.

    \vspace{0.2cm}
    \textbf{Input}\\
    \textit{\{Sampled Frames 0\}}\\
    \textit{\{Sampled Frames 1\}}\\
    \ldots\\
    \textit{\{Sampled Frames $N$\}}\\

    \vspace{0.2cm}
    Question: \textit{\{Question\}}
    \end{tcolorbox}
    \vspace{-6pt}
    \caption{\centering The standardized inference prompt template utilized in PAI-Bench-U.}
    \vspace{-6pt}
    \label{fig:reason-system-prompt}
\end{figure}

\thinparagraph{Experimental Settings}
We employ default inference settings for all proprietary models. Specifically for GPT-5~\cite{openai2025gpt5}, we differentiate between \textit{medium reasoning effort} (enabling "thinking mode") and \textit{minimal reasoning effort} (disabling "thinking mode"). The model versions and checkpoints are detailed in Table~\ref{tab:modelcfg}.Open-source models are served via the vLLM backend~\cite{kwon2023efficient} using their respective official default parameters. 

\begin{table}[h]
    \centering
    \caption{\centering List of proprietary models with their versions evaluated in our experiments.}
    \resizebox{0.6\linewidth}{!}{%
        \begin{tabular}{lcc}
        \toprule
        \textbf{Model} & \textbf{Vendor} & \textbf{Version} \\
        \midrule
        GPT-4o & OpenAI & gpt-4o-2024-08-06 \\
        GPT-5 & OpenAI & gpt-5-2025-08-07 \\
        Claude 3.5 Sonnet & Anthropic & claude-3-5-sonnet-20241022-v2 \\
        \bottomrule
        \end{tabular}%
    }
    \label{tab:modelcfg}
\end{table}

\subsection{Discussion on Input Frame Count}
To analyze how the number of sampled video frames affects model performance in PAI-Bench-U, we systematically evaluate GPT-5~\cite{openai2025gpt5}, Qwen3-VL-235B-A22B~\cite{qwen_vl_3}, Qwen3-VL-30B-A3B~\cite{qwen_vl_3}, and Qwen3-VL-8B~\cite{qwen_vl_3} with 0, 1, 2, 4, 8, 16, and 32 uniformly sampled frames.

\thinparagraph{Impact of Temporal Sampling Rate.}
As illustrated in Figure~\ref{fig:metrics_three_col_multi_model}, increasing the input frame count ($<8$) initially yields consistent performance gains, as the model acquires the effective visual information necessary for question answering. However, performance saturates and remains stable beyond 8 frames. We attribute this plateau to two primary factors. First, we hypothesize a reasoning bottleneck: while the visual information provided by 8 frames is likely sufficient for the task, the models' failure to achieve further gains highlights intrinsic limitations in their spatial-temporal reasoning capabilities, rather than a deficiency in visual data. Second, excessive sampling may introduce temporal ambiguity. High frame rates significantly reduce the visual variance between adjacent frames, which can paradoxically hinder motion perception. For instance, when determining the movement direction of a robotic arm, sparse sampling (e.g., start and end frames) presents distinct visual states that make the trajectory obvious. Conversely, dense sampling results in minute inter-frame differences. Current models often lack the sensitivity to process these fine-grained temporal shifts, leading them to misinterpret the high similarity between consecutive frames as a static scene, thereby degrading judgment.

\subsection{User Study Details}
\begin{figure*}[tbp]
    \centering
    \includegraphics[width=0.85\textwidth]{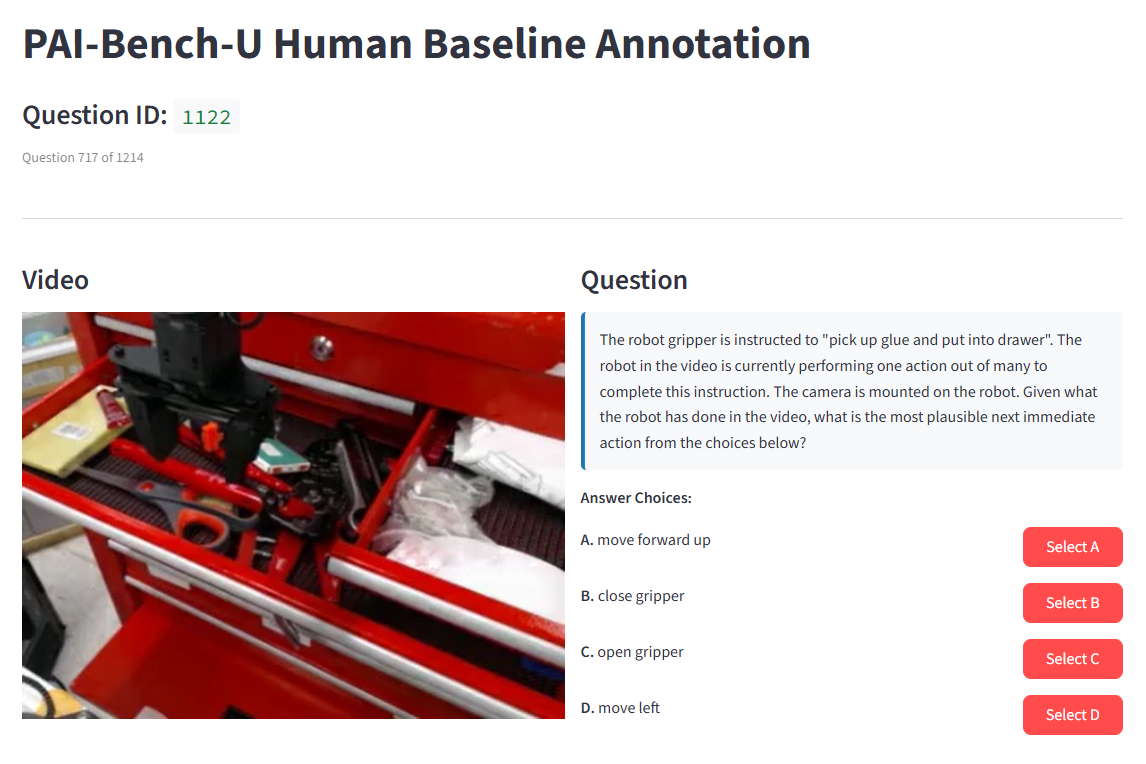}
    \caption{User study interface for establishing human baselines on PAI-Bench-U.}
    \label{fig:user_study_interface_reason}
\end{figure*}

To establish a human performance baseline for PAI-Bench-U, we conducted a user study wherein participants responded to benchmark questions conditioned on the provided video clips. The resulting aggregated accuracy serves as a comparative reference for model evaluation. The annotation interface utilized for this data collection is illustrated in Figure~\ref{fig:user_study_interface_reason}.
\section{PAI-Bench Leaderboard Visualization}

\begin{figure*}[t]
    \centering

    \begin{minipage}{0.47\linewidth}
        \centering
        (a) PAI-Bench-G: Domain radar comparison\vspace{0.3em}

        \includegraphics[width=\linewidth]{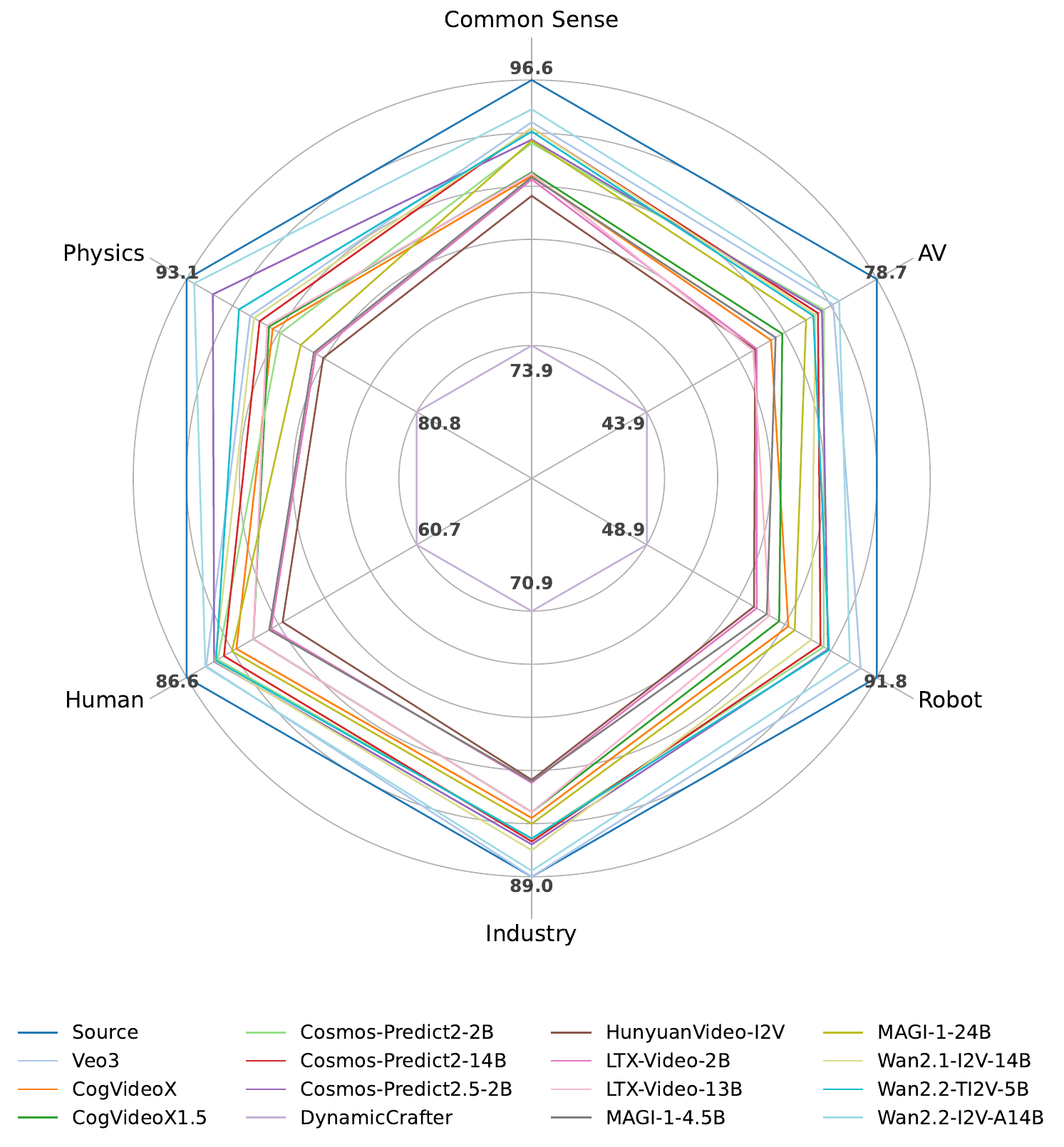}
    \end{minipage}
    \hfill
    \begin{minipage}{0.51\linewidth}
        \centering
        (b) PAI-Bench-G: Quality radar comparison\vspace{0.3em}

        \includegraphics[width=\linewidth]{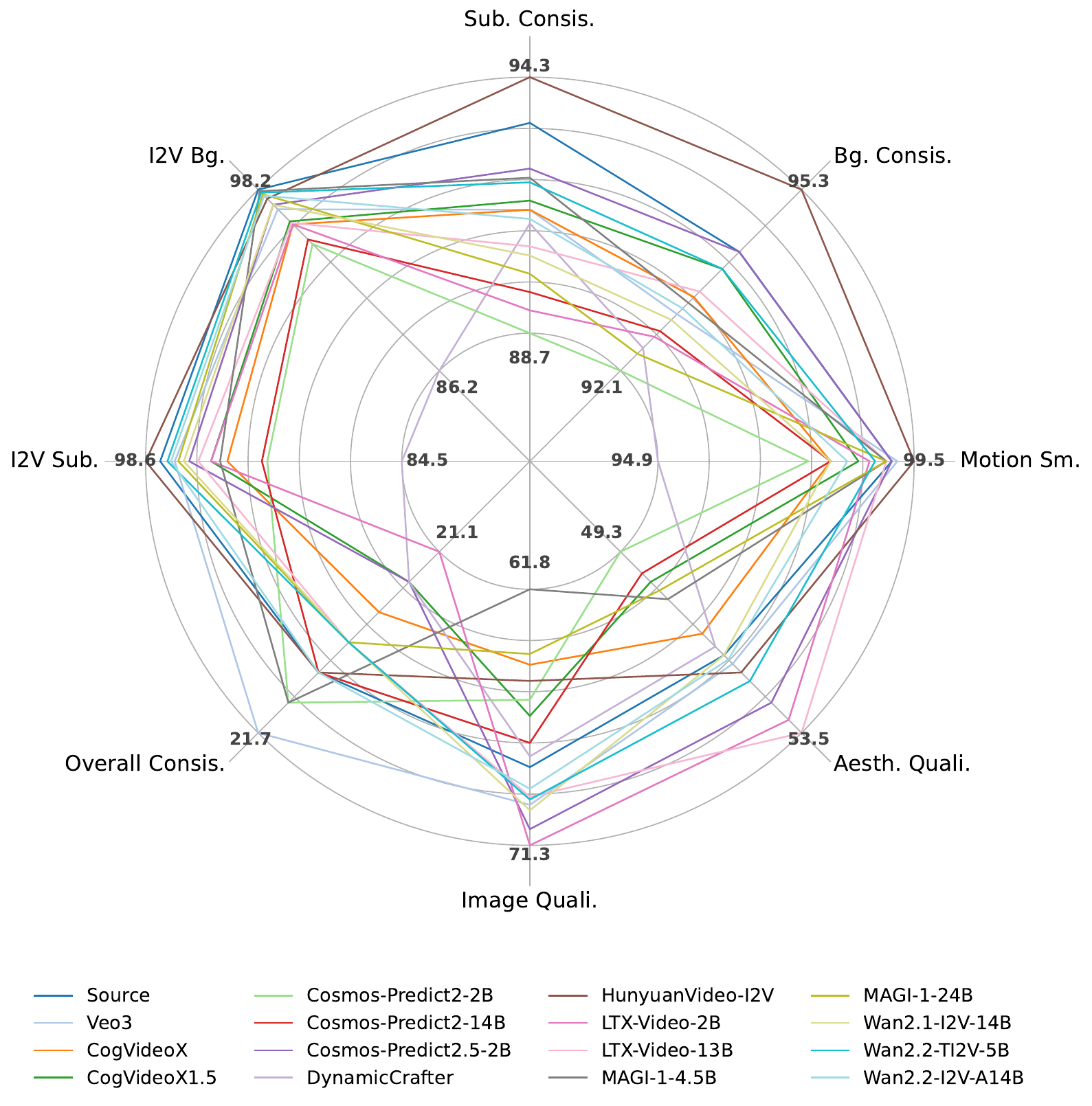}
    \end{minipage}

    \vspace{0.8em}

    \begin{minipage}{0.48\linewidth}
        \centering
        (c) PAI-Bench-C: Radar comparison\vspace{0.3em}

        \includegraphics[width=\linewidth]{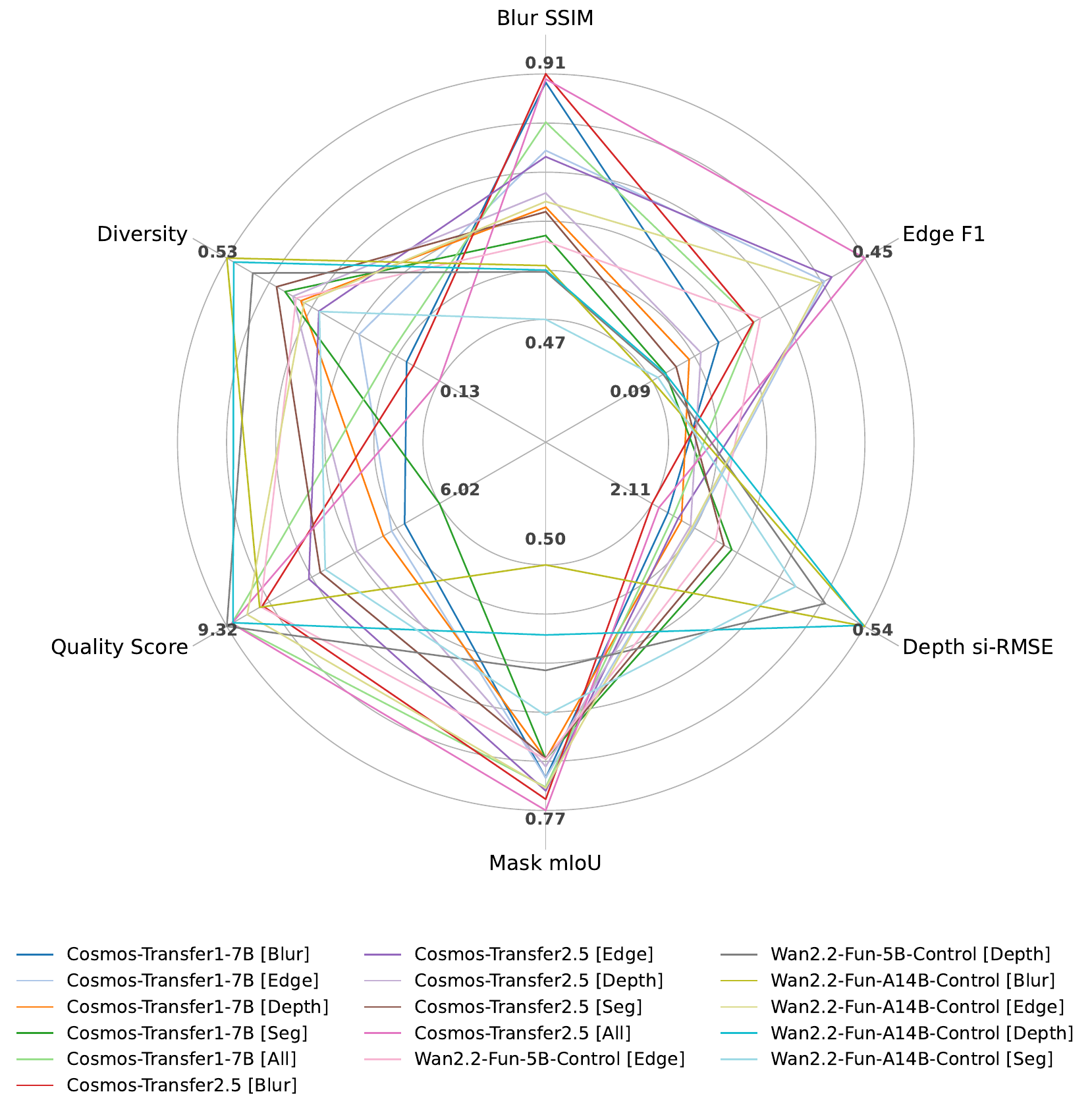}
    \end{minipage}
    \hfill
    \begin{minipage}{0.50\linewidth}
        \centering
        (d) PAI-Bench-U: Radar comparison\vspace{0.3em}

        \includegraphics[width=\linewidth]{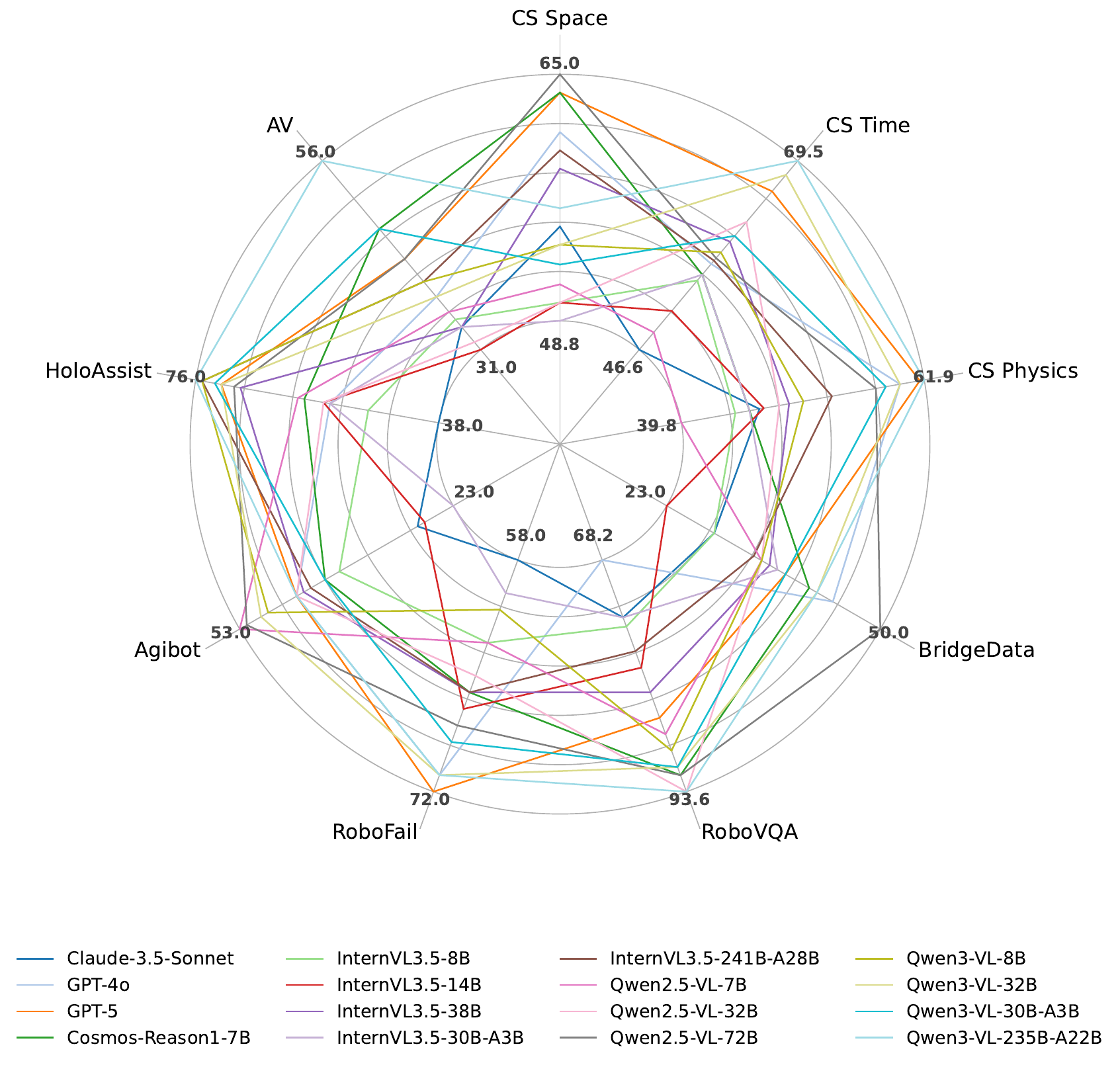}
    \end{minipage}

    \caption{
        \textbf{Radar chart visualizations of model capabilities across PAI-Bench tracks.}
    }
    \label{fig:pai_bench_radar_overview}
\end{figure*}

To provide a holistic view of model performance across the diverse dimensions of PAI-Bench, we visualize the evaluation results using radar plots (Figure~\ref{fig:pai_bench_radar_overview}). To account for the heterogeneity of evaluation metrics, we normalize scores to ensure comparative consistency.
\section{Discussion and Future Work}
In this section, we identify several open challenges and outline promising directions for future research:

\noindent \textbf{Enhancing Metric Robustness with Advanced Encoders.}
The precision of overall consistency metrics is tied to the capabilities of the underlying video--text foundational models. Currently, models like VCLIP~\cite{weng2023openvcliptransformingclipopenvocabulary} exhibit constraints when processing lengthy and highly detailed textual prompts. Future work could leverage next-generation encoders to improve sensitivity to complex instructions, thereby enabling more granular consistency evaluations.

\noindent \textbf{Addressing Conservative Generation Strategies.}
Our user study highlights a nuanced phenomenon in video generation: a trade-off between motion dynamism and generation safety. We observe that some models tend to adopt a conservative strategy that prioritizing static fidelity (e.g., holding a racket) over complex, high-risk dynamics (e.g., rapidly swinging it) to avoid artifacts. While this has a limited impact on quantitative rankings, developing methods that encourage risk-taking in motion generation without compromising visual quality remains a vital direction.

\noindent \textbf{Necessity and Imperfection of MLLM Judges.}
While we utilize SOTA MLLMs like GPT-5~\cite{openai2025gpt5} to assess high-level semantics, automated evaluation remains an evolving field. Even the most advanced models possess inherent boundaries as visual judges, particularly when interpreting temporal dynamics in video. However, given the lack of effective alternatives for scalable semantic assessment, MLLM-based evaluation represents the current best practice.

\end{document}